\documentclass[11pt]{article}

\usepackage[english]{babel}
\usepackage{standalone}
\usepackage[T1]{fontenc}
\usepackage[singlespacing]{setspace}
\usepackage[a4paper,top=2.5cm,bottom=2cm,left=2.4cm,right=2.4cm]{geometry}

\usepackage{amsmath, amssymb, amsthm, amsbsy, amsfonts, natbib, wrapfig,graphicx, comment, enumitem, algorithm, algcompatible,indentfirst} 
\usepackage[colorinlistoftodos]{todonotes}
\usepackage[colorlinks=true, allcolors=blue]{hyperref}
\usepackage[noend]{algpseudocode}
\usepackage{caption}
\usepackage{colortbl, titlesec}
\usepackage[font=small,skip=0pt]{caption}

\theoremstyle{plain}
\newtheorem{thm}{Theorem}
\newtheorem{defn}{Definition}

\newtheorem{cor}{Corollary}
\newtheorem{pro}{Proposition}
\newtheorem{rem}{Remark}
\newtheorem{cla}{Claim}

\newcommand{\x}{\mathbf{x}}
\newcommand{\X}{\mathbf{X}}

\newcommand{\z}{\mathbf{z}}

\newcommand{\e}{\mathbf{e}}

\newcommand{\E}{\mbox{E}}
\newcommand{\V}{\mbox{V}}
\newcommand{\A}{\mathbf{A}}

\newcommand{\bs}{\boldsymbol}

\newcommand{\1}{\mathbf{1}}
\newcommand{\0}{\mathbf{0}}
\newcommand{\diag}{\mbox{diag}}

\allowdisplaybreaks
\titlespacing*{\section}
{0pt}{6pt plus 3pt minus 3pt}{1pt plus 1pt}
\titlespacing*{\subsection}
{0pt}{6pt plus 3pt minus 3pt}{1pt plus 1pt}

\title{\Large AdaPtive Noisy Data Augmentation for Regularized Estimation of Undirected Graphical Models}
\author{\normalsize Yinan Li$^{1}$, Xiao Liu$^2$, and Fang Liu$^1$\footnote{Corresponding author email: fang.liu.131@nd.edu}\\
\small$^1$ Department of Applied and Computational Mathematics and Statistics\\
\small$^2$ Department of Psychology\\
\small University of Notre Dame, Notre Dame, IN 46556, U.S.A. }
\date{}

\begin{document}
\maketitle
\begin{abstract}	
We propose an AdaPtive Noise Augmentation (PANDA) technique to regularize the estimation and construction of undirected graphical models. PANDA iteratively optimizes the objective function given the noise augmented data until convergence to achieve regularization on model parameters.  The augmented noises can be designed to achieve various regularization effects on graph estimation, such as the bridge (including lasso and ridge), elastic net, adaptive lasso, and SCAD penalization; it also realizes the group lasso and fused ridge. We examine the tail bound of the noise-augmented loss function and establish  that the noise-augmented loss function and its minimizer converge almost surely to the expected penalized loss function and its minimizer, respectively. We derive the asymptotic distributions for the regularized parameters through PANDA in generalized linear models, based on which, inferences for the parameters can be obtained simultaneously with variable selection. We show the non-inferior performance of PANDA in constructing graphs of different types in simulation studies and apply PANDA to an autism spectrum disorder data to construct a mixed-node graph. We also show that the inferences based on the asymptotic distribution of regularized parameter estimates via PANDA achieve nominal or near-nominal coverage and are far more efficient, compared to some existing post-selection procedures.  Computationally, PANDA can be easily programmed in software that implements (GLMs) without resorting to complicated  optimization techniques.
\vspace{6pt}

\noindent \textbf{keywords}:  adjacency matrix, augmented Fisher information, generalized linear model (GLM), maximum a posterior estimation,   sparsity, inference
\end{abstract}

\pagebreak

\section{Introduction}\label{sec:intro}
\subsection{Noise Injection}
\emph{Noise injection (NI)} is a simple and effective  regularization technique that can improve the generalization ability of statistical learning and machine learning methods. We can roughly classify the NI techniques into two types. The first type refers to additive or  multiplicative noises injected into the observed data or latent variables without changing the dimension of the original data $(n\times p)$. We refer to the second type as  \emph{noise augmentation} which expands the dimensionality of the original data (either $n$ or $p$ increases). Both types would lead to less overfitting and smaller generalization errors of the trained models and and parameters  as compared to those learned from the original data without any regularization.  

NI has wide applications in regularizing and learning  neural networks (NN). \citet{ni19922} proves that injecting noises to the input layers when training NN decreases the learned NN's sensitivity to small input perturbation. \citet{ni19921} interpret NI in the input nodes from the perspective of kernel smoothing in classification and mapping problems. The best known NI technique in NN training is the multiplicative Bernoulli noise, which is shown to achieve the $l_2$ regularization effect (for dropout) \citep{dropout14} or the $l_2$ plus some sparsity regularization on model parameters in the setting of generalized linear models (GLMs) \citep{Kang2016}.  \citet{Grandvalet1997} and \citet{Wager2013} also show that Bernoulli and constant-variance Gaussian NI in GLMs is equivalent to the Tikhonov regularization, after taking expectation of the  second order approximated loss function over the distribution of injected noises. Whiteout \citep{whiteout} injects adaptive additive and multiplicative Gaussian noises in NNs, where the variance of the Gaussian noise is a function of NN parameters and contains tuning parameters that lead to a variety of regularizers, including the bridge, ridge, lasso, adaptive lasso, elastic net,  and group lasso.   
\citet{Gal} develop a theoretical framework that connects dropout in deep NNs with approximate Bayesian inference in deep Gaussian processes.  \citet{Noh2017} suggest that NI regularization optimizes the lower bound of the objective function marginalized over the distribution of hidden nodes. 


Given the success of NI in regularizing NNs, one would conjecture that, conditional on properly  designed  noises, NI can be potentially useful in regularizing other types of large and complex models. That is indeed the case; but we only found a couple of cases beyond the framework of NNs. In the first case, NI is applied to the linear discriminant analysis \citep{Marina1999}, where redundant features are augmented to the observed data ($p$ increases while $n$ remains the same), yielding similar effects as other regularization techniques. In the second case, the $l_2$ regularization in linear regression setting can be realized by appending a $p\times p$ diagonal matrix $\sqrt{\lambda}\mathbf{I}$ (where $\lambda$ is the tuning parametric) to the design matrix $\X$ and $p$ rows of 0 to the centered outcome $Y$  \citep{David1974, learning96}. Both cases also happen to be noise augmentation.

In this discussion, we explore the utility of NI, more specifically NA, in regularizing undirected graphical models (UGMs), where the injected noises are adaptive to the most updated parameters during an iterative computation procedure rather than being drawn from a fixed distribution, and can be designed to achieve various regularization effects on the model parameters.


\subsection{Undirected Graphical Models (UGM)}
A graphical model is a probabilistic model that expresses the conditional dependence structure among random variables in a graph. The random variables are often referred to as the nodes of the graph. If two nodes are dependent conditional on all the other nodes in the graph, then  an edge is drawn between the two node; otherwise, there is no edge. The edges in a UGM have no direction.  We denote a UGM by $\mathcal{G}_p(\X, \A)$ with $p$ nodes, where $\X$ refers to the data observed in the $p$ nodes and $\A$ is a $p\times p$ unknown symmetric adjacency matrix (weighted or unweighted). A non-zero entry $a_{ij}$  represents conditional dependence between nodes $i$ and $j$. Construction and estimation of $\A$ given $X$ is often the main goal in UGM problems. 

Many real-life graphs are believed to be sparse, such as biological networks \citep{sparse2008}, meaning that the proportion of none-zero $a_{ij}$'s in $\A$ is small. In addition, data collected for estimating edges often have $n<p$. Given both the practical and technical needs,  regularization techniques that promote sparsity in $\A$ are often employed when constructing a UGM. One popular approach is the neighborhood selection (NS) method which estimates $\A$ by columnwise modeling the conditional  distribution of each node  given all the other nodes, leading to $p$ regression models. When the conditional distributions belong to an exponential family, the generalized linear models (GLM) can be employed, including linear regression for  Gaussian nodes,  logistic regression and Ising models for Bernoulli nodes \citep{ising2010, binary2009,Kuang2017, multinomial2011}, and Poisson regression for count nodes \citep{poisson2012}. Also noted is that the nodes in a graph do not have to be of the same type; and there exist works for mixed graph models (MGMs) with nodes of mixed types \citep{mixed2013,yang2012,yang2014}.  In terms of the regularization techniques that promote sparsity in the relationships among the nodes, the lasso \citep{Meinshausen2006}, the graphical Dantzig selector \citep{Yuan2010,Cai2011}, the graphical scaled lasso \citep{Sun2012}  and the SQRT-Lasso \citep{Liu2012,Belloni2012} have been proposed, among others.

When all $p$ nodes in a UGM follow a multivariate Gaussian distribution, the UGM is referred to as the Gaussian graphical model (GGM). There exist approaches for edge estimation specifically for GGM in addition to the general NS approach mentioned above. For example, \citet{Huang2006} obtain the Cholesky decomposition (CD) of the precision matrix $\Omega$ of the multivariate Gaussian distribution, and then apply the  $l_1$ penalty to the elements of the triangular matrix from the CD. \citet{Levina2008} apply the adaptive banding method \citep{Bickel2008} with a nested Lasso penalty on the regression coefficients of linear regressions after the  CD.
\citet{weidong2015} reformulate the NS as a regularized quadratic optimization  problem without directly employing the Gaussian likelihood.  Another line of research focuses on estimating $\Omega$ as a whole while ensuring its positive-definiteness (PD) and sparsity. For example, \citet{Yuan2007} propose a $l_1$ penalized likelihood approach that accomplishes model selection and estimation simultaneously and also ensures the PD of the estimated $\Omega$. \citet{Friedman2008,Banerjee2008,Rothman2008} propose efficient computational algorithms to implement the $l_1$ penalized likelihood approach.   Theoretical properties of the penalized likelihood methods are developed in \citet{Ravikumar2008,Rothman2008,Lam2009}.



\subsection{Our Contributions}
We propose AdaPtive Noisy Data Augmentation  (PANDA) - a general, novel, and effective NI technique to regularize the estimation and construction of UGMs. Denote the sample size of the observed data by $n$, PANDA augments the $n$ observations with properly designed $n_e$ noise terms to achieve the desired regularization effects on model parameters. One requirement on $n_e$ is $n^*=n+n_e>p$, which allows for the ordinary least squares (OLS) or maximum likelihood estimation (MLE) approaches to be employed to estimate the model parameters without resorting to complicated  algorithms to optimize objective functions with regularizers.

To the best of our knowledge,
PANDA is the first  NI, more specifically, the data or noise augmentation technique for regularizing UGMs. Our overarching goal is to show that PANDA delivers non-inferior performance while enjoying learning, inferential, and computational advantages compared to the existing UGM estimation approaches. Our contributions are listed below.
\vspace{-9pt}
\begin{enumerate}[leftmargin=0.2in]
\setlength\itemsep{0.0em}
\item By properly designing the variance of the augmented  noise, PANDA can achieve various regularization effects, including bridge ($l_{\gamma}$) ($0\!<\gamma\!\le2$) with lasso ($\gamma\!=\!1$) and ridge ($\gamma\!=\!2$) as special cases, elastic net ($l_1+l_2$), SCAD, group lasso, and graphical ridge for single graph estimation. 
\item PANDA can be used to construct mixed graph models, without additional complexity compared to constructing a graph with the same types of nodes.
\item Computation in PANDA is straightforward and only employs the OLS in linear regression and the MLE in GLMs to iteratively estimate the model parameters on the augmented data. The variance terms of the augmented noise are adaptive to the most updated parameter estimates  until the algorithm converges.  
\item  We establish the Gaussian tail of the noise-augmented loss function  and the almost sure convergence to its expectation as $n_e$ or $m$ increases, which is a penalized loss function with the targeted regularizer, providing theoretical justification for PANDA as a regularization technique and that the noise-augmented loss function is trainable for practical implementation. 
providing the theoretical justification for PANDA.
\item We connect PANDA with the  Bayesian framework and show that the regularized parameter estimate in PANDA is equivalent to the ``maximum a posterior'' (MAP) in the Bayesian framework.
\item PANDA offers an alternative approach to post-selection procedures for obtaining inferences for regression coefficients from GLMs with sparsity regularization, whether the estimates are zero-valued or not. 
Our empirical results suggest the inferences based on PANDA are valid and more efficient compared to some existing post-selection procedures.
\end{enumerate}

The rest of the paper is organized as follows. Section \ref{sec:single} presents several PANDA algorithms and their associated regularization effects for constructing GGM and UMG in general. Section \ref{sec:EB} presents the Bayes interpretation for PANDA. Section \ref{sec:theory} establishes the consistency on the noise-augmented loss function and the regularized parameter estimates,  presents the Fisher information of the model parameters in augmented data and a formal test for convergence in PANDA algorithms. It also provides the asymptotic distributions for the parameter estimates via PANDA in the GLM setting. Section \ref{sec:simulation} compares PANDA to the constrained optimization approach in edge detection for  several types of UGMs, and to the post-selection inferential approach in statistical inferences in  GLMs. Section \ref{sec:case} applies PANDA to estimating the association among the attributes in a real autism spectrum disorder data set. Section \ref{sec:discussion} provides some concluding remarks and offers future research directions on PANDA.

\section{Methodology}\label{sec:single}
In this section, we present PANDA to regularize the construction of GGMs and UGMs  via NS (Sec \ref{sec:NS}, \ref{sec:NSGGM}, \ref{sec:nongaussian1}). In the case of GGM construction, in addition to NS, PANDA can also be implemented in the context of other types of regularization than the NS framework, which will be detailed in Sec \ref{sec:CD}, \ref{sec:scio}, and \ref{sec:graphical}.

\subsection{Neighborhood selection (NS) via PANDA in UGM}\label{sec:NS}
Let $j=1,\ldots,p$ be the index for the $p$ nodes in a UGM. The neighborhood selection (NS) approach, as referred to in this paper, for constructing a UGM assumes the conditional  distribution of $X_j$ given  $\X_{-j}=(X_1,\ldots,X_{j-1},X_{j+1},\ldots,X_p)^T$ comes from an exponential family
\begin{equation}\label{eqn:expfam}
p(X_j|\X_{-j})=\exp\left(X_j\eta_j-B_j(\eta_j)+h_j(X_j)\right),
\end{equation}
where $\eta_j=\theta_{j0}+\sum_{k\ne j}\theta_{jk}X_k$ if the canonical link is used  (e.g., the identity link for Gaussian $X_j$; the logit link for Bernoulli $X_j$). Eqn (\ref{eqn:expfam}) suggests that the relationship among the nodes can be recovered by running GLMs $p$ times; that is, there is no edge between nodes $j$ and $k$ in the graph if $\theta_{jk}=\theta_{kj}=0$; otherwise, the two nodes are connected with an edge. \citet{yang2012,UGMEXP2015} establish, under some regularity conditions,  that the structure of a UGM can be recovered exactly
via M-estimators with high probability when node-conditional distributions belong to an exponential family in Eqn (\ref{eqn:expfam}). Regularization (e.g., sparsity regularization) is often imposed when running the node-wise GLM to estimate $\bs{\theta}=\{\theta_{jk}\}$,
followed by developing an optimization algorithms to solve for $\hat{\bs{\theta}}$ (refer to Section \ref{sec:intro} for some existing work in this direction). When a graph contains nodes of different types (e.g. node $j$ is Gaussian and node $k$ is Bernoulli), due to the asymmetry in the regression modles on $X_j$ and $X_k$, $\theta_{jk}$ and $\theta_{kj}$ would have different interpretation from a regression perspective. However,  the actual magnitude of $\theta_{jk}$ would not be important if the goal is to decide there is an edge between $j$ and $k$ or not.

\begin{wrapfigure}{r}{0.4\textwidth}
\centering\includegraphics[width=0.35\textwidth]{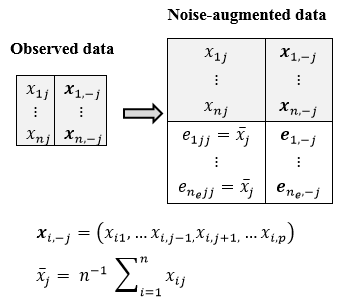}
\caption{A schematic of the data augmentation for a single graph in PANDA}\label{fig:panda1}
\end{wrapfigure}
PANDA estimates $\bs{\theta}$ by first augmenting the observed data $\x$ with a noisy data matrix. We recommend centerizing the observed data on each ``covariate'' node in $\X_{-j}$ in a UGM  and  standardizing all nodes in a GGM (or standardizing $\X_{-j}$ and centering the ``outcome'' node $X_j$) prior to the augmentation. Figure \ref{fig:panda1} depicts a schematic  of the data augmentation step in PANDA for a single graph. The augmented values to  $X_j$ is a constant and is the sample average of the outcome node ($e_{ijj}=\bar{x}_j=0$ for $i=1,\ldots,n_e$ for GGM unless stated otherwise).  The augmented observations for the covariate node $X_k$ ($k\ne j$) are drawn independently from a Gaussian distribution with mean 0 and variance that depends on $\theta_{jk}$ and the tuning parameters (Eqns (\ref{eqn:bridge}) to (\ref{eqn:fusedridge})). We refer to these distributions as the \emph{Noise Generating Distributions} (NGD).
\begin{align}
e_{jk}&\sim N\left(0, \lambda|\theta_{jk}|^{-\gamma}\right) \label{eqn:bridge}\\
e_{jk}&\sim N\left(0, \lambda|\theta_{jk}|^{-1}+\sigma^2\right) \label{eqn:en}\\
e_{jk}&\sim N\left(0, \lambda |{\theta}_{jk}|^{-1}|\hat{{\theta}}_{jk}|^{-\gamma}\right),\mbox{where  $\hat{\theta}_{jk}$ is a consistent estimate for ${\theta}_{jk}$}\label{eqn:adaptivelasso}\\
e_{jk}&\sim N\!\left(0,\frac{\lambda}{|{\theta}_{jk}|}1_{(0,\lambda n_e)}(|{\theta}_{jk}|)+\frac{1}{(a-1)}\! \left(\! \frac{a\lambda}{|{\theta}_{jk}|}\!-\!\frac{\lambda^2n_e}{2{\theta}_{jk}^2}-\!\frac{1}{2} \right) 1_{[\lambda n_e,a\lambda n_e]} (|{\theta}_{jk}|)+\right.\notag\\
&\qquad\quad\left.\frac{(a+1)\lambda^2n_e}{2{\theta}_{jk}^2} 1_{(a\lambda n_e,\infty)}(|{\theta}_{jk}|)\!\right)\!,\mbox{where $1_{(l,u)}(|{\theta}_{jk}|)=1$ if $l<|{\theta}_{jk}|<u$, 0 otherwise.} \label{eqn:scad}
 \end{align}
$\sigma^2\ge0, \lambda>0$, $0\le\gamma<2$, $a>2$ are tuning parameters, either user-specified or chosen by a model selection criterion such as  cross validation (CV), AIC, or BIC.  Different formulation of the variance term leads to different regularization effects on $\theta_{jk}$. Specifically, Eqn (\ref{eqn:bridge}) leads to the bridge-type regularization which including the lasso  ($\gamma=1$) and ridge regression ($\gamma=0$)  as special cases,  Eqn (\ref{eqn:en}) to elastic net,  Eqn (\ref{eqn:adaptivelasso}) to adaptive lasso, and Eqn (\ref{eqn:scad}) to SCAD, respectively. Eqns (\ref{eqn:bridge})  to  (\ref{eqn:scad}) suggest that the dispersion of the noise terms varies by node: nodes associated with small-valued $|{\theta}_{jk}|$ will be augmented with more spread out noises, and those with large-valued $|{\theta}_{jk}|$ will be augmented with noises around zero.

In addition to Eqns (\ref{eqn:bridge})  to  (\ref{eqn:scad}), PANDA can also realize other types of regularization. For example, to simultaneously regularize a group of $q$ nodes that share connection patterns with the same node $k$ (e.g., genes on the same pathway, binary dummy variables created from the same categorical node), we can generate augmented noises $\e=(e_1,\ldots, e_q)^T$ in these $q$ nodes simultaneously from Eqn (\ref{eqn:group}) to yield a group lasso-like penalty on $(\theta_{1k},\ldots,\theta_{qk})$, and from Eqn (\ref{eqn:fusedridge}) to yield  a fused-ridge type penalty on $(\theta_{1k},\ldots,\theta_{qk})$.
\begin{align}
\e &\sim N_{(q)}\left(\mathbf{0},\lambda\mbox{diag}\left\{\left(\textstyle\sum_{j=1}^{q}\theta_{jk}^2\right)^{-1/2}\right\}\right), \label{eqn:group}\\
\e&\sim N_{(q)}\left(0,\lambda(\mathbf{T}\mathbf{T}')\right),\label{eqn:fusedridge} \\
\mbox{where} &\mbox{ entries in $\mathbf{T}$ are $T_{s,s}=1,T_{s+1-s\cdot1(s=q),s}=-1$ for  $s=1,\ldots,q$; and 0 otherwise.}\notag
\end{align}
The group-lasso regularization in Eqn (\ref{eqn:group}) sets $(\theta_{1k},\ldots,\theta_{qk})$  either at zero or nonzero simultaneously, whereas the fused ridge regularization in Eqn (\ref{eqn:fusedridge}) promotes numerical similarity among $(\theta_{1k},\ldots,\theta_{qk})$ in the same group 
We could also obtain a fused-lasso type of regularization on $(\theta_{1k},\ldots,\theta_{qk})$  by letting $T_{jj'}=\lambda|\theta_{jk}-\theta_{j'k}|^{-1}$ ($j\ne j'$) in Eqn \ref{eqn:fusedridge}. However, it does not necessarily outperform the fused ridge regularizer in terms of promoting similarity on parameter estimates. Since the fused ridge is more stable computational in the context of PANDA, we  therefore focus our discussion on the fused ridge in the rest of the paper.

Eqns (\ref{eqn:bridge}) to (\ref{eqn:fusedridge}) suggest that the variance of the augmented noise depend on the unknown $\bs{\theta}$. When implementing PANDA in practice,  we start with some initial values for $\bs{\theta}$ and then estimate it iteratively. In each iteration,  the augmented noises are drawn from the NGD with the variance constructed using the most updated $\bs{\theta}$. 
The iterative procedure continues until the convergence criterion is met.  

\subsubsection{PANDA-NS for GGM}\label{sec:NSGGM}
Let $\X\sim N_{p}(\bs{\mu},\Sigma)$, where $\Sigma$ is the covariance matrix, then the conditional distribution $X_j$ given $\X_{-j}$  is $\mbox{N}(\mu_j+\Sigma_{j,-j}\Sigma_{-j,-j}^{-1}(\X_{-j}-\bs{\mu}_{-j}), \Sigma_{j,j}-\Sigma_{j,-j}\Sigma_{-j,-j}^{-1}\Sigma_{-j,j})$  for $j=1,\ldots,p$, where $\Sigma_{j,j}$ is the $j$-th diagonal element of $\Sigma$, $\Sigma_{-j,-j}$ is the submatrix of $\Sigma$ with the $j$-th row and the $j$-th column removed, and $\Sigma_{j,-j}$ is the $j$-th row of $\Sigma$ with the $j$-th element removed, and $\Sigma_{-j,j}=\Sigma_{j,-j}^T$.  The conditional distribution suggests the following linear model
\begin{align}
&X_j =\alpha_j+\X_{-j}^T\bs{\theta}_j+\epsilon_j, \mbox{ where } \epsilon_j\!\sim\!\mbox{N}(0,\sigma_j^2), \label{eqn:colreg}\\
& \alpha_j=\mu_j-\Sigma_{j,-j}\Sigma_{-j,-j}^{-1}\mu_{-j}, \bs{\theta}_j\!=\!\Sigma_{-j,-j}^{-1}\Sigma_{-j,j}, \mbox{ and } \sigma_j^2=\Sigma_{j,j}\!-\!\Sigma_{j,-j}\Sigma_{-j,-j}^{-1}\Sigma_{-j,j} \notag
\end{align}
The intercepts  $\alpha_j$ ($j=1,\ldots, p$) can be set at 0 with centered $\X$.   Let $\Omega=\Sigma^{-1}$ be the precision matrix and $\omega_{jk}$ are the $[j,k]$-th entry in $\Omega$; then $\omega_{jj}=\sigma_j^{-2}$ and $\theta_{jk}=-\omega^{-1}_{jj}\omega_{jk}$ for $k\ne j$ \citep{learning631}, implying that $\theta_{kj}=0$ ($k\ne j$) is equivalent to $\omega_{kj}=0$.  Running $p$ regressions separately in Eqn (\ref{eqn:colreg}), with or without regularization on $\bs{\theta}$, does not lead to a symmetric $\Omega$ estimate nor does it guarantee its positive definiteness. If the main goal is to determine the existence of an edge between nodes $j$ and $k$, there are two common practices leading to a null edge between nodes $j$ and $k$ \citep{Meinshausen2006}: the intersection  rule  $\{\hat{\theta}_{jk}=0\}\cap\{\hat{{\theta}}_{kj} =0\}$ and the union rule $\{\hat{\theta}_{jk} =0\}\cup\{\hat{\theta}_{kj} =0\}$, with the latter resulting in less edges.

PANDA  regularizes the estimation of $\bs{\theta}_j$ with iterative injection of Gaussian noises drawn from the NGDs. During an iteration, in the regression with outcome node $X_j$, PANDA augments centered observed data in node $j$ with  $\e_{jj}=(e_{1,jj},\ldots,e_{n_e,jj})^T=(0,\ldots,0)$, and those in node $k$ ($k\ne j$) with  $\e_{jk}=(e_{1,jk},\ldots,e_{n_e,jk})^T$ drawn from a NGD in Eqns (\ref{eqn:bridge}) to (\ref{eqn:scad}). $n_e$, the size of augmented noisy data, should be large enough so that $n+n_e>p$ and $\bs{\theta}_{j}$ can be estimated with OLS by running the regression model in Eqn (\ref{eqn:colreg}) on the augmented data.

Proposition \ref{prop:regularization} establishes that PANDA, in expectation over the distribution of the injected noise, minimizes the overall  penalized SSE in the $p$ linear regression models with a penalty term on $\Theta=\{\bs{\theta}_{j}\}$ for $j=1,\ldots,p$. In other words, PANDA achieves the same global optimum as in \citet{Yuan2010} by iteratively solving the OLS of $\Theta$ until convergence. The proof of Proposition \ref{prop:regularization} is given in Appendix \ref{app:expectedloss}.   
\begin{pro}[\textbf{regularization effect of PANDA-NS for  GGM}]\label{prop:regularization}
The loss function given the original data $\x$ is the overall sum of squared errors (SSE) $l(\Theta|\x)\!=\!\sum_{j=1}^{p}\!\sum_{i=1}^{n}\!\left(x_{ij}\!\!-\!\!\sum_{k\ne j}x_{ik}\theta_{jk}\right)^2$, and the loss function based on the augmented data $\tilde{\x}$ is $l_p(\Theta|\tilde{\x})=l_p(\Theta|\x,\e)=\\
\sum_{i=1}^{n+n_e}\sum_{j=1}^{p}\!\left(\tilde{x}_{ij}\!-\!\sum_{k\ne j}\tilde{x}_{ik}\theta_{jk}\!\right)^2\!$. The  expectation of $l_p(\Theta|\x,\e)$ over the distribution of noise $\e$ is
\begin{align}\label{eqn:Elp}
\E_{\e}(l_p(\Theta|\x,\e))&=\textstyle l(\Theta|\x)+ P(\Theta).
\end{align}
\end{pro}
The penalty term $P(\Theta)$ takes different forms for different NGDs. Specifically,   $P(\Theta)=$
\begin{itemize}
\item $(\lambda n_e)\sum_{j=1}^{p}\sum_{j\neq k}|\theta_{jk}|^{2-\gamma}$ when $e_{jk}\sim N\left(0,\lambda|\theta_{jk}|^{-\gamma}\right)$, resulting in a bridge-type penalty (the lasso and ridge-type penalties are special cases at $\gamma=1$ and $\gamma=0$, respectively).
\item $(\lambda n_e)\sum_{j=1}^{p}\sum_{j\neq k}|\theta_{jk}|+(\sigma^2 n_e)\sum_{j=1}^{p}\sum_{j\neq k}\theta_{jk}^2$ when $e_{jk}\sim N\left(0,\lambda|\theta_{jk}|^{-1}+\sigma^2\right)$,  resulting in a elastic net-type penalty.
\item $(\lambda n_e)\sum_{j=1}^{p}\sum_{j\neq k}|\theta_{jk}||\hat{\theta}_{jk}|^{-\gamma}$ when $e_{jk}\sim N\left(0, \lambda |\theta_{jk}|^{-1}|\hat{\theta}_{jk}|^{-\gamma}\right)$, where $\hat{\theta}_{jk}$ is a\\ $\sqrt{n}$-consistent estimator of $\theta_{jk}$, resulting in an adaptive-lasso-type penalty.
\item  $\!\sum_{j=1}^{p}\!\sum_{j\neq k}\!\left(\!n_e\lambda|\theta_{jk}|1_{(0,\lambda n_e)}(|\theta_{jk}|)\!+\!\frac{1}{2(a\!-\!1)}\!\left(2a\lambda n_e| \theta_{jk}|\!-\!(\lambda n_e)^2\!-\!\theta_{jk}^2 \right)\!1_{[\lambda n_e,a\lambda n_e]} (|\theta_{jk}|)\!+\right.$\\
$\left.\frac{a\!+\!1}{2}(\lambda n_e)^2 1_{(a\lambda n_e,\infty)}(|\theta_{jk}|)\!\right)$ when $e_{jk}\sim N\!\left(\!0,\frac{\lambda}{|\theta_{jk}|}1_{(0,\lambda n_e)}(|\theta_{jk}|)\!+\frac{(a+1)\lambda^2n_e}{2\theta_{jk}^2} 1_{(a\lambda n_e,\infty)}(|\theta_{jk}|)+\right.$\\
$\left.\frac{1}{(a-1)}\! \left(\! \frac{a\lambda}{| \theta_{jk}|}\!-\!\frac{\lambda^2n_e}{2\theta_{jk}^2}-\!\frac{1}{2} \right) 1_{[\lambda n_e,a\lambda n_e]}(|\theta_{jk}|)\!\right)$ for $a>2$, resulting in a SCAD-type penalty.
\item $(\lambda n_e)\sum_{l=1}^{g}\sqrt{p_l}||\bs\theta_l||_2$ when $e_{lk}\sim N\left( 0, \frac{\lambda\sqrt{p_l}}{||\bs\theta_l||_2} \right)$, where $\bs\theta_l=\{\theta_{l1},\ldots,\theta_{lp_l} \}, l=1,\ldots,g$ is the index for the $g$ groups, resulting in a group-lasso-type penalty.
\end{itemize}

Algorithm \ref{alg:NSGGM} lists the computational steps of PANDA for constructing GGM, along with some remarks on setting some algorithmic parameters and convergence criterion (Remarks \ref{rem:T} to \ref{rem:nonconvex}).
\begin{algorithm}[!htb]
\caption{PANDA-NS for GGM}\label{alg:NSGGM}
\begin{algorithmic}[1]
\small
\State \textbf{Pre-processing}:  standardize the observed data $\x$
\State \textbf{Input}
\begin{itemize}[leftmargin=0.18in]\setlength\itemsep{-1pt}
\item Initial parameter estimates $\bar{\bs{\theta}}_j^{(0)}$ for $j=1,\ldots,p$.
\item A NGD from Eqns (\ref{eqn:bridge}) to (\ref{eqn:scad}) and the associated tuning parameters, the maximum iteration $T$ (Remark \ref{rem:T}),  noisy data size $n_e$ (Remark \ref{rem:nem}), moving average (MA) window width $m$  (Remark \ref{rem:nem}),  thresholds $\tau_0$ (Remark \ref{rem:tau0r}), banked parameter estimates after convergence $r$ (Remark \ref{rem:tau0r}).
\end{itemize}
\State $t\leftarrow 0$; convergence $\leftarrow 0$
\State \textbf{WHILE} $t<T$ \textbf{AND} convergence $= 0$
\State \hspace{0.3cm} $t\leftarrow t+1$
\State \textbf{\hspace{0.3cm} FOR} $j = 1:p$
\begin{enumerate}[leftmargin=0.5in]\setlength\itemsep{-1pt}
\item[a)] Generate noisy data $\e_j$ from the NGD with  $\bar{\bs{\theta}}_j^{(t-1)}$ plugged in the variance term of the NGD.
\item[b)] Obtain augmented data $\tilde{\x}_{j}$ by row-combining $(\x_j,\x_{-j})$ and  $(\mathbf{0},\e_j)$.
\item[c)] Obtain OLS estimate $\hat{\bs{\theta}}_{j}^{(t)}$ in the regression of $\tilde{\x}_{j}$ on $\tilde{\x}_{-j}$
\item[d)] If $t>m$, calculate  $\bar{\bs{\theta}}^{(t)}_j=m^{-1}\sum_{l=t-m+1}^t \hat{\bs{\theta}}_{j}^{(l)}$; otherwise $\bar{\bs{\theta}}^{(t)}_j=\hat{\bs{\theta}}^{(t)}_j$. Calculate SSE$^{(t)}_{j}$ on the original data at $\bar{\bs{\theta}}^{(t)}_j$.
\end{enumerate}
\State \hspace{0.45cm} \textbf{END FOR}
\State Calculate the loss function $\bar{l}^{(t)}=m^{-1}\!\sum_{l=t-m+1}^t \sum_{j=1}^p \mbox{SSE}^{(l)}_{j}$ and apply one of the convergence criteria listed in Remark \ref{rem:converge} to $\bar{l}^{(t)}$.  convergence $\leftarrow 1 $ if the convergence is reached.
\State \textbf{END WHILE}
\State Run lines 5 to 7 for another $r$ iterations after convergence, record $\bar{\bs\theta}^{(l)}_j$ for $l=t+1,\ldots,t+r$, and calculate the degrees of freedom $\nu_j^{(t)}=\mbox{trace}(\x_j(\tilde{\x}'_j\tilde{\x}_j)^{-1}\x'_j)$, and mean squared error (MSE) $\hat{\sigma}_{j}^{2(l)}=$ SSE$^{(t)}_{j}/(n-\nu_j^{(l)})$ for $j=1,\ldots, p$. Let $\bar{\bs{\theta}}_{jk}=(\bar{\theta}_{jk}^{(t+1)},\ldots,\bar{\theta}_{jk}^{(t+r)})$ for $j\ne k=1,\ldots,p$.
\State Set $\hat{\theta}_{jk}=\hat{\theta}_{kj}=0$
if $\left(\big|\max\{\bar{\bs{\theta}}_{jk}\}\cdot\min\{\bar{\bs{\theta}}_{jk}\}\big|<\tau_0\right) \cap \left(\max\{\bar{\bs{\theta}}_{jk}\}\cdot\min\{\bar{\bs{\theta}}_{jk}\}<0\right)$ or
$\left(\big|\max\{\bar{\bs{\theta}}_{kj}\}\!\cdot\!\min\{\bar{\bs{\theta}}_{kj}\}\big|<\tau_0\right) \cap \left(\max\{\bar{\bs{\theta}}_{kj}\}\!\cdot\!\min\{\bar{\bs{\theta}}_{kj}\}<0\right)$; and $\hat{\theta}_{jk}\!=\!\hat{\theta}_{kj}=\min\left\{\bar{\theta}_{jk}^{(t+r)},\bar{\theta}_{kj}^{(t+r)}\right\}$ otherwise.
\State To estimate $\Omega$, set $\hat{\omega}_{jj}=\hat{\sigma}_{j}^{-2(t+r)}$ and $\hat{\Omega}_{-j,j}=-\hat{\bs{\theta}}_{j}\hat{\omega}_{jj}$, where $\hat{\bs{\theta}}_{j}=\{\hat{\theta}_{jk}\}$ for $k\ne j$.
\State \textbf{Output}: $\hat{\bs{\theta}}$ and $\hat{\Omega}$.
\end{algorithmic}
\end{algorithm}
\normalsize
\begin{rem}[\textbf{convergence criterion}]\label{rem:converge}
We provide three choices to evaluate the convergence of the PANDA algorithm: 1) eyeball the trace plots of  $\bar{l}^{(t)}$, which is the most straightforward and often sufficient and effective; 2) use a cutoff value, say $\tau$ on the absolute percentage change on $\bar{l}^{(t)}$ from two consecutive iterations: if $|\bar{l}^{(t+1)}-\bar{l}^{(t)}|/\bar{l}^{(t)}<\tau$, then we may declare convergence; 3) apply a formal statistical test on  $\bar{l}^{(t)}$, the details of which is provided in Section \ref{sec:convergence}. Note that due to the randomness of the augmented noises from iteration to iteration, there is always some fluctuation around  $\bar{l}^{(t)}$ for finite $m$ and $n_e$. It is important to keep this in mind when evaluating convergence. For example, in the second criterion,  $\tau$ is expected to be small upon convergence, but being arbitrarily close to 0 would be difficult to achieve with finite $m$ or $n_e$. In the empirical studies in Sections \ref{sec:simulation} and \ref{sec:case}, $\tau$ was on the order of $O(10^{-2})$ upon convergence.
\end{rem}

\begin{rem}[\textbf{maximum iteration  $T$}]\label{rem:T}
$T$ should be set at a number large enough so to allow the algorithm to reach  convergence criterion in a reasonable time period. With a large $n_e$, we expect the algorithm to converge fast. For example, in the empirical studies in Sections \ref{sec:simulation} and \ref{sec:case}, convergence is achieved with $T\le20$ for PANDA-NS. 
\end{rem}


\begin{rem}[\textbf{choice of $n_e$ and $m$}]\label{rem:nem}
The expected regularization in Proposition \ref{prop:regularization}  can be realized  either by letting  $m\rightarrow \infty$ as  in $\sum_{j=1}^{p}\lim_{m\rightarrow\infty} m^{-1}\sum_{t=1}^m\!
\sum_{i=1}^{n_e}\!\left(e^{(t)}_{ijj}\!-\!\sum_{k\ne j}e^{(t)}_{ijk}\theta_{jk}\right)^2$, or  by letting  $n_e\rightarrow\infty$ as in
$\textstyle n_e\sum_{j=1}^{p}\lim_{n_e\rightarrow\infty} n_e^{-1}\!
\sum_{i=1}^{n_e}\left(e^{(t)}_{ijj}\!-\!\sum_{k\ne j}e^{(t)}_{ijk}\theta_{jk}\right)^2$ under the constraint $n_e\mbox{V}(e_{ijk})=O(1)$ for a given $\theta_{jk}$. The constraint $n_e\mbox{V}(e_{jk})=O(1)$ guarantees that injected noise $\e$ does not over-regularize or trump the information about $\Theta$ contained in the observed data $\x$ even when $n_e$ is large. For example, $\mbox{V}(e_{jk})=\lambda|\theta_{jk}|^{-1}$, for the lasso-type noise, and $n_e\lambda$ would be treated together as one tuning parameter. In practice, we can set either $m$ or $n_e$ at a large number to achieve the regularization effect. Our empirical results suggest the algorithm seems to converge faster and the loss function experiences less fluctuation by using a large $n_e$ ($m$ can be as small as 1 or 2) than using a large $m$.  Regarding what specific value to use on $n_e$, the only requirement is $n+n_e>p$ so that an unique OLS can be obtained from each regression in each iteration; but a large $n_e$ would need less iterations to converge. Regarding the choice of $m$, it more or less depends on $n_e$; if a large $n_e$ still results in noticeable fluctuation around  $\hat{\Theta}$, then a large $m$ can be used to speed up the convergence on $\hat{\Theta}$. There are also other considerations on the choices of $m$ and $n_e$ in non-Gaussian UGMs and when using PANDA to obtain inferences on parameters, which are discussed in Sections \ref{sec:nongaussian1}) and \ref{sec:asymp.dist}), respectively.
\end{rem}

\begin{rem}[\textbf{hard thresholding $\tau_0$ and choice of $r$}]\label{rem:tau0r}
The hard thresholding $\tau_0$ is necessary  as well as justified. It is needed for setting non-significant edges at 0 because,  though the estimates of the zero-valued $\theta_{jk}$ can get arbitrarily  close,  the exact 0 estimate cannot be achieved computationally in PANDA.  The hard thresholding is justified because of  the estimation and selection consistency property of PANDA established in Section \ref{sec:consist}. 
In addition, after the convergence of the PANDA algorithm, there is still mild fluctuation around the parameter estimates, especially when  $n_e$ or $m$ are not large. We would need a sequence of estimates on $\theta_{jk}$ to average out the random fluctuation; and we refer to this sequence as the banked estimates, the length of which is $r$. In the empirical studies we have conducted, $r=O(10^2)$ is sufficient.
\end{rem}

\begin{rem}[\textbf{non-convex targeted regularizers}]\label{rem:nonconvex}
PANDA optimizes a convex objective function in each iteration in the regression framework once the NA step is completed even when the targeted regularizer itself is non-convex, such as the SCAD. As such, PANDA will not run into computational difficulties as experienced by non-convex optimization.  That said, the final solutions for parameter estimates will depend highly on the staring values of the parameter -- different starting value could lead to different local optima.
\end{rem}


\subsubsection{Connection between PANDA-NS and weighted ridge regression for GGM}
Algorithm \ref{alg:NSGGM} shows the OLS estimator is obtained from the noise-augmented data in each iteration. Corollary \ref{cor:wrr} states that this OLS estimator is also a \emph{weighted ridge estimator}. Compared to the regular ridge estimator, where the same constant $\lambda$ is used for all the diagonal elements of $\x_{j,-j}'\x_{j,-j}$, different constants are used for different diagonal elements in the weighted ridge estimator.
\begin{cor}[\textbf{PANDA and weighted ridge regression}]\label{cor:wrr}
The OLS estimator from the regression with outcome node $X_j$ in PANDA on the noise augmented data is equivalent to the weighted ridge estimator $\hat{\bs{\theta}}_{j}
\!\!=\!\!\left(\x_{j,-j}'\x_{j,-j}\!+\!\e_{j,-j}^T\e_{j,-j}\right)^{-1}\!\!\x_{j,-j}{\x}_{j}$.
\end{cor}
The proof is straightforward. Let $\tilde{\x}=(\x,\e)^T$. The OLS estimator on the augmented data is $\hat{\bs{\theta}}_{j}=(\tilde{\x}_{j,-j}^T\tilde{\x}_{j,-j})^{-1}\tilde{\x}^T_{j,-j}(\x_j,\mathbf{0})=(\x_{j,-j}^T\x_{j,-j}+\e_{j-j}^T\e_{j,-j})^{-1}{\x}_{j,-j}{\x}_{j}$, leading to Corollary \ref{cor:wrr}. When $n_e\rightarrow \infty$,  $\e_{j-j}^T\e_{j,-j}\rightarrow n_e\mbox{V}(\e_{j,-j})$. For example, if $e_{j,k}\sim\mbox{N}(0,\lambda|\theta|^{-\gamma}_{jk})$ ($k\ne j$), then  $n_e\mbox{V}(\e_{j,-j})=(n_e\lambda) \mbox{diag}(|\theta|^{-\gamma}_{jk})$. Therefore, the regularization effect varies by the magnitude of $\theta_{jk}$ -- the closer $\theta_{jk}$ is to 0, the more regularization (shrinkage to 0) there is on the estimate $\hat{\theta}_{jk}$.

\subsubsection{PANDA for UGM with non-Gaussian nodes}\label{sec:nongaussian1}
When the conditional distribution of every node given the other nodes follow an exponential family, then regardless whether the nodes are of the same or mixed types, PANDA-NS can regularize the graph construction via running GLM with the canonical link functions. 
Proposition \ref{prop:glmregularization} states the expected regularization effects of PANDA in UGM. The proof is given in Appendix \ref{app:glmregularization}.

\begin{pro}[\textbf{Regularization effects of PANDA in UGMs}]\label{prop:glmregularization}
Let the loss function given the observed data $\x$  be $l(\Theta|\x)\!=\!-\!\sum_{j=1}^{p}\! \sum_{i=1}^n \!\left\{\!h_j(x_{ij})\!+\!\left(\theta_{j0}\!+\!\!\sum_{k\ne j}\!\theta_{jk}x_{ik}\right)x_{ij}\!-\!B_j(\theta_{j0}\!+\!\!\sum_{k\ne j}\!\theta_{jk}x_{ik})\!\right\}$ (summation of $p$ negative log-likelihood functions), and the loss function given with the noise augmented data $\tilde{\x}=(\x,\e)$ be
\begin{equation}\label{eqn:GLMloss}
\textstyle l_p(\Theta|\tilde{\x})\!=\! -\sum_{j=1}^{p}\left\{\sum_{i=1}^{n+n_e} \left(h_j(\tilde{x}_{ij})\!+\!\left(\theta_{j0}+\sum_{k\ne j}\theta_{jk}\tilde{x}_{ik}\right)\tilde{x}_{ij}\right)
\!-\!B_j\left(\theta_{j0}+\sum_{k\ne j}\theta_{jk}x_{ik}\right)\right\}.
\end{equation}
Apply the Taylor expansion to $l_p$ around $\sum_{k\ne j}\theta_{jk}x_{ik}=0$ and take expectation over the distribution of $\e$, we have
\begin{align}
&\E_{\e}(l_p(\Theta|\x,\e))= l(\Theta|\x)+P(\Theta), \mbox{ where}\notag \\
&P(\Theta)\!=\!\textstyle
n_e\sum_{j=1}^{p}\!\!\left(C_{1j}\!\sum_{k\ne j}\theta^2_{jk}\mbox{V}(e_{jk})\right) \!+\!O\!\left(\!n_e\!\sum_{j=1}^{p}\!\sum_{k\ne j}\!\left(\theta_{jk}^4\V^2(e_{jk})\!\right)\!\right)\!+\!C,\label{eqn:glmregularization}
\end{align}
where $C_{1j}=2^{-1}B''_j(\theta_{j0})$ and $C=\sum_{j=1}^{p}\sum_{i=1}^{n_e} \left(h_j(e_{ijj})+e_{ijj}\theta_{j0}\right)+B_j(\theta_{j0})$ are constants independent of $\Theta=\{\theta_{jk}\}$.
\end{pro}

The actual form $P(\Theta)$ in Eqn \ref{eqn:glmregularization} depends on the node type of $X_j$ and the NGD from which $\e$ is drawn. Table \ref{tab:glm} lists some examples on $P(\Theta)$ if the lasso-type NSG  is used ($\gamma=1$ in Eqn (\ref{eqn:bridge})) for graphs with the same type of nodes.  For examples, if all nodes follow a Bernoulli distribution given all the other nodes, then the graph is called Bernoulli graph model (BGM); similarly for EGM (Exponential), PGM (Poisson), and NBGM (Negative Binomial).
\begin{table}[!htp]
\begin{center}
\begin{tabular}{lll}
\hline
graph   & $P(\Theta)$\\
\hline
GGM & $\lambda n_e\sum_{j=1}^{p}\sum_{k\ne j}|\theta_{jk}|$\\
BGM& $\frac{\lambda n_e}{2}\sum_{j=1}^{p}\frac{\exp(\theta_{j0})}{(1+\exp(\theta_{j0}))^2}\sum_{k\ne j}|\theta_{jk}|+O(\lambda^2 n _e ||\Theta||_2^2)+C$\\
EGM &  $\frac{\lambda n_e}{2}\sum_{j=1}^{p}\exp(\theta_{j0})\sum_{k\ne j}|\theta_{jk}|+O(\lambda^2 n _e ||\Theta||_2^2)+C$\\
PGM &  $\frac{\lambda n_e}{2}\sum_{j=1}^{p}\exp(\theta_{j0})\sum_{k\ne j}|\theta_{jk}|+O(\lambda^2 n _e ||\Theta||_2^2)+C$\\
NBGM & $\frac{\lambda n_e}{2}\sum_{j=1}^{p}\frac{r_j\exp(\theta_{j0})}{(r_j+\exp(\theta_{j0}))}\sum_{k\ne j}|\theta_{jk}|+O(\lambda^2 n _e ||\Theta||_2^2)+C$ ($r$ is the \# of failures)\\
\hline
\end{tabular}
\end{center}
\caption{Expected penalty term in PANDA with lasso-type noise augmentation for various graphs} \label{tab:glm}
\end{table}

Similar to Proposition \ref{prop:regularization}, the expectation of $l_p(\Theta|\x,\e)$ in Proposition \ref{prop:glmregularization} can be achieved by letting $m\rightarrow\infty$ as in
$\lim_{m\rightarrow\infty} m^{-1}\sum_{t=1}^m \sum_{j=1}^{p}\sum_{i=1}^{n_e} l(\bs\theta_j|\X_i, \e^{(t)}_{i,-j})$,
or, suggested by Eqn (\ref{eqn:glmregularization}), by  letting $n_e\rightarrow\infty$ with the constraint $n_e\mbox{V}(e_{ijk})\!=\!O(1)$; that is,
$n_e\sum_{j=1}^p
\left(C_{1j}\sum_{k\ne j}\theta_{jk}^2\mbox{V}(e_{ijk})\right)=
n_e\sum_{j=1}^{p}C_{1j}\sum_{k\ne j}\!\left(\!\theta^2_{jk}\lim_{n_e\rightarrow\infty} n_e^{-1}\!
\sum_{i=1}^{n_e}\e^2_{ijk})\!\right)$.
Between $m\rightarrow\infty$  and $n_e\rightarrow\infty$, the latter would be preferable in  that the higher-order term $O\left(\sum_{j=1}^{p}\!\sum_{k\ne j}\!\left(\theta_{jk}^4n_e\V^2(e_{jk}))\!\right)\!\right)\!\rightarrow\! 0$ in Eqn (\ref{eqn:glmregularization}), meaning the targeted regularizer can be achieved arbitrarily well.  $m\rightarrow\infty$ with $n_e$ fixed has no effect on the higher-order term, which can only reply on small $V(e_{jk})$ or small $\theta^2_{jk}$, to become ignorable relative to the lower-order term $n_e\sum_{j=1}^{p}\!\!\left(C_{1j}\!\sum_{k\ne j}\theta^2_{jk}\mbox{V}(e_{jk})\right)$, the targeted regularizer. In other words, the higher-order term, which is a function of $\Theta$, might bring additional regularization to $\Theta$ on top of the targeted regularization.

\begin{figure}[!htb]
\begin{minipage}{1\textwidth}
\includegraphics[width=0.245\linewidth]{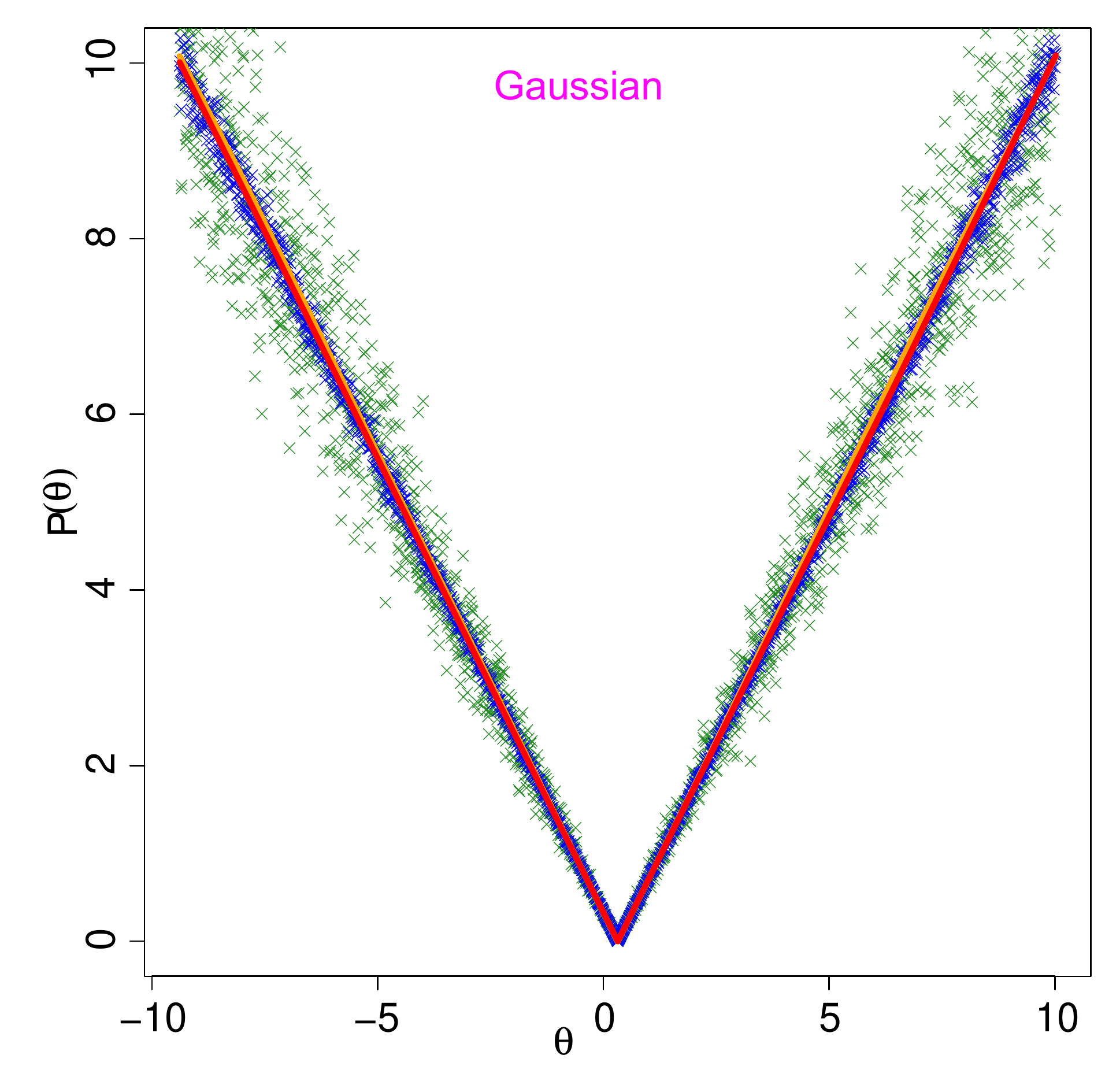}
\includegraphics[width=0.245\linewidth]{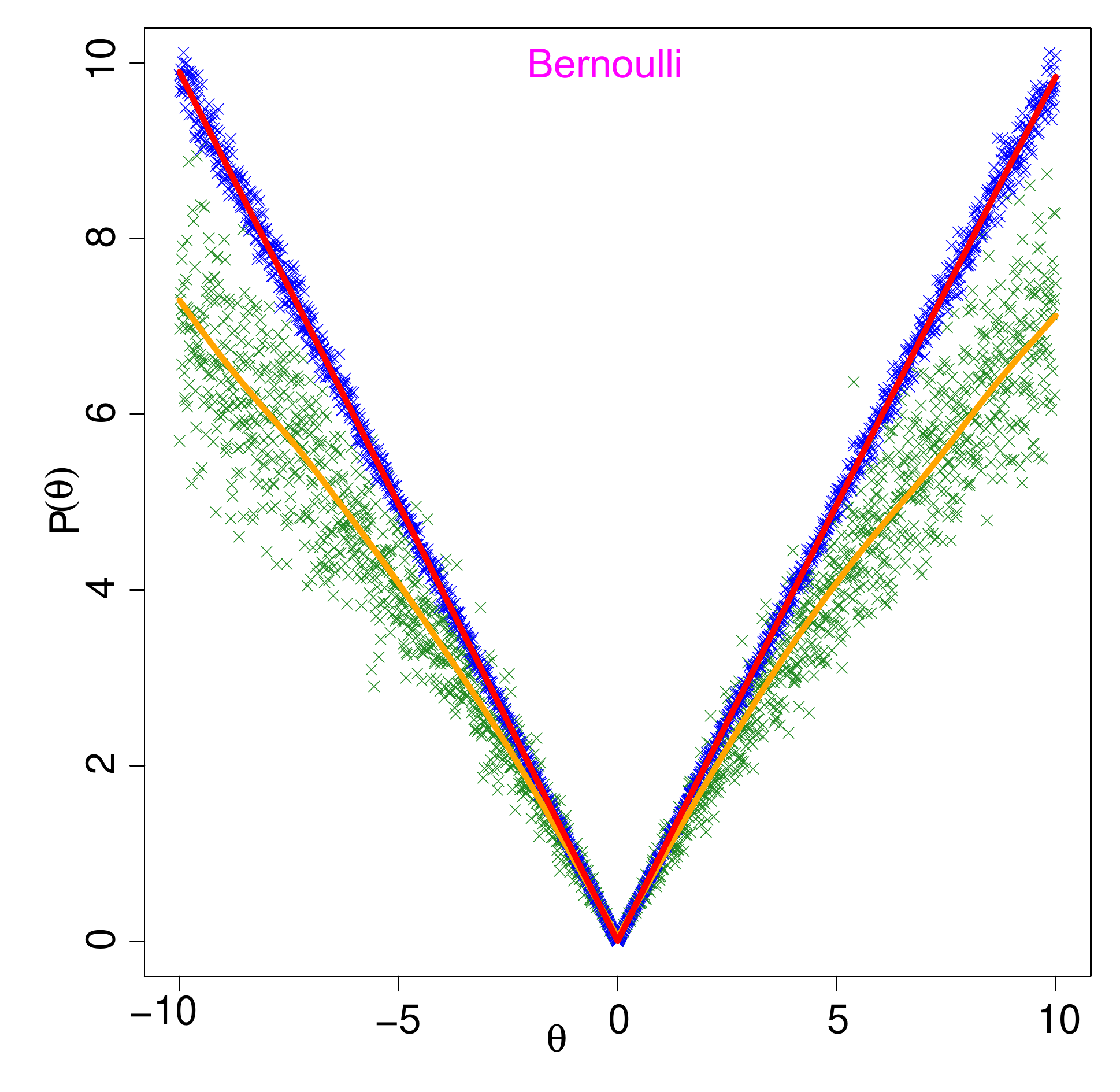}
\includegraphics[width=0.248\linewidth]{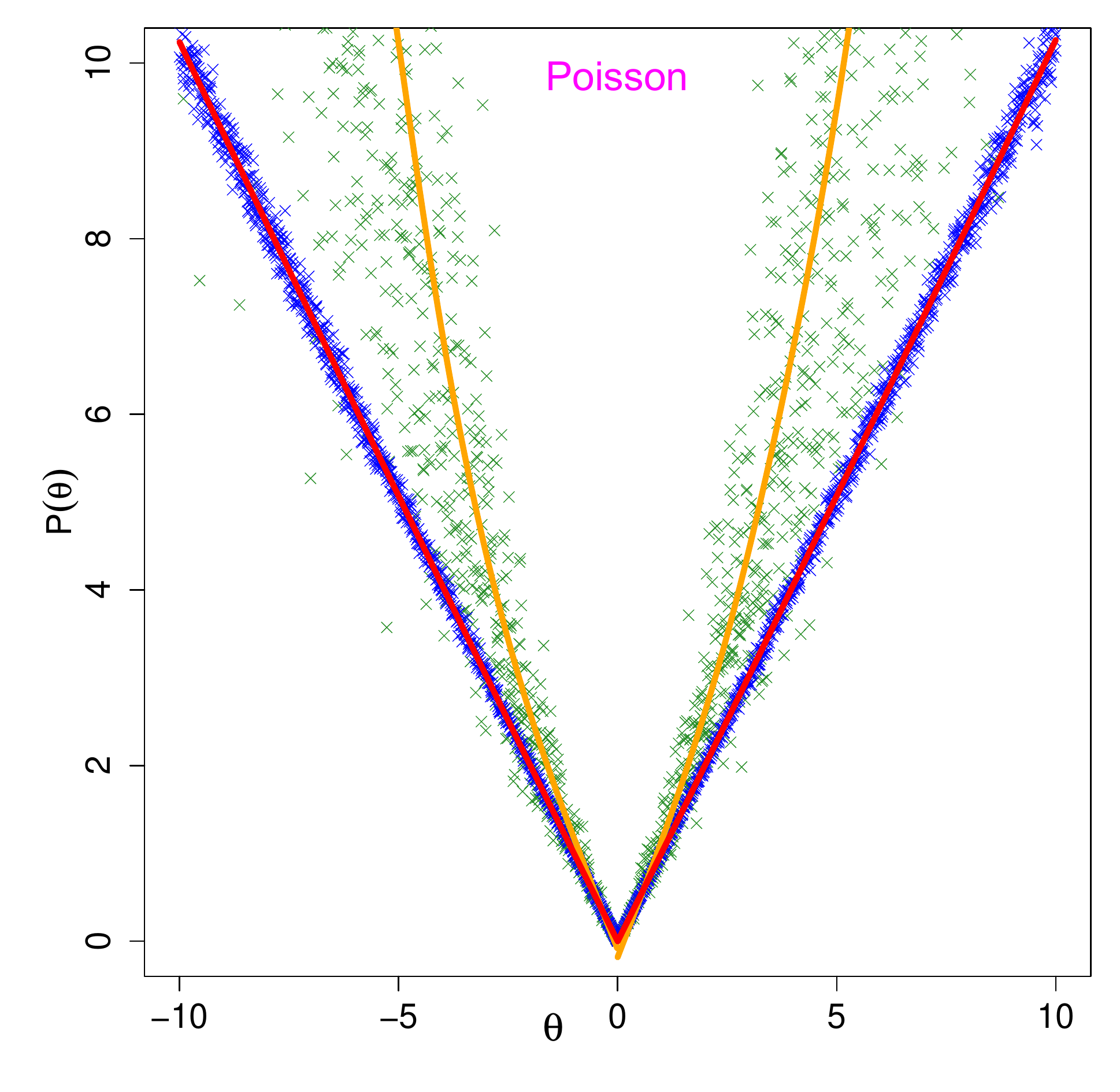}
\includegraphics[width=0.245\linewidth]{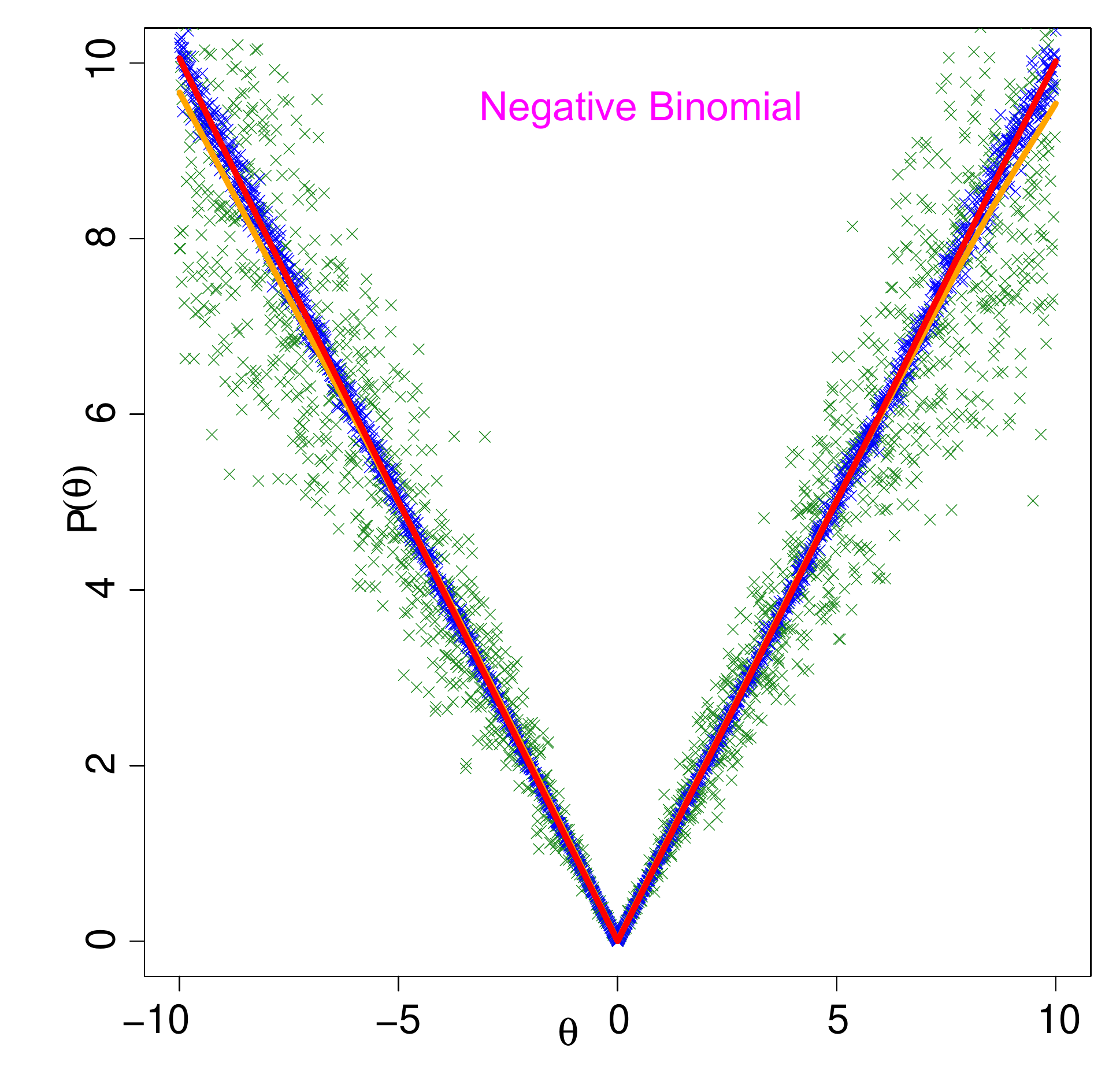}
\end{minipage}
\caption{Realized regularization by PANDA in different graphs for the targeted regularization $P(\theta)=|\theta|$. Red and orange lines are the realized penalty at $(n_e\rightarrow\infty) \cap (\lambda n_e=1)$ with $m=50$, and  at $m\rightarrow\infty$ with $n_e=5,\lambda=1/5$, respectively; blue and green dots are their respective empirical penalties by setting $(n_e=100, m=50)$ and $(n_e=5, m=50)$.}
\label{fig:regeffect}
\end{figure}

To illustrate the differences between the regularization effects between letting $n_e\rightarrow\infty$ and $m\rightarrow\infty$, we display  in Figure \ref{fig:regeffect}  the relationships between the realized $P(\theta)$ by PANDA and $\theta$ for several graph types, along with their empirical versions when the lasso-typed augmented noises are used (the regularization effect in EGM looks very similar to the PGM and the results from EGM are not provided).  The targeted regularizer is lasso ($P(\theta)=|\theta|$). With $n_e\rightarrow\infty$ ($\lambda n_e=1$ fixed at 1, and $m=50$), the realized penalty (red lines) is identical to lasso in all four graphs; and its empirical version (the blue dots) at $n_e=100$ is very close to the analytic form except for some very mild fluctuation. The realized regularization on $\theta$ with $m\rightarrow\infty$ while $n_e$ is small (orange lines) varies by graph. When $|\theta|$ is small, the distinction between $n_e\rightarrow\infty$  and $m\rightarrow\infty$ is minimal in four cases as the higher-order term is ignorable in each graph. As as $|\theta|$ increases, the regularization deviates from linearity (the target regularization) since the the higher-order residual term in Eqn (\ref{eqn:glmregularization}) becomes less ignorable. Specifically, the realized regularization is sub-linear for BGM through logistic regression and for NBGM through NB regression (though not obvious), and super-linear in PGM through Poisson regression (and EGM). The only exception is GGM through linear regression where the higher-order term is analytically 0.

\begin{figure}[!htb]
\begin{minipage}{0.45\textwidth}
\includegraphics[width=1\linewidth]{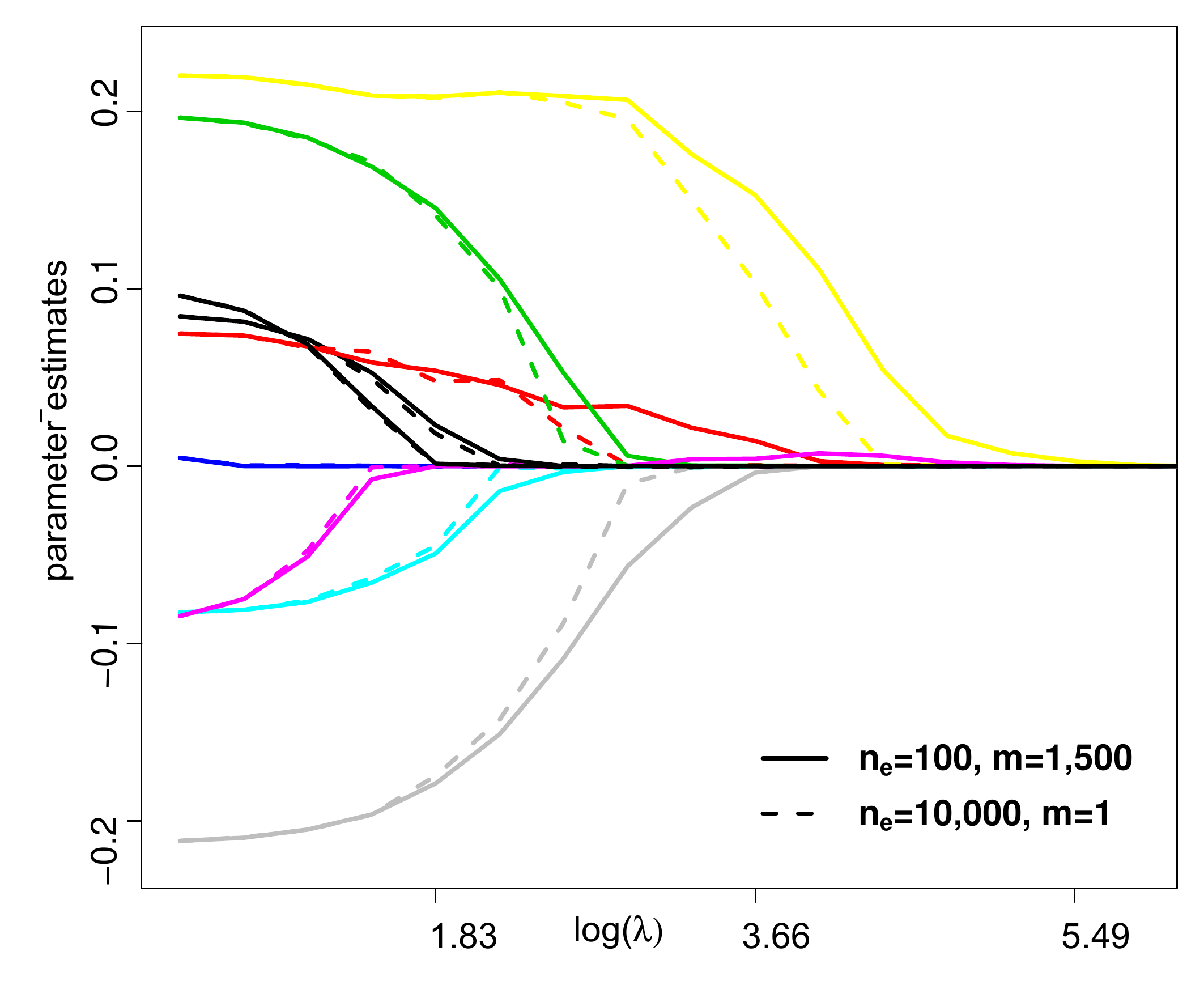}
\end{minipage}
\begin{minipage}{0.46\textwidth}
\includegraphics[width=1\textwidth]{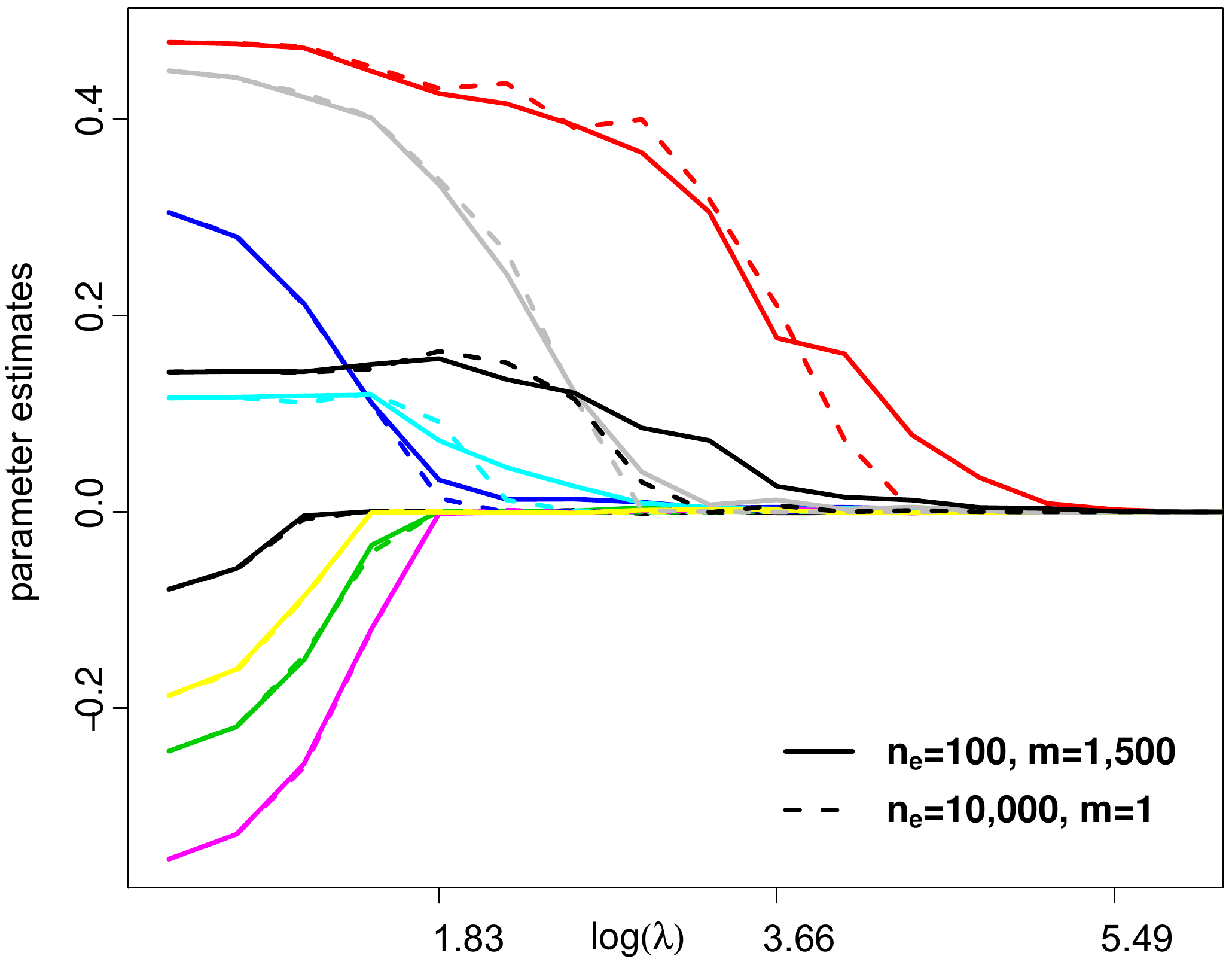}
\end{minipage}
\caption{Trajectories of estimates for zero-valued regression coefficients in linear regression (left) and Poisson regression (right) with different $\lambda$ when the lasso-type noise is used in PANDA}\label{fig:trace}
\end{figure}

In Figure \ref{fig:trace}, we show how the regularized parameter estimates obtained with large $n_e$ vs. with large $m$ change with $\lambda$ when the lasso-type noise is used in PANDA.  Specifically, we run PANDA in simulated data in linear regression 
and Poisson regression, respectively, with ($n_e=10^4, m=1$) and ($n_e=100, m=1.5\times 10^3$). In both cases, there are 30 predictor ($p=30$) and $n=100$. In the linear regression, the predictors were simulated from N$(0,1)$; in the Poisson regression, the predictors were simulated from Unif$(-0.3,0.5)$. Out of the 30 regression coefficients, 9 of them were set at 0, and the 21 nonzero coefficients ranged from 0.5 to 1. Under these settings, the trajectories of the regularized estimates for the 9 zero-valued parameters are similar for large $m$ and large $n$ in both regression; but large $m$ had a higher computation cost. 

In practice, when $n_e$ is large, setting $m=1$ in a PANDA algorithm is sufficient to achieve the expected regularization effect. On the other hand, a very large $n_e$ will slow the computation in each iteration. Therefore, we would recommend set $n_e$ at a somewhat large value to yield the expected regularization effect, and then set $m$ at a small value to average out the fluctuation around the estimated parameters.

Due to space limitation, we list the computational algorithm in PANDA for constructing UGM in Algorithm \ref{alg:NSUGM} in the Supplementary Materials. Most of the steps are similar to  Algorithm \ref{alg:NSGGM} for GGM, with a few  differences. First, there is no standardization of data; second, the loss function optimized in each iteration is the sum of the negative log-likelihood in Eqn (\ref{eqn:GLMloss}) across the nodes; third, MLE $\hat{\bs{\theta}}_{j}$ (not OLS) is calculated from regressing $\tilde{\x}_{j}$ on all other nodes $\tilde{\x}_{-j}$ for $j=1,\ldots,p$ in each iteration. The guidelines for choosing of the algorithmic parameters (e.g., $T,n_e,m,\tau_0,r$) and evaluating the convergence as laid out in Remarks \ref{rem:T} to \ref{rem:nonconvex} also apply to the UGM algorithm. 

\subsection{Other regularization for GGM via PANDA}\label{sec:otherGGM}
For GGM, given the connection between the graph structure and the precision matrix of the multivariate Gaussian distribution,  additional approaches have been proposed to construct a GGM. We list three of these approaches that can all be realized through  PANDA.

\subsubsection{PANDA-SPACE for hub nodes detection in GGM}\label{sec:SPACE}
The elements in the precision matrix $\Omega$ of a multivariate Gaussian distribution are related to the partial correlation coefficients  in linear regression. Specifically, the partial correlation between node $X_j$ and node $X_k$  is $\rho_{jk}\!=\!-\omega_{jk}/\sqrt{\omega_{jj}\omega_{kk}}=\beta_{jk}\sqrt{\sigma_{jj}/\sigma_{kk}}$ ( Lemma 1 in \citet{Jie2009}). 
SPACE (Sparse PArtial Correlation Estimation) is an approach to select nonzero partial correlation when $n<p$ \citep{Jie2009}. Non-zero  $\rho_{jk}$ implies non-zero $\omega_{jk}$ and an edge between nodes $j$ and $k$ in GGM. The biggest advantage of SPACE, compared to NS, is that not only does it identify edges, it is also efficient for identifying hub nodes. Corollary \ref{cor:space} shows that PANDA can realize SPACE by imposing a bridge-type penalty on $\rho_{jk}$. The data augmentation scheme is similar to Figure \ref{fig:panda1}. In each iteration,  PANDA runs $p$ linear regressions based on the noise-augmented data, obtain estimates for $\beta_{jk}, \sigma_{jj}$, and $\sigma_{kk}$, and calculates $\rho_{jk},\omega_{jj}$ and $\omega_{kk}$.
\begin{cor}[\textbf{PANDA-SPACE}]\label{cor:space}
Let $e_{jk}\!\!\overset{\text{ind}}{\sim}\!\!N\left(0,\lambda|\rho_{jk}|^{-\gamma}\omega_{jj}\omega_{kk}^{-1}\right), e_{jj}\equiv0, \Theta=\{\rho_{jk},\omega_{jj},\omega_{kk}\}$ for $j\ne k=1,\ldots,p$, and $l(\Theta|\x) =\!\sum_{i=1}^{n}\!\sum_{j=1}^{p}\!\left(\!x_{ijj}\!-\!\sum_{k\ne j}x_{ijk}\rho_{jk}\sqrt{\omega_{kk}/\omega_{jj}}\right)^2$. Then
\begin{align*}
l_p(\Theta|\x,\e)&= \textstyle l(\Theta|\x)+ \sum_{i=1}^{n_e}\sum_{j=1}^{p}\left(\sum_{k\ne j}e_{ijk}\rho_{jk}\sqrt{\omega_{kk}/\omega_{jj}}\right)^2, \\
\E(l_p(\Theta|\x,\e))&=\textstyle l(\Theta|\x)+\lambda n_e\sum_{j=1}^{p}\sum_{j\neq k}|\rho_{jk}|^{2-\gamma}.
\end{align*}
\end{cor}

\subsubsection{PANDA-CD for GGM}\label{sec:CD}
The Cholesky decomposition (CD) approach refers to estimating $\Omega$ through the LDL decomposition, a variant of the CD. Compared to the NS in Section \ref{sec:NSGGM}, the CD approach guarantees symmetry and positive definiteness of the estimated $\hat{\Omega}$. WLOG, let $\x_{n\times p}\sim N_p(\mathbf{0},\Omega)$, and the corresponding negative log-likelihood is
$l(\Omega|\x)\!=\!-n\log(|\Omega|)+\textstyle\frac{1}{2}\sum_{i=1}^{n}\!\x_i^T\Omega\x_i$.
There exists a unique LDL decomposition  $\Omega =L^TD^{-1}L$, such that  $|\Omega|=|D|^{-1}=\prod_{j=1}^{p}\sigma_{j}^{-2}$, where $D=\mbox{diag}(\sigma_{1}^2,\ldots,\sigma_{p}^2)$ and $L$ is a lower uni-triangular matrix with elements $-\theta_{jk}$ for $j>k$, 0 for $k<j$, and 1 for $j=k$. Therefore,
\begin{align}
l(\Omega|\x)= l(L,D|\x)& =\textstyle n\log|D|\!+\!\sum_{i=1}^{n}\!\x_i^TL^TD^{-1}L\x_i= n\log|D|\!+\!\sum_{i=1}^{n}\!(L\x_i)^TD^{-1}L\x_i\label{eqn:LDL1}\\
& =\textstyle n\!\sum_{j=1}^{p}\!\log\!\sigma_{j}^2\!+\!\sum_{i=1}^{n}\!\sum_{j=1}^{p}\!\sigma_{j}^{-2}\left(\!x_{ij}-\sum_{k=1}^{j-1}x_{ik}\theta_{jk}\!\right)^2.\label{eqn:LDL2}
 \end{align}

\citet{Huang2006} apply the $l_\gamma$  regularization $(\gamma>0)$ on $\theta_{jk}$ to the negative log-likelihood in Eqn (\ref{eqn:LDL2}) and minimize it by solving Eqns  (\ref{eqn:CDtheta}) and (\ref{eqn:CDsigma})  alternatively in an iterative manner.
 \begin{align}
 \hat{\bs{\theta}}_j&=\textstyle \arg\min\limits_{\bs\theta_{j}}\left\{\hat{\sigma}_j^{-2}\sum_{i=1}^{n}\left(\x_{ij}-\sum_{k=1}^{j-1}\x_{ik}\theta_{jk}\right)^2+\xi\sum_{k=1}^{j-1}|\theta_{jk}|^\gamma\right\},\label{eqn:CDtheta}\\
\hat{\sigma}_j^2&=\textstyle n^{-1}\sum_{i=1}^{n}\left(\x_{ij}-\sum_{k=1}^{j-1}\x_{ik}\hat{\bs{\theta}}_{jk}\right)^2.\label{eqn:CDsigma}
\end{align}
Optimization and regularization occur only on $\bs\theta_j$ in Eqn (\ref{eqn:CDtheta}), whereas Eqn (\ref{eqn:CDsigma}) can be calculated analytically once $\bs{\theta}_j$ is estimated. We show below how PANDA realizes the above framework.  Instead of solving the optimization problem in Eqn (\ref{eqn:CDtheta}), PANDA calculates the OLS of $\bs{\theta}_j$ from noise-augmented data. Specifically, let $\bs{\epsilon}=L\X$, then $\bs{\epsilon}\sim\mbox{N}(\mathbf{0},D)$ and  Eqn (\ref{eqn:LDL1}) can  be expressed as the summation of the likelihood functions from a series of linear  models
\begin{align}\label{eqn:CDreg}
X_{1}=\epsilon_{1}\mbox{ and }  X_{j}= \textstyle \sum_{k=1}^{j-1} X_{k}\theta_{jk}+ \epsilon_{j} \mbox{ for } j=2,\ldots,p;
\end{align}
\setlength{\belowcaptionskip}{-10pt}
\begin{wrapfigure}{r}{0.5\textwidth}
\centering\includegraphics[width=0.45\textwidth]{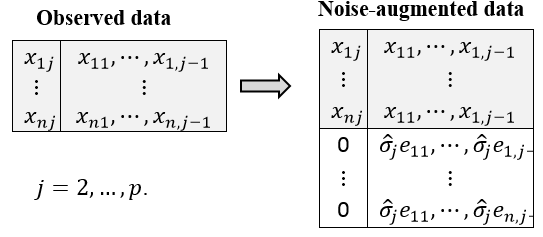}
\caption{A schematic of the data augmentation in PANDA-CD for GGM ($\sigma^2_j$ varies by iteration; see Algorithm \ref{alg:CD} in the supplementary materials)}\label{fig:pandaCD1}
\end{wrapfigure}
that is, the model on $X_1$ has only the known intercept term of 0 (on centered data) plus an error term, $X_2$ is regressed on $X_1$, $X_3$ is regressed on $(X_1,X_2)$, and so on.  PANDA  augments the observed data in the outcome node $X_j$ with 0 and those in each of the covariate nodes in Eqn (\ref{eqn:CDreg}) with $n_e$ noise terms sampled from a NGD (Eqns \ref{eqn:bridge} to \ref{eqn:scad}).  Figure \ref{fig:pandaCD1} depicts a schematic  of the data augmentation in PANDA-CD. Though the earlier regression model have less predictors and do not have the $n<p$ problem, $n_e$ and the tuning parameters in the NGD should be kept the same in every regression in Eqn (\ref{eqn:CDreg}), so to achieve the targeted regularization effect.

The steps of the PANDA-CD algorithm are listed in Algorithm \ref{alg:CD} in the supplementary materials.  Proposition \ref{prop:orderreg} establishes that the expected noised-augmented likelihood function over the distribution of  $\e$ drawn from the NGD in Eqn (\ref{eqn:bridge}) is  equivalent to the penalized likelihood function in Eqn (\ref{eqn:CDtheta}),  with turning parameter  $\lambda n_e$ (same role as $\xi$ in  Eqn (\ref{eqn:CDtheta})).   The proof of Proposition \ref{prop:orderreg} is given in Appendix \ref{app:cholesky}.  It is straightforward to extend Proposition \ref{prop:orderreg} to other types of noises by using any NGD from Eqns (\ref{eqn:en}) to (\ref{eqn:scad}), leading to other types of regularization  on $\bs{\theta}_{j}$.
\begin{pro}[\textbf{Regularization effects of PANDA-CD for GGM}]\label{prop:orderreg}
Let $l(L|\x)=\\ \sum_{j=1}^{p}\hat{\sigma}_j^{-2}\sum_{i=1}^{n}\left({x}_{ij}-\sum_{k=1}^{j-1}{x}_{ik}\theta_{jk}\right)^2$ be the loss function given the observed data $\x$, and  $l_p(L|\x,\e)\\ =
\sum_{j=1}^{p}\hat{\sigma}_j^{-2}\sum_{i=1}^{n}\left({x}_{ij}-\sum_{k=1}^{j-1}{x}_{ik}\theta_{jk}\right)^2+\sum_{j=1}^{p}\hat{\sigma}_j^{-2}\sum_{i=1}^{n_e}\left({e}_{ijj}-\sum_{k=1}^{j-1}{e}_{ijk}\theta_{jk}\right)^2$ be the loss function based on noise-augmented data. The expectation of $l_p(L|\x,\e)$ over the distribution of $\e$ drawn from the NGD in Eqn (\ref{eqn:bridge}) is
\begin{align}
\E_{\e}(l_p(L|\x,\e))=\textstyle l(L|\x)+\lambda n_e\sum_{j=1}^{p}\sum_{k=1}^{j-1}|\theta_{jk}|^{2-\gamma}\label{eqn:ElLDL}
\end{align}
\end{pro}

\subsubsection{PANDA-SCIO for GGM} \label{sec:scio}
The Sparse Columnwise Inverse Operator (SCIO) estimator \citep{weidong2015} of the precision matrix $\Omega$ of a GGM is realized by solving $p$ $l_1$-regularized \emph{quadratic optimization} problems:
\begin{equation}\label{eqn:qo}
\hat{\bs{\theta}}_{j}=\textstyle \arg\min\limits_{\bs{\bs{\theta}_{j}}\in R^p}\left\{l_j+\lambda\sum_{k\ne j}|{\theta}_{jk}|\right\}, \mbox{ where the objective function } l_j=\frac{1}{2}\bs{\theta}_{j}^t\hat{\Sigma}\bs{\theta}_{j}-\mathbf{1}_j\bs{\theta}_{j}
\end{equation}
for $j=1,\ldots,p$. $\bs{\theta}_{j}$ is the $j$-th column of $\Omega$, $\hat{\Sigma}=n^{-1}\x'\x$, $\mathbf{1}_j$ is a row binary vector of dimension $p$ with 1 at the $j^{th}$ entry and 0 otherwise, and $\lambda > 0$ is a tuning parameter. 
After $\bs{\theta}_{j}$ is estimated for $j=1,\ldots,p$, $\omega_{jk}$ can be estimated by $\hat{\omega}_{kj}=\min\{\hat{{\theta}}_{jk}, \hat{{\theta}}_{kj}\}$.

The PANDA technique can be used to obtain the SCIO estimator that only needs to take the inverses of a positive definitive matrices
without resorting to complicated optimization algorithms with constraints. Since the SCIO estimator  in Eqn (\ref{eqn:qo}) is defined with the $l_1$ regularization, we thus use the lasso-type  noise to illustrate PANDA-SCIO; but any type of noises from NGDs in Eqns (\ref{eqn:bridge}) to (\ref{eqn:group}) can also be applied in the SCIO framework. In brief, PANDA draws $e_{ijk}$  from N$\left(0, \lambda|{\theta}_{jk}|^{-1}\right)$ and sets $\e_{ijj}$ at 0 for $i=1,\ldots,n_e$. It then scales the observed data $\x$ to obtain $\z=\sqrt{(n+n_e)n^{-1}}\x$, and the augmented data $\e$ to obtain $\mathbf{d}=\sqrt{2(n+n_e)n_e^{-1}}\e$;
and calculates $\tilde{\Sigma}=(n+n_e)^{-1}\tilde{\x}^T\tilde{\x}$, where $\tilde{\x}=(\z,\mathbf{d})$. Plugging $\tilde{\Sigma}$ in the objective function in  Eqn (\ref{eqn:qo}), we have $\tilde{l}_j=\textstyle \frac{1}{2}\bs{\theta}_{j}^t\tilde{\Sigma}\bs{\theta}_{j}-\1_j\bs{\theta}_{j}$, the minimizer of which can be easily obtained analytically, which is $\hat{\bs{\theta}}_j=\tilde{\Sigma}^{-1}\1_j$. With the data augmentation, the inverse of $\tilde{\Sigma}$ always exists.  The computational steps of the PANDA-SCIO algorithm are given in Algorithm \ref{alg:scio} in the Supplementary Materials.

It is easily to prove that the expectation of the loss function $\tilde{l}_j$  over the distribution of $\e$ has the same regularization as the SCIO in Eqn (\ref{eqn:qo}). Specifically,
$l_j\!=\!\frac{1}{2}\bs{\theta}_{j}^T \left(\!\sum_{i=1}^n\!\z_i\z_i^T\!+\!\sum_{i=1}^{n_e}\!\mathbf{d}_{ij}\mathbf{d}_{ij}^T\right)
\bs{\theta}_{j}-\1_j\bs{\theta}_{j}
=\frac{1}{2}\bs{\theta}_{j}^t\textstyle
\left(\frac{1}{n}\sum_{i=1}^{n}\x_i\x_i^T +\frac{2}{n_e}\sum_{i=1}^{n_e}\e_{ij}\e_{ij}^t\right)\bs{\theta}_{j}-\mathbf{1}_j\bs{\theta}_{j}$
and $\E(l_j)=\frac{1}{2}\bs{\theta}_{j}^t\hat{\Sigma}\bs{\theta}_{j}-\1_j\bs{\theta}_{j}+\lambda\textstyle\sum_{k=1}^{p}\left|{\theta}_{jk}\right|.$

\subsubsection{PANDA-graphical ridge for GGM}\label{sec:graphical}
The PANDA technique can also be employed to regularize the off-diagonal elements in $\Omega$ simultaneously for GGM, instead of in a columnwise fashion as in PANDA- NS, SPACE, CD, and SCIO. Existing work on simultaneous regularization of $\omega_{jk}$ ($k\ne j$) includes the graphical lasso \citep{Friedman2008} and the graphical ridge \citep{rope2017}. The graphical lasso imposes the $l_1$ penalty $\sum_{k\ne j}|\omega_{jk}|$ while the graphical ridge imposes the $l_2$ penalty $\sum_{k\ne j}\omega_{jk}^2$. The $l_2$ penalty is used when achieving sparsity is not the main goal such as in principal component analysis or in prediction problems. 

PANDA starts with an initial value for $\Omega$ and draws $\e_i$ from $\mbox{N}_{(p)}(0,\lambda\Omega)$ for $i=1,\ldots,n_e$.  It then row-combines $\sqrt{n^{-1}(n+n_e)}\x$ and $\sqrt{n_e^{-1}(n+n_e)}\e$ to get the scaled augmented data $\tilde{\x}$ and calculates the MLE of $\Omega$, which is the inverse of the sample covariance matrix of $\tilde{\x}$. The $\Omega$ estimate is plugged in the NGD to draw a new set of $\e$ to augment $\x$, based on which a new  $\Omega$ estimate is obtained. The procedure continues until convergence.  PANDA achieves the same global optimum on $\Omega$ as the graphical ridge in expectation over the distribution of $\e$, as stated in Proposition \ref{prop:graphicalridge}. The proof is straightforward given that $E_\e(e_{ij}e_{ik})=\lambda \omega_{jk}$.
\begin{pro}[\textbf{Regularization effects of PANDA-graphical ridge for GGM}]\label{prop:graphicalridge}
The negative log-likelihood of $\Omega$ based on the augmented data in PANDA-graphical ridge is
$l_p(\Omega|\x,\e)=
(n+n_e)n^{-1}l(\Omega|\x)+
(n+n_e)n^{-1}_e\sum_{i=1}^{n_e}\sum_{j,k=1}^p (e_{ij}e_{ik})\omega_{jk}$, the expectation of which over the distribution of $\e$ is
\begin{equation}\label{eqn:nmgr}
\textstyle\E_{\e}(l_p(\Omega|\x,\e))=(n+n_e)\left(n^{-1}l(\Omega|\x)+\lambda  \sum_{j=1}^{p}\sum_{k=1}^{p}\omega_{jk}^2\right).
\end{equation}
\end{pro}


\subsection{An Additive NI Counterpart to PANDA}
As mentioned in Section \ref{sec:intro}, PANDA is inspired by the adaptive NI approach used in whiteout for regularizing NNs \citep{whiteout}. The additive NI approach directly perturbs the observed data with additive noise drawn from a NGD without altering the data dimension. Proposition \ref{prop:additive} shows there exists an additive NI counterpart to PANDA that achieves the same regularization effect on the parameters from a UGM. The proof of which is given in Appendix \ref{app:additive}.
\begin{pro}[\textbf{an additive NI counterpart to PANDA}]\label{prop:additive}
The expected regularization effects in PANDA-NS, PANDA-CD, PANDA-SCIO and PANDA-graphical ridge for GGM, and PANDA-NS for UGM can also be achieved via minimizing the second-order approximation of the expected loss function constructed with using additive NI  $\tilde{\x}_{ij}=\x_{ij}+\e_{ij}$ ($i=1,\ldots,n;j=1,\ldots,p$) in the regression on the outcome node $X_i$, where $\e_{ij}=(e_{ijj},\e_{ij,-j})$ is designed in the same way as the PANDA procedure.
\end{pro}
Though  the additive NI can be used to construct UGM with the same regularization effects as PANDA per  Proposition \ref{prop:additive}, it cannot be easily realized computationally in practice. In addition, PANDA provides the exact expected regularization effects as $n_e\rightarrow\infty$ while the expected regularization effects are only second-order approximate for the NI for non-Gaussian UGM.

\section{Bayesian Interpretation of PANDA}\label{sec:EB}
PANDA introduces endogenous information into the observed data as a way to regularize large models. This bears a resemblance to the Bayesian framework, where the endogenous information is often formulated in a prior distribution. Below we connect PANDA with the  Bayesian framework in two aspects.

Propositions \ref{prop:regularization} to  \ref{prop:graphicalridge}   show that the expectation of noise-augmented loss function over the distribution of noises in PANDA is equivalent to the original loss function (negative likelihood function) plus a penalty term $P(\Theta)$. We can always regard $\exp(-P(\Theta))$ as a prior on $\Theta$, regardless of whether it is proper or improper. For almost every regularizer discussed in Section \ref{sec:single}, there exists a Bayesian version, such as the Bayesian lasso \citep{bayeslasso2008}, Bayesian bridge \citep{bayesbridge2012}, Bayesian elastic net \citep{bayesen2010}, Bayesian group lasso \citep{bayesgroup2015}, and Bayesian graphical lasso \citep{bayesgl2012}. Despite the conceptual  connection with the Bayesian framework, the endogenous information introduced in PANDA, technically speaking, is not a prior distribution on the parameters per se, but represents  prior information in the form of ``noisy data'' in parallel to the observed data. Procedurally, PANDA optimizes a convex regularized  objective function  iteratively and outputs regularized MLEs whereas the full Bayesian hierarchical modeling often replies on posterior sampling to make inferences on model parameters.  


The generative distribution (or more specifically, its variance) of the noisy data in PANDA depends on the up-to-date parameter estimates, which is a function of the observed data. This conceptually relates to the empirical Bayesian (EB) framework, which refers to a Bayesian inferential procedure where the prior distribution is estimated from the data. We further explore the connection between PANDA and EB using two examples.

We first consider the regression coefficients $\theta_{jk}$ ($k\ne j$) with outcome node $X_j$ in the context of PANDA-NS. Specifically, we can reformulate the $(t+1)$-th iteration of the PANDA-NS algorithm in the EB framework, where the prior $\pi(\theta_{jk})$ is constructed from the data. For example, the bridge-type EB prior for $\theta_{jk}$ in GGM and UGM is
\begin{align}
\mbox{GGM: }&\pi(\theta_{jk}|\hat{\theta}^{(t)}_{jk}, \sigma_{j}^2)= \mbox{N}\left(0,\lambda^{-1}\left|\hat{\theta}^{(t)}_{jk}\right|^\gamma\sigma_{j}^{2}\right)\mbox{ for }k\ne j;\mbox{ and } \pi(\sigma_{j}^{2})\propto\sigma_{j}^{-2},\label{eqn:EBGGM}\\
\mbox{UGM: }&\pi(\theta_{jk}|\hat{\theta}^{(t)}_{jk})= \mbox{N}\left(0,(2\lambda)^{-1}\left|\hat{\theta}_{jk}^{(t)}\right|^{\gamma}\right),\label{eqn:EBUGM}
\end{align}
respectively, where $\hat{\theta}^{(t)}_{jk}$ is either the MAP estimate or a random posterior sample for $\theta_{jk}$ from the $t$-th iteration. The negative logarithm of the joint posterior distribution of $\bs{\theta}$ and $\bs{\sigma}^2\!=\!\{\sigma^2_1,\ldots,\sigma^2_p\}$ is 
$
{2}\!\sum_{j=1}^{p}\log(\sigma_{j}^2)\!+\!\sum_{j=1}^{p}\left(2\sigma_{j}^2\right)^{-1}\!\left(\sum_{i=1}^{n}\!\left(\!x_{ij}\!-\!\!\sum_{k\ne j}x_{ik}\theta_{jk}\!\right)^2\!\!\!\!+\!\lambda\!\sum_{k\ne j}\left|\theta_{jk}\right|^{2}|\hat{\theta}_{jk}^{(t)}|^{-\gamma}\right)$
for GGM; and that of $\bs{\theta}$ is
$
-\sum_{j=1}^{p}\!\!\left(\!\sum_{i=1}^{n}\!\left(\! h_j(x_{ij})\!+\!\!\sum_{k\ne j}\theta_{jk}x_{ij}x_{ik}\!-\!B_j\!\!\left(\!\sum_{k\ne j}\theta_{jk}x_{ik}\!\right)\!\right)\!-\!\lambda\!\sum_{k\ne j}\left|\theta_{jk}\right|^{2}|\hat{\theta}_{jk}^{(t)}|^{-\gamma}\!\right)$ for UGM.
When the per-iteration EB prior is constructed using a random posterior sample from the last iteration, we can also obtain the posterior distributions for $\theta_{jk}$ upon convergence. If the posterior distributions from all $\theta_{jk}$'s are graphed in one plot, a notable separation would be expected between the distributions the MAPs of which are approximately zero and the spread of which a very narrow, and those the MAPs of which are clearly not zero the scales of which are evidently larger.  If the per-iteration EB prior is constructed using the MAP from the last iteration, there will still be some fluctuation around the MAP samples over the iterations upon convergence, reflecting the Monte Carlo errors. In other words, if the MAP was calculated with an  infinite number of posterior samples in an iterative or if the closed-form MAP exists, then no fluctuation around the MAPs across the iterations would he expected. Regardless of whether $\theta_{jk}^{(t)}$ used throughout the iterations is a MAP or a random posterior sample, as long as the expected regularizer is convex, the MAP for $\theta_{jk}$ is the same as the minimizer of the regularized loss functions in Eqns (\ref{eqn:Elp}) and (\ref{eqn:glmregularization}) respectively upon the convergence of the posterior distribution of $\theta_{jk}$ through the iterative procedure.

As a second example, we consider the  graphical ridge regularization for GGM. In the $(t+1)$-th  iteration, rather than augmenting $\x$ with a noise matrix, we impose an EB prior on $\Omega$
$$\pi(\Omega|\hat{\Omega}^{(t)})= \mbox{Wishart}_p\left(\left(\lambda\hat{\Omega}^{(t)}\right)^{-1}\!\!,\;\nu=p+1\right),$$
where $\hat{\Omega}^{(t)}$ is either the MAP estimate or a random posterior sample for $\Omega$. Due to conjugacy of the EB prior for  the Gaussian likelihood function of $\Omega$, we obtain easily obtain the posterior distribution for $\Omega$, the negative logarithm of which is
$$\textstyle\frac{n}{2}\log(|\Omega|)+\frac{1}{2}\mbox{tr}\left(\Omega\cdot\left(\X^T\X+\lambda\hat{\Omega}^{(t)}\right)\right)+\mbox{const.},$$
as well as the MAP estimate or a random posterior sample for $\Omega$ to be used for constructing the EB prior for the iteration. Upon the convergence of the iterative procedure, the MAP would be equivalent to the minimizer with the graphical ridge regularization in Eqn (\ref{eqn:nmgr}), regardless of whether $\hat{\Omega}^{(t)}$ is a MAP or a random sample.

The above two examples demonstrate that the minimizer of $l_p(\Theta|\x,\e)$ in PANDA  is equivalent to the MAP of the posterior distribution $\Theta$ obtained via an iterative procedure with an \emph{adaptive} EB prior constructed using the MAP or a random posterior sample. PANDA  with other regularizers can also be derived in a similar manner. When the targeted regularizer is non-convex, if the MAP  from the last iteration  is used to construct the prior from the last iteration, then the converged MAP value would depend on the starting value; if a random sample from the posterior distribution  from the last iteration is used to construct the prior, then the converged posterior distribution could exhibit multi-modality.

\section{Theoretical Properties and Statistical Inferences with PANDA}\label{sec:theory}
Section \ref{sec:single}  establishes PANDA as a  regularization technique for UGM and GGM construction.  In this section, we establish the almost sure (a.\ s.) convergence of the data augmented $l_p(\Theta|\x,\e)$ to its expectation and the a.\ s.\ convergence of the minimizer of the former to the minimizer of the expected loss function  as $n_e\rightarrow\infty$ or $m\rightarrow\infty$ in the framework of PANDA-NS for GGM and UGM (Sec \ref{sec:consist}). 
In addition, we examine the Fisher information of the parameters in noise-augmented data (Sec \ref{sec:fisher}) and statistical inferences of the parameters via PANDA in the GLM setting (Sec \ref{sec:asymp.dist}). Finally, we provide a formal test on the convergence of the PANDA algorithms (Sec \ref{sec:convergence}).

\subsection{Almost sure convergence of noise-augmented loss function and its minimizer for PANDA-NS}\label{sec:consist}
Let $\Theta$ denote the collection of all parameters from the $p$ regression models for UGM-NS. The conditional distribution of each node given others is modelled by an exponential family, depending on the node type. 
For example, the averaged loss function for GGM, PGM, and NBGM over $m\ge1$ iterations in the PANDA algorithm  $\bar{l}_p(\Theta|\x,\e)$  is 
\begin{align*}
&\textstyle l(\Theta|\x)\!+\! m^{-1}\sum_{t=1}^m\sum_{i=1}^{n_e}\sum_{j=1}^{p}\!\left(\!\sum_{k\ne j}e^{(t)}_{ijk}\theta_{jk}\!\right)^2,\\
&l(\Theta|\x)-m^{-1}\textstyle\!\sum_{t=1}^{m}\!\sum_{i=1}^{n_e}\!\sum_{j=1}^{p}\!\left(\!e_{ijj}(\theta_{j0}\!+\!\!\sum_{k\ne j}\!e_{ijk}\theta_{jk})\!-\! \log(e^{(t)}_{ijj}!)-\exp\left(\theta_{j0}\!+\!\!\sum_{k\ne j}\!e^{(t)}_{ijk}\theta_{jk}\!\right)\!\right),\\
&l(\!\Theta|\x)\!-\!m^{-1} \!\textstyle\sum_{t=1}^{m}\!\sum_{i=1}^{n_e}\!\sum_{j=1}^{p}\!\!\left\{\log\!\left(\!\frac{\Gamma(e_{ijj}+r_j)r_j^{r_j}}{\Gamma(e_{ijj}\!\!+\!1)\Gamma(r_j)}\!\!\right)\!\!+\!e_{ijj}\!\left(\theta_{j0}\!+\!\!\sum_{k\ne j}\!e_{ijk}\theta_{jk}\right)\right.\\
&\mbox{\hspace{150pt}}\textstyle\left.-(r_j\!+\!e_{ijj})
\log\left[r_j\!+\!\exp\left(\theta_{j0}\!+\!
\sum_{k\ne j}\!e_{ijk}\theta_{jk}\right)\right]\right\},
\end{align*}
respectively, where $l(\Theta|\x)=\sum_{i=1}^{n}\!\sum_{j=1}^{p}\!\left(\!x_{ij}\!-\!\sum_{k\ne j}x_{ik}\theta_{jk}\!\right)^2$ for GGM, and  is the  negative log-likelihood for PGM and NBGM, respectively. Theorem \ref{thm:pelf2nmelf} presents the asymptotic properties of $\bar{l}_p(\Theta|\x,\e)$ in a UGM under two scenarios: 1) $n_e\rightarrow\infty$ while $n_e\V(e_{jk})=O(1)$ for a given $\theta_{jk}$ and $m\;(\ge 1)$ is fixed at a constant; 2) $m\rightarrow\infty$ while $n_e$ and $(n_e>p-n)$ is fixed at a finite constant.
\begin{thm}{\textbf{(asymptotic properties of the noise-augmented loss function and its minimizer in PANDA)}}\label{thm:pelf2nmelf}
Assume $\Theta$ belongs to a compact set. Let $l_p(\Theta|\x)=\E_{\e}(l_p(\Theta|\x,\e))$. \\
1) If $n_e\rightarrow\infty$ while $n_e\V(e_{jk})=O(1)$ for a given $\theta_{jk}$ and $m\;(\ge1) $ is held at a constant, then
\begin{align}
n_e^{1/2}C_1^{-1}\left(\bar{l}_p(\Theta|\x,\e)-l_p(\Theta|\x)\right)&\overset{d}{\longrightarrow} N(0,1)\label{eqn:d1}\\
\bar{l}_p(\Theta|\x,\e)&\overset{a.s.}{\longrightarrow}l_p(\Theta|\x)\!\overset{n_e\rightarrow \infty}{\longrightarrow}\!\textstyle l(\Theta|\x)\!+P(\Theta)+C\\
\arg\inf\limits_{\Theta}\bar{l}_p(\Theta|\x,\e)&\overset{{a.s.}}{\longrightarrow}\arg\inf\limits_{\Theta}l_p(\Theta|\x),\label{eqn:as1argmin}
\end{align}
where $P(\Theta)$ is the same as defined in Proposition \ref{prop:glmregularization}, and $C_1$ is a function of $\Theta$ and  takes different forms for different distributions in an exponential family. 

2) If $m\rightarrow\infty$ while  $n_e$ is fixed, then
\begin{align}
m^{1/2}C_2^{-1}\left(\bar{l}_p(\Theta|\x,\e)-l_p(\Theta|\x)\right)&\overset{d}{\longrightarrow}N(0,1)\label{eqn:d2}\\
\bar{l}_p(\Theta|\x,\e)&\overset{a.s.}{\longrightarrow}l_p(\Theta|\x)
\overset{m\rightarrow \infty}{\longrightarrow}
l(\Theta|\x)+ P(\Theta)+C\label{eqn:as2l}\\
\arg\inf\limits_{\Theta}\bar{l}_p(\Theta|\x,\e)&\overset{{a.s.}}{\longrightarrow}\arg\inf\limits_{\Theta}l_p(\Theta|\x),\label{eqn:as2argmin}
\end{align}
where $P(\Theta)$ is the same as defined in Proposition \ref{prop:glmregularization}, and $C_2$ is a function of $Theta$ and takes different forms for different exponential.
\end{thm}
The proofs of Theorem \ref{thm:pelf2nmelf} are provided in Appendix \ref{app:pelf2nmelf} for GGM, BGM, PGM, EGM, and NBGM. The theorem can be proved for other graph types, including mixed graphs, in a similar manner. It can be shown Theorem \ref{thm:pelf2nmelf} holds for PANDA-graphical ridge (the proof is available in Appendix \ref{app:pelf2nmelf}), where the average loss over $m$ iterations is $\bar{l}_p(\Omega|{\x},{\e})=n^{-1}(n+n_e)l(\Omega|\x)+ m^{-1}n_e^{-1}(n+n_e)\sum_{t=1}^m\sum_{i=1}^{n_e}\sum_{j,k=1}^{p}(e_{ijj}e_{ijk})\omega_{jk}$.

There are two important takeaways from Theorem \ref{thm:pelf2nmelf}. First, it states that $\bar{l}_p(\Theta|\x,\e)$ follows a Gaussian distribution at the rate of $\sqrt{n_e}$ and $\sqrt{m}$ under the two scenarios, respectively, suggests the augmented loss function in PANDA is trainable for practical implementation. Specifically, the fluctuation in $\bar{l}_p(\Theta|\x,\e)$ around its expected value is controlled and the tail of the distribution of $d=\bar{l}_p(\Theta|\x,\e)-l_p(\Theta|\x)$  decays
to zero exponentially fast in $n_e$ and $m$ as $\Pr(d>t)\le \exp(-n_et^2/2C^2)$ and $\Pr(d>t)\le \exp(-mt^2/2C^2)$ for any $t>0$. 
Second, $\bar{l}_p(\Theta|\x,\e)$ converges a.\ s.\ to its expectation, which is the penalized loss function given $\x$ with the targeted penalty term (arbitrarily well for $n_e\rightarrow\infty$ and under certain scenarios which are often the case in practice or $m\rightarrow\infty$ ), guaranteeing that PANDA optimizes what it is supposed to optimize.


When there exists multi-collinearity among the covariates  and if the imposed sparsity regularization is not strong enough, then the loss function minimized in PANDA would have a global optimum region rather than a single  optimum point. To examine the asymptotic properties in this case, we first define  the \emph{optimum parameter set} (Definition \ref{def:optimum}), then show that the parameters learned by PANDA from minimizing the $l_p(\Theta|\x,\e)$  fall into the optimum parameter set asymptotically (Proposition \ref{prop:consistmcl}).  The proof is given in Appendix \ref{app:consistmcl}.

\begin{defn}{(\textbf{optimum parameter set})}\label{def:optimum} Let the expected loss function $l_p(\bs{\theta}|\x)$ be a continuous function in $\bs{\theta}$. The optimum parameter set is defined as
$\hat{\bs\Theta}^0\!=\!\left\{\bs{\theta}^0\!\in\!{\bs\Theta} \mid l_p( \bs{\theta}^0|\x)\!\leq\!l_p(\bs{\theta}|\x), \forall\;\bs{\theta}\!\in\!{{\bs\Theta}}\right\}$, and the distance from $\bs{\theta}\in\bs\Theta$ to $\hat{\bs\Theta}^0$ is defined as $d\left(\bs{\theta}, {\hat{\bs\Theta}}^0\right)=\min\limits_{\bs{\theta}^0\in\hat{\bs\Theta}^0}||\bs{\theta}-\bs{\theta}^0||_2$.
\end{defn}
\begin{pro}{\textbf{(consistency of parameter estimate in the presence of multicollinearity)}}\label{prop:consistmcl}
Let $\hat{\bs{\theta}}_p^0\!=\!\arg\min\limits_{\Theta}\bar{l}_p(\Theta|\x,\e)$ in PANDA. 
Given
\begin{equation}\label{eqn:sup}
\sup\limits_{\Theta}\left|\bar{l}_p(\Theta|\x,\e)\!-\!\bar{l}_p(\Theta|\x)\right|\rightarrow0 \mbox{ as  $n_e\!\rightarrow\!\infty \bigcap n_e\V(e_{jk})\!=\!O(1)\; \forall j\ne k=1,...,p$, or $m\rightarrow\infty$;}
\end{equation}
and assume ${\bs\Theta}$ is compact, then
$\Pr\left(\limsup\limits_{m\rightarrow\infty\mbox{ or }n_e\rightarrow\infty} d\left(\hat{\bs{\theta}}_p^0,\hat{{\bs\Theta}}^0 \right)\leq\delta \right) =1\;\forall\; \delta>0.$
\end{pro}
Multi-collinearity does not affect the convergence of the loss functions in PANDA-NS; therefore, Eqn (\ref{eqn:sup}) holds per the proof of Theorem \ref{thm:pelf2nmelf}.

\subsection{Fisher information in noise augmented data}\label{sec:fisher}
The augmented noisy data in PANDA bring endogenous information to observed data $\x$ to regularize the estimation of $\Theta$. The expected regularization can be achieved  by either letting $(n_e\rightarrow\infty) \cap (n_e\V(e_{jk})=O(1))$ or $m\rightarrow\infty$ (Sec \ref{sec:nongaussian1} and \ref{sec:consist}).  At first sight,  it seems that the large amount of augmented noisy data could potentially overshadow the information on parameters contained in the observed data, leading to over-regularization. We claim that this is not the case because of the constraint  $n_e\V(e_{jk})=O(1)$ for a given $\theta_{ik}$. In other words, $n_e$ combined with the tuning parameters from the NGD noise term is treated as one tuning parameter. For example, with the lasso-type noise, $n_e\lambda$ is tuned together: if $n_e$ is large, then $\lambda$ would take a small value so to keep $n_e\lambda=O(1)$ and lead to the targeted regularization. Proposition \ref{prop:fisher} provides theoretical justification that as long as $n_e\V(e_{jk})=O(1)$  for any given $\theta_{jk}$, the amount of regularization brought by the augmented data to $\theta_{jk}$ remains at constant even as  $n_e\rightarrow\infty$. Proposition \ref{prop:fisher} is established in the context of the bridge-type noise; the same conclusion can be obtained with other noise types in a similar fashion. The proof is provided in Appendix \ref{app:fisher}.  

\begin{pro}\label{prop:fisher}
The regularization on the regression coefficient $\bs\theta_j$ in the regression of $X_j$ on $\X_{-j}$ introduced through the augmented bridge-type noise drawn is proportional to $n_e\lambda|\theta_k|^{-\gamma}$. Specifically, $I_{\tilde{\x}}(\bs\theta_j)$, the Fisher information on $\bs\theta_j$ contained in the augmented data $\tilde{\x}$ is the summation of $I_{\x}(\bs\theta_j)$, the Fisher information on $\bs\theta_j$ contained in the observed data, and $I_{\e}(\bs\theta_j)$, the amount of regularization on  $\bs\theta_j$.
\begin{equation}
\textstyle {I_{\tilde{\x}}}(\bs\theta_j)= {I_\x}(\bs\theta_j)+(\lambda n_e){B}_j''(\theta_{j0}+0) \mbox{Diag}\{|\theta_{j1}|^{-\gamma},\ldots,|\theta_{jp}|^{-\gamma}\} +O\left(\lambda n_e^{1/2}\right)J_p),\label{eqn:fisher}
\end{equation}
\end{pro}
where $J_p$ is a $p\times p$ matrix with all elements equal to 1. The higher-order term $O\left(\lambda n_e^{1/2}\right)$ becomes $O(\lambda^{1/2})$ if $\lambda n_e=O(1)$ and be ignorable if $\lambda$ is small.  Eqn (\ref{eqn:fisher}) suggests that the information about $\theta_{jk}$ (for $k \ne j$) does not increase with $n_e$ as along as $\lambda n_e|\theta_{jk}|^{-\gamma}$ is kept at a constant. In practice, we could treat $\lambda n_e$ as one tuning parameter. In addition, the closer $|\theta_{jk}|$ is to 0, the more regularization the augmented information brings to  $\theta_{jk}$.

\subsection{Asymptotic distribution of regularized  parameters via PANDA in GLM}\label{sec:asymp.dist}
In each iteration of the PANDA algorithm, a GLM is run with each node as the outcome in the PANDA-NS approach (as well as in PANDA-CD and PANDA-SPACE for GGM). We derive the asymptotic distribution  for the regularized $\hat{\bs\theta}$ in the GLM (with linear regression included as a special case), based on which we can obtain inferences, such as  confidence intervals (CI), for $\bs{\theta}$. In contrast to some existing post-selection inferential approaches in GLM, where inferences follow  variable selection in a two-stage manner, PANDA achieves variable selection and parameter estimation and inferences simultaneously,  regardless of whether a parameter estimate is zero or not,  with much better coverage rates (see the simulation results in Sec \ref{sec:sim2}). The results presented below  focus on the inferences for a single GLM through PANDA rather than UGMs due to two considerations. First, the analysis of UGMs often focuses on the construction of the network, that is, whether an edge exists or not between two nodes. Achieving this goal does not necessarily enlist the help of statistical inferences if the construction method itself has a build-in thresholding rule that leads to sparse solutions. Second, given the large amount of parameters involved in a UGM, inferences can be difficult to comprehend, and multiplicity correction procedures would become necessary, depending on the context.


\begin{pro}[\textbf{Asymptotic distribution of parameter estimates via PANDA in GLM}]\label{prop:asymp.dist.UGM}
WLOG, denote the outcome in a GLM by $X_1$ and covariate by $\X_{-1}$. Let $I_{\x}(\bs{\theta})$ denote the Fisher information  in the original data $\x$, and $I_{\tilde{\x}}(\bs\theta)$  be the  Fisher information in the noise-augmented data $\tilde{\x}=(\x,\e)$.  Let $\hat{\bs\theta}^{(t)}$ be the estimate of $\bs{\theta}$ in iteration $t$ and the final estimate for $\bs{\theta}$ is $\bar{\bs{\theta}}=r^{-1}\sum_{t=1}^r\hat{\bs\theta}^{(t)}$ from $r\ge1$ iterations after the convergence of the PANDA algorithm in the GLM. If $n_e\V(e)=o(\sqrt{n})$ for any given $\theta$, then
\begin{align}
\sqrt{n}(\hat{\bs{\theta}}^{(t)}-\bs{\theta})
&\overset{d}{\rightarrow}  N(\0,\Sigma^{(t)})
\mbox{ as } n\rightarrow\infty, \label{eqn:conditional.asym.disn}\\
\sqrt{n}(\bar{\bs{\theta}}-\bs{\theta})
&\overset{d}{\rightarrow} N\left( \0,\bar{\Sigma}+ \Lambda \right)\mbox{ as } n\rightarrow\infty;r\rightarrow\infty, \label{eqn:asym.disn}
\end{align}
where $\Sigma^{(t)}= I_{\tilde{\x}^{(t)}}(\bs\theta)^{-1}I_{\x}(\bs\theta)I_{\tilde{\x}^{(t)}}(\bs\theta)^{-1}$ in iteration $t$,  $\bar{\Sigma}=r^{-1}\sum_{t=1}^r \Sigma^{(t)}$, and
$\Lambda=\V(\hat{\bs{\theta}}^{(t)})$, the between iteration variability of $\hat{\bs{\theta}}^{(t)}$.
\end{pro}
The  proof of Proposition  \ref{prop:asymp.dist.UGM} is given in Appendix \ref{app:CI.UGM}.  The regularity condition  $n_e\V(e)=o(\sqrt{n})$ takes different forms for different NGDs (e.g., for the bridge-type noise, it would be $\lambda n_e=o(\sqrt{n})$).  The asymptotic variance of $\hat{\bs{\theta}}^{(t)}$ involve the inverse of $I_{\tilde{\x}^{(t)}}(\bs\theta)$, which exists with the augmented data in PANDA.  Eqn (\ref{eqn:asym.disn}) suggests the overall variance on $\bar{\theta}$ is the summation of two variance components, $\bar{\Sigma}$, the per-iteration variance of $\hat{\bs{\theta}}^{(t)}$,  and  $\Lambda$, the between-iteration variance of $\hat{\bs{\theta}}^{(t)}$.  $\bar{\Sigma}$  contains the unknown $\bs\theta$ and can be estimated by plugging in $\hat{\bs{\theta}}^{(t)}$, with the caveat that the uncertainty around $\hat{\bs{\theta}}^{(t)}$ is not accounted for. $\Lambda$ can be estimated by the sample variance of  $\hat{\bs{\theta}}^{(t)}$ over $r$ iterations; that is, $(r-1)^{-1}\sum_{t=1}^r\left(\hat{\bs{\theta}}^{(t)}-\bar{\bs\theta}\right)
\left(\hat{\bs{\theta}}^{(t)}-\bar{\bs\theta}\right)'$.

A special case of Proposition \ref{prop:asymp.dist.UGM} is linear regression, where the asymptotic distribution of $\hat{\bs{\theta}}^{(t)}$ in Eqn (\ref{eqn:conditional.asym.disn}) becomes
\begin{equation}\label{eqn:conditional.asym.disn.gaussian}
\sqrt{n}(\hat{\bs{\theta}}^{(t)}-\bs{\theta})\overset{d}{\rightarrow} N\left(\0,\sigma^2 (\bs M^{(t)})^{-1}(\x_{-1}'\x_{-1})(\bs M^{(t)})^{-1} \right),
\end{equation}
where $\mathbf{M}^{(t)}\!=\!(\x_{-1}'\x_{-1}+n_e\mbox{diag}(\V(\e))$. $\V(\e)$ is the variance of the augmented noise to the covariates; e.g., $\V(\e)=\lambda|\bs\theta|^{-1}$ for the lasso-type noise. The asymptotic variance in Eqn (\ref{eqn:conditional.asym.disn.gaussian}) contains  unknown $\sigma^2$ and can be estimated by  $\hat{\sigma}^2=\mbox{SSE}/(n-\nu)=\left({\x}_1-{\x}_{-1}\hat{\bs{\theta}}^{(t)}\right)'\left({\x}_1-{\x}_{-1}\hat{\bs{\theta}}^{(t)}\right)/(n-\nu)$, where the degree of freedom $\nu\!=\!\mbox{tr}(\x_{-1}(\bs M^{(t)})^{-1}\x_{-1}')$.  $\hat{\sigma}^2$ converges to
$\sigma^2 \chi_{n-\nu}^2$ in distribution. 


When applying the PANDA technique to obtain statistical inferences in GLMs in addition to variable selection, we should set $n_e$ at a small number and $m$ at a large number to achieve valid statistical inferences and the targeted regularization effect  simultaneously. We recommend $n_e=o(n)$ as long as $n_e+n>p$ (e.g., one-order of magnitude smaller than $n$), especially when $n$ is relatively small.  This is different from when the main goal is just variable selection, where  a large $n_e$ can be used to achieve the expected regularization effect with less iterations. The reason for this is that large $n_e$ (relative to $n$) tends to lead to underestimated $\bar{\Sigma}+\Lambda$, the asymptotic variance of $\bar{\bs{\theta}}$, resulting in lower-than-nominal coverage rates and inflated type I error rates. As mentioned above, $\bar{\Sigma}=r^{-1}\sum_{t=1}^r \Sigma^{(t)}$ is estimated by plugging in $\hat{\bs{\theta}}^{(t)}$ for $t=1,\ldots,r$ upon convergence, pretending it is the true parameter value and ignoring the variability around it. Though this issue exists regardless of whether a large or a small $n_e$ is used, using  a small $n_e$ helps to re-capture this lost variability with the between-iteration variability $\Lambda$. The rationale behind this is given below. $\hat{\bs{\theta}}^{(t)}$ is a regularized estimate with an externally imposed constraint by minimizing a loss  function summed over the data component $\x$ and the regularizer component, or equivalently, a summation of loss functions constructed with the data component $\x$ and with the augmented data component $\e$ in the context of PANDA. Instead of focusing on how $\hat{\bs{\theta}}^{(t)}$ changes with sample data $\x$, which is fixed throughout iterations, we shift to quantifying how it changes with $\e$.  If a large $n_e$ is used,
the ignored sampling variability around $\hat{\bs{\theta}}^{(t)}$ can hardly be recovered through $\Lambda$ as it is close to 0, which is easy to understand as the realized regularization effect with a large $n_e$ is close to its expectation and it is almost like solving the same analytical constrained optimization at every iteration,  leading to very similar $\hat{\bs{\theta}}^{(t)}$ across iterations upon convergence.


\subsection{Test of convergence of PANDA algorithm}\label{sec:convergence}
When presenting the PANDA algorithms  in Section \ref{sec:single}, we recommend 3 criteria for evaluating the convergence of the PANDA algorithms, one of which is a formal statistical test. This test is asymptotic in the sense that it assumes $n_e\rightarrow\infty$ or $m\rightarrow\infty$, but should work well when either $n_e$ or $m$ is relatively large in practice, which is often the case when  PANDA is implemented. WLOG, we establish the test for $n_e\rightarrow\infty$; the procedure is similar for $m\rightarrow\infty$ by replacing $n_e$ with $m$..

Theorem \ref{thm:pelf2nmelf} shows that as  $n_e\rightarrow\infty$, the distribution of the loss function in iteration $t$ converges to a Gaussian distribution (Eqn (\ref{eqn:d1})). The asymptotic Gaussian distribution involves $C_1(\Theta)$, which is unknown and can be estimated by plugging the $\hat{\Theta}^{(t)}$ from the current iteration. Specifically,
$$\textstyle C_1^{(t)}= \frac{\lambda n_e}{2}\!\left(\!\sum_{j=1}^{p}\kappa  \bigg|\bigg|\!\left(\hat{\bs\theta}^{(t)}_{j,-j}\big|{\hat{\bs\theta}_{j,-j}^{(t)}}\big|^{-\gamma/2}\right)\left(\hat{\bs\theta}^{(t)}_{j,-j}\big|{\hat{\bs\theta}_{j,-j}^{(t)}}\big|^{-\gamma/2}\right)^T\bigg|\bigg|_2^2\right)^{1/2},$$
where $\kappa$ is a constant that depends on the graph type ($\kappa=8$ for GGM, $2\exp(2\theta_{j0})$ for PGM, $2$ for EGM, $2\exp(2\theta_{j0})/(1+\exp(2\theta_{j0}))^4$ for BGM and $2r^2_j\exp(2\theta_{j0})/(r_j+\exp(\theta_{j0}))^2$ for NBGM where $r_j$ is the failure numbers in the regression with outcome node $X_j$; see Eqns (\ref{app:asygauss1}), (\ref{app:asypoi1}) and (\ref{app:asynb1})).

Let $d^{(t)}=\bar{l}_p(\x,\e^{(t+1)})-\bar{l}_p(\x,\e^{(t)})$ be the difference in the loss function from two consecutive iterations of the PANDA algorithm, which is  $n_e^{-1/2}\left(C_1^{(t+1)}z^{(t+1)}-C_1^{(t)}z^{(t)}\right)$ per Eqn (\ref{eqn:d1}). If the PANDA algorithm converges, the estimates $\hat{\Theta}^{(t)}$ stabilizes, so does $C_1^{(t)}$; in other words, $C_1^{(t+1)}\approx C_1^{(t)}$ and the difference $d^{(t)}$ should mostly be due to the randomness of the Gaussian noise terms with an expected mean of 0; that is,
\begin{equation}\label{eqn:d}
z^{(t)}= d^{(t)}/\sqrt{n^{-1}_e \left[C_1^{(t)2}+C_1^{(t+1)2}\right]}.
\end{equation}
since $z^{(t)}$ is independent from $z^{(t+1)}$ (augmented noises  are drawn independently across iteration).  If $|z^{(t)}|>z_{1-\alpha/2}$, then we may claim the PANDA algorithm has not converged at iteration $t$ at the significance level of $\alpha$.

\section{Simulation}\label{sec:simulation}
\subsection{Graph construction}\label{sec:sim1}
\begin{figure}[!htbp]
\begin{minipage}{0.96\textwidth}
\includegraphics[width=0.32\linewidth]{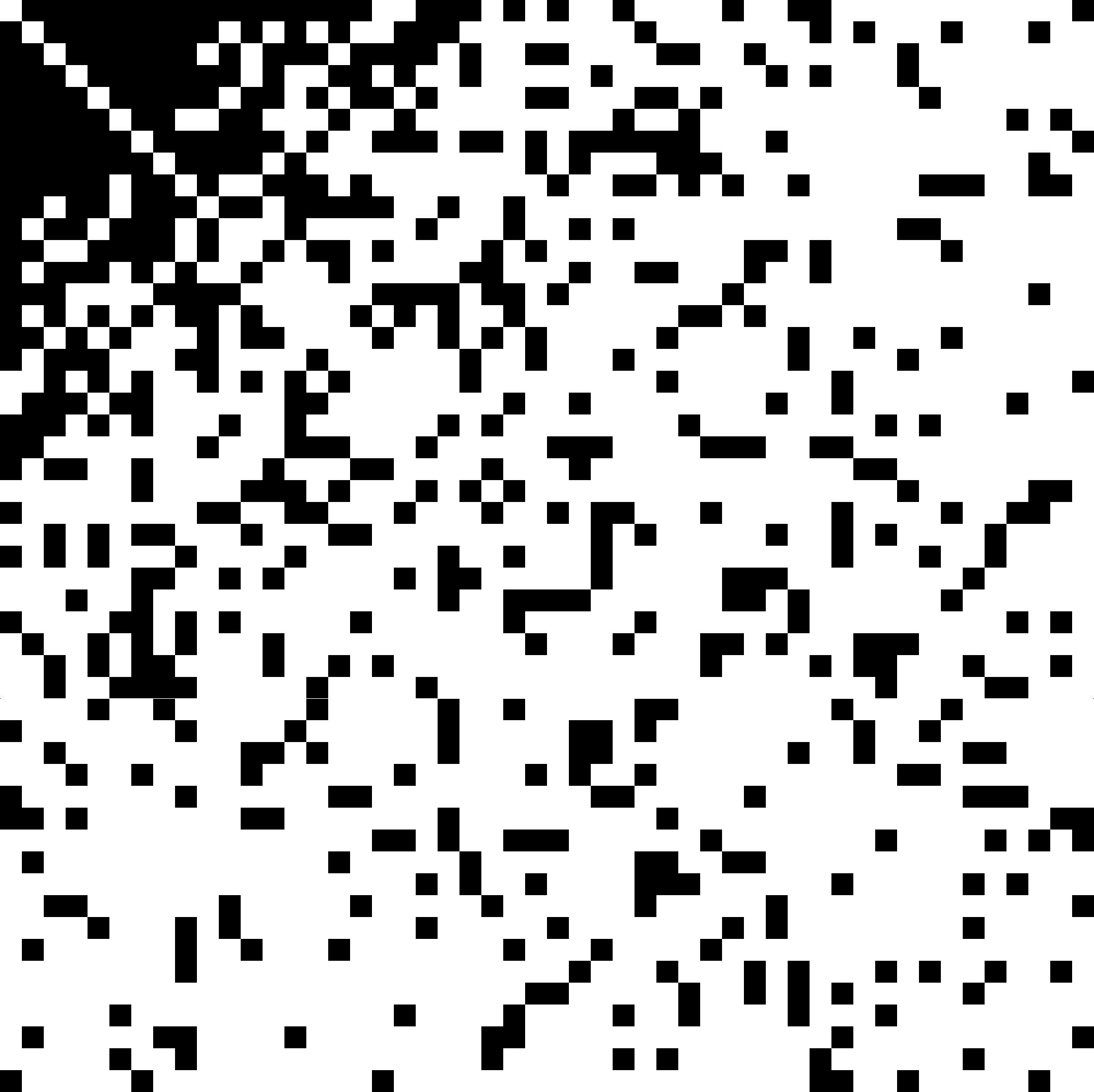}
\includegraphics[width=0.32\linewidth]{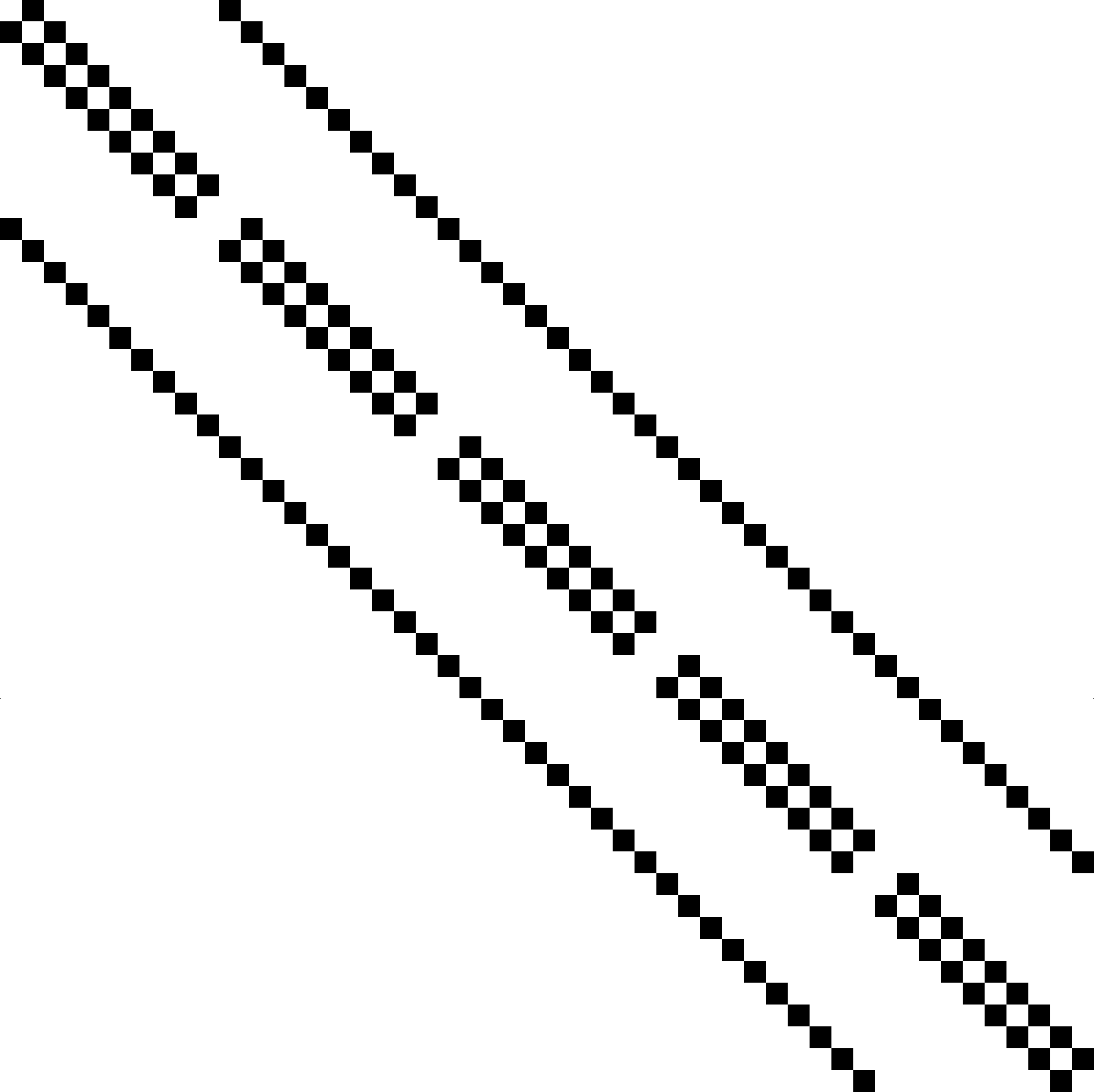}
\includegraphics[width=0.32\linewidth]{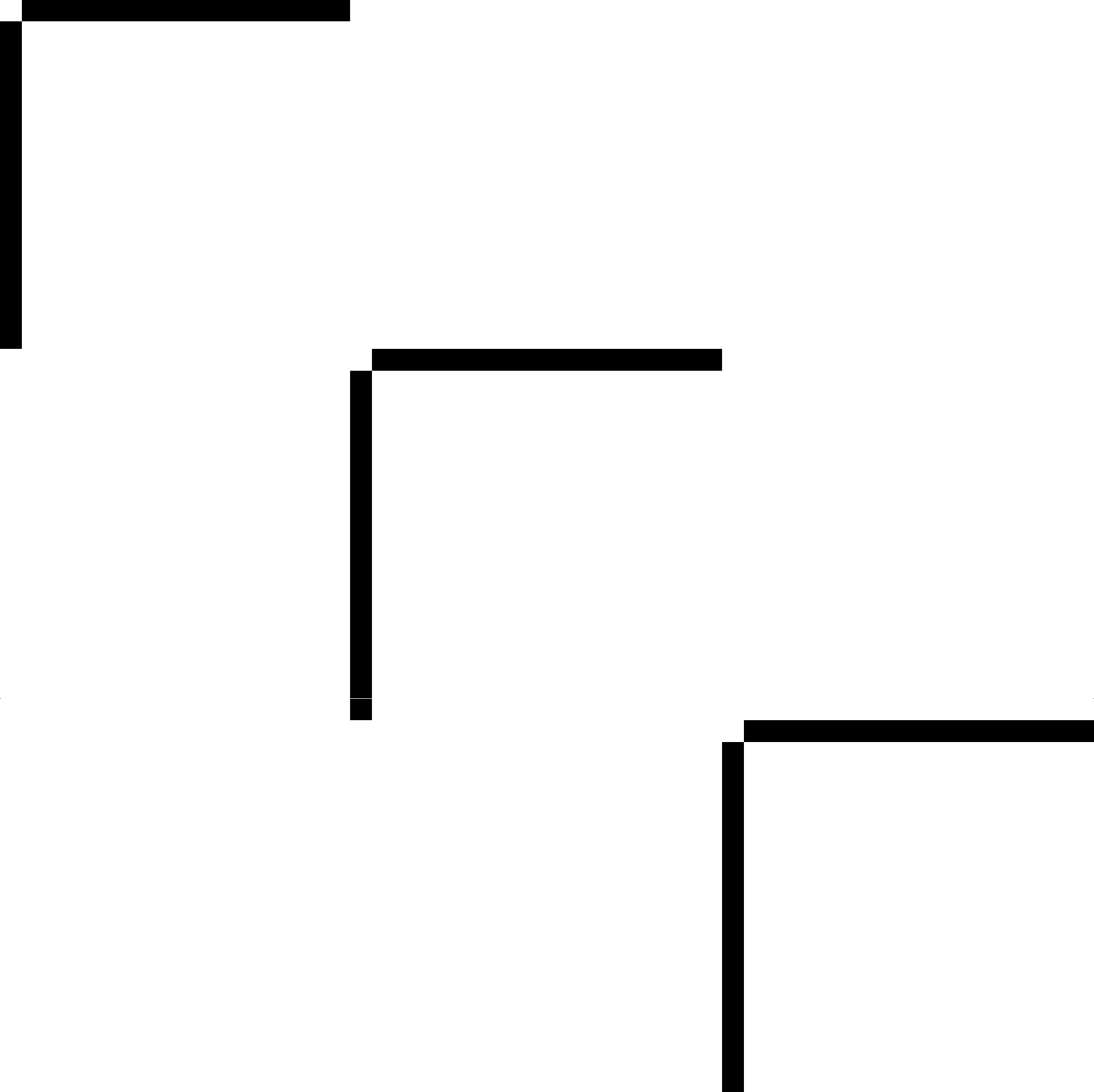}\\
\centering{scale-free \hspace{3.5cm} lattice-structured \hspace{2.5cm} 3-hub}
\end{minipage}
\caption{Adjacency matrices ($50\times50$) of the 3 types of networks examined in the simulation studies (edges are represented by the solid points)}\label{fig:single.adjacency}
\end{figure}

We implement PANDA-NS with the lasso-type penalty for constructing three  graphs (GGM, PGM, and BGM) and benchmark its performance against the NS approach with the constrained optimization when the graph takes on three types of adjacency matrix as depicted in Figure \ref{fig:single.adjacency}. The first  is a scale-free network the degree distribution of which follows a power law; the second has a banded structure, resulting in a lattice connection pattern among the nodes;  and the third network has 3 hub nodes.
In the estimation of the GGM, besides the NS approach, we also construct the graph using the CD and SCIO approaches.  To simulate the graphs, we first generated the adjacency matrix $\mathbf{A}$, conditional on which nodes $\X$ were generated using via R function \texttt{XMRF.sim} in package \texttt{XMRF} \citep{RXMRF}.  Table \ref{tab:sim1} summarizes the simulation schemes, and the tuning parameter and the algorithmic parameter specification for the PANDA algorithms.   
\begin{table}[!htp]
\begin{center}
\begin{tabular}{cccc |  ccc| ccccccc} 
\hline
\multicolumn{4}{c|}{simulation scheme} &
\multicolumn{9}{c}{tuning and algorithmic parameter in PANDA} \\
\hline
graph & n & p & non-zero edges${}^\star$& & $\gamma$ & $\sigma^2$ & $T$ & $n_e$ & $m$ & $\tau_0$ &  $r$ & other \\
\hline
GGM & 100 & 50 & 322, 85, 47 & NS & 1 & 0 & 70& 2000 & 1 & $10^{-6}$ & 100 & -\\ 
 &&&& CD& 1 & 0 & 70 &  2000 & 1 & $10^{-6}$  & 100& $k\!=\!5$\\
 &&&& SCIO& 1 & 0 & 150 & 2500 & 1 & $10^{-5}$  & 100 & $\tau_1\!=\!10^{-6}$\\
\hline
BGM & 100 & 50 & 322, 85, 47 & & 1 & 0 & 100 & 2000 &20 & $10^{-5}$ & 100 &- \\
\hline
PGM & 100 & 50 & 322, 85, 47 &  & 1 & 0 & 70  & 2000 & 1 &$10^{-5}$ & 100&- \\
\hline
\end{tabular}
\begin{tabular}{l}
${}^\star$\footnotesize For scale-free, lattice-structured, and 3-hub networks, respectively, out of a total of 1225 possible edges.\\
\hline
\end{tabular}
\end{center}
\caption{Simulation schemes, and tuning and algorithmic parameter specifications in PANDA}  \label{tab:sim1}
\end{table}

\begin{figure}[!htb]
\centering\textcolor{magenta}{\textbf{scale-free}}\\
\begin{minipage}{0.32\textwidth}
\includegraphics[width=1\linewidth]{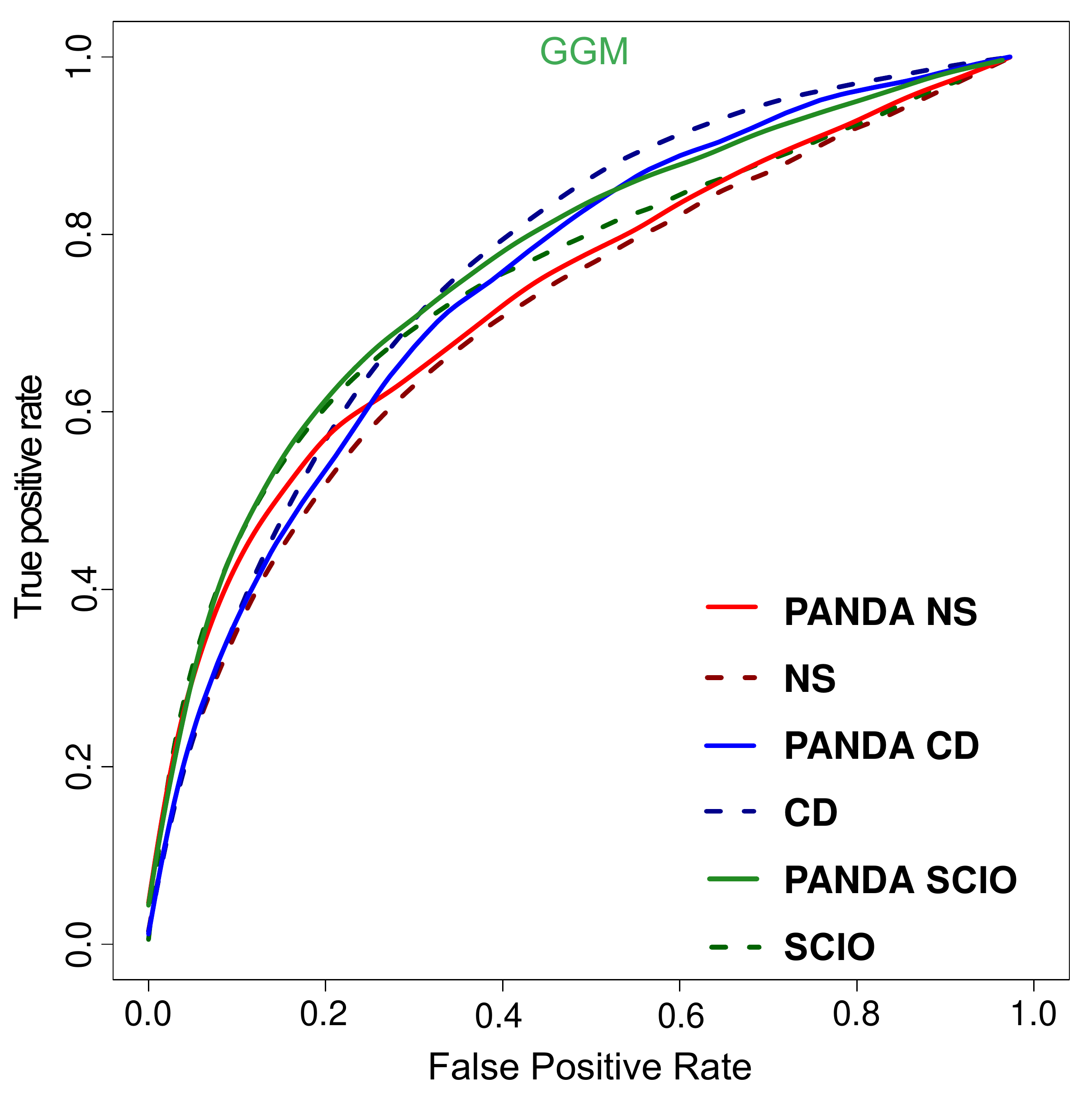}
\end{minipage}
\begin{minipage}{0.32\textwidth}
\includegraphics[width=1\textwidth]{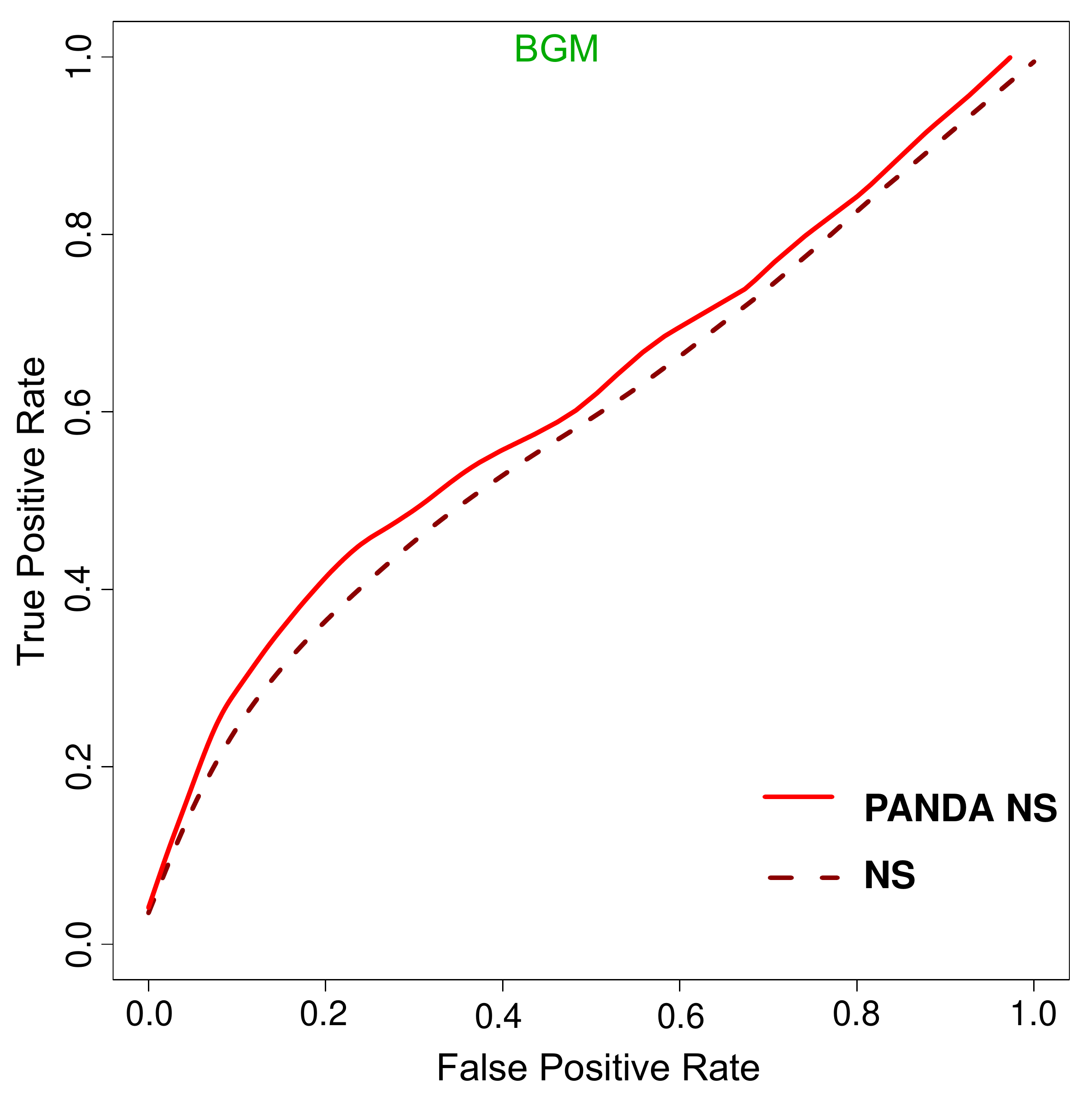}
\end{minipage}
\begin{minipage}{0.32\textwidth}
\includegraphics[width=1\linewidth]{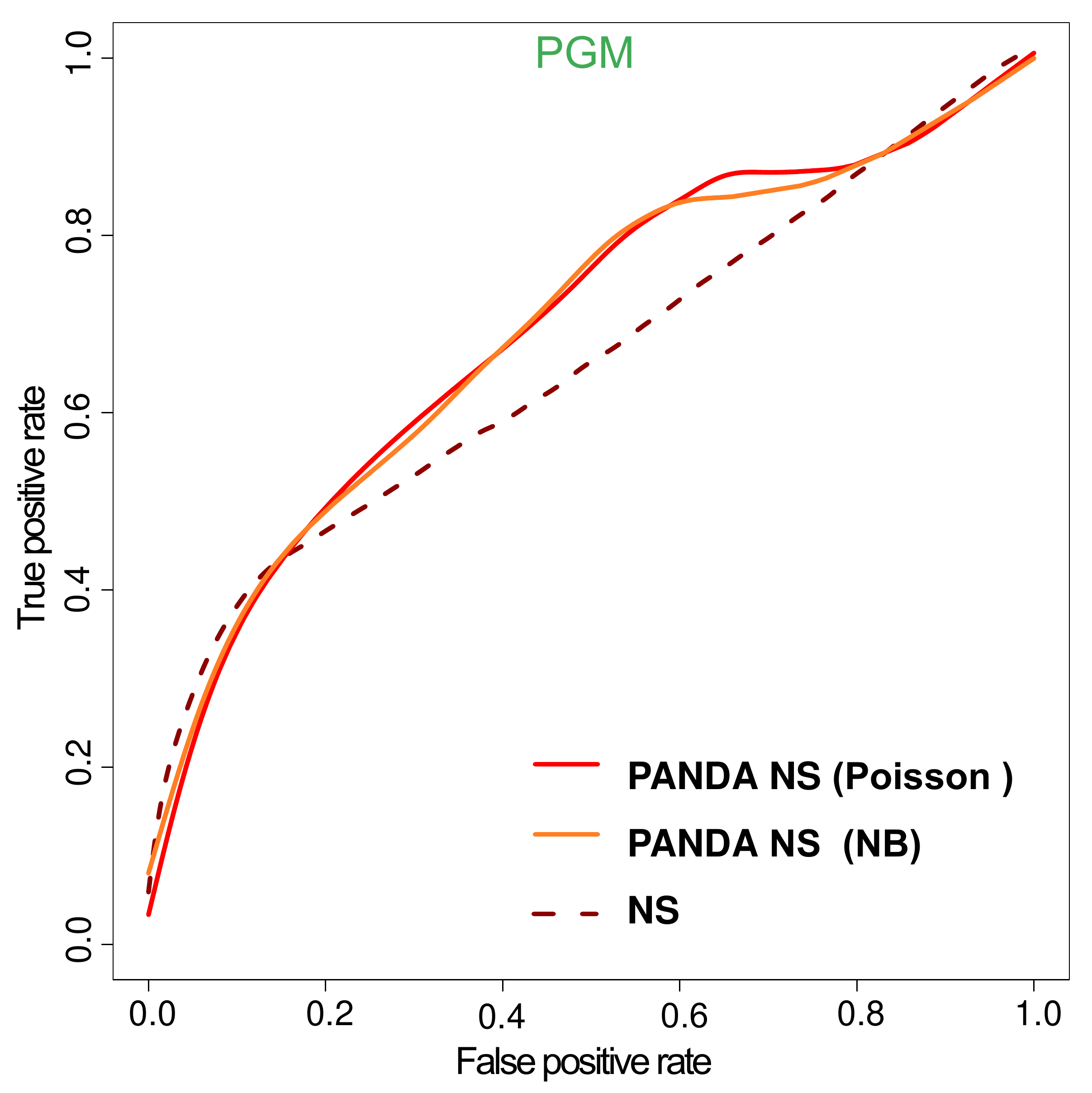}
\end{minipage}\\
\centering\textcolor{magenta}{\textbf{lattice-structured}}\\
\begin{minipage}{0.32\textwidth}
\includegraphics[width=1\linewidth]{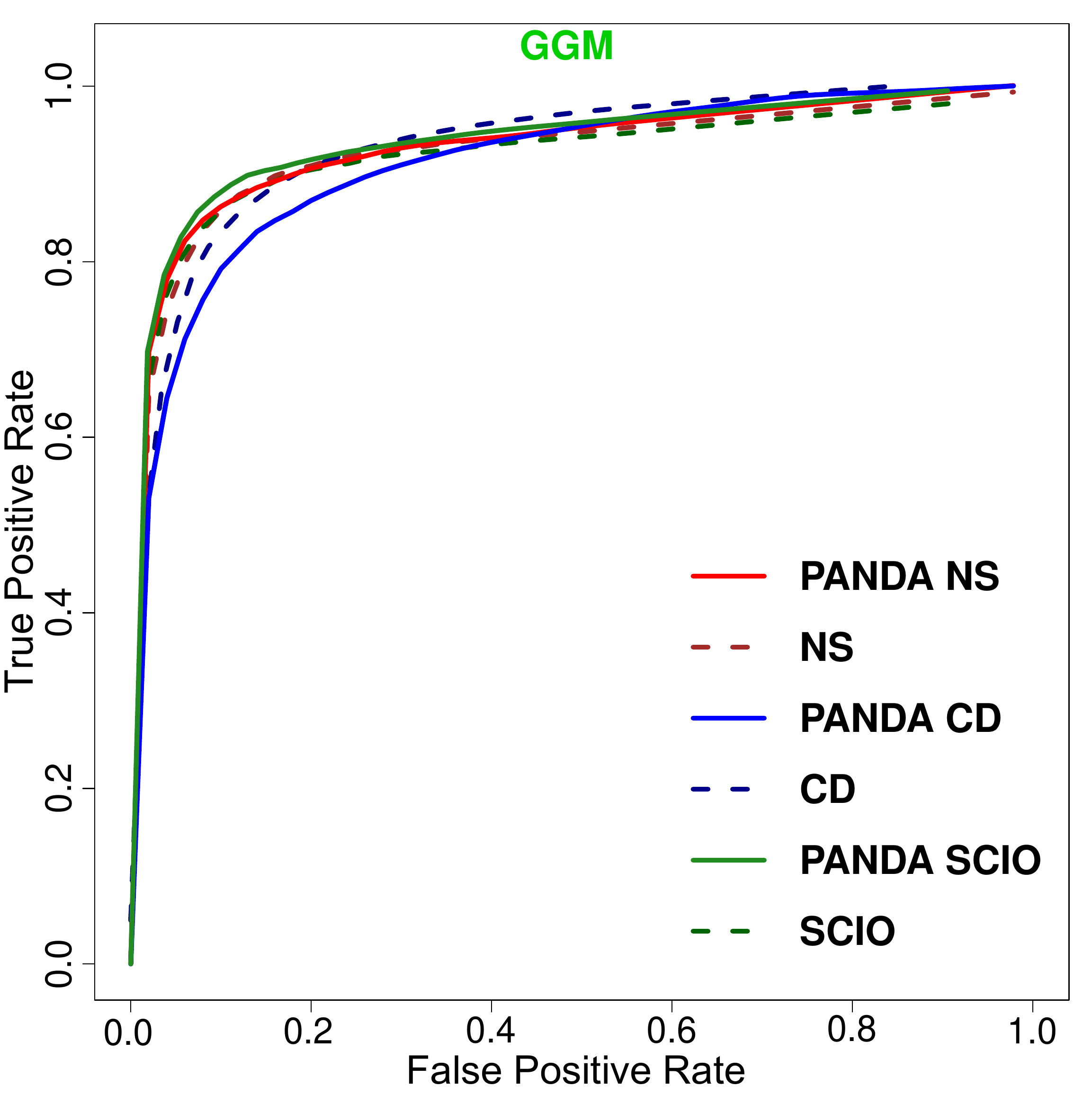}
\end{minipage}
\begin{minipage}{0.32\textwidth}
\includegraphics[width=1\textwidth]{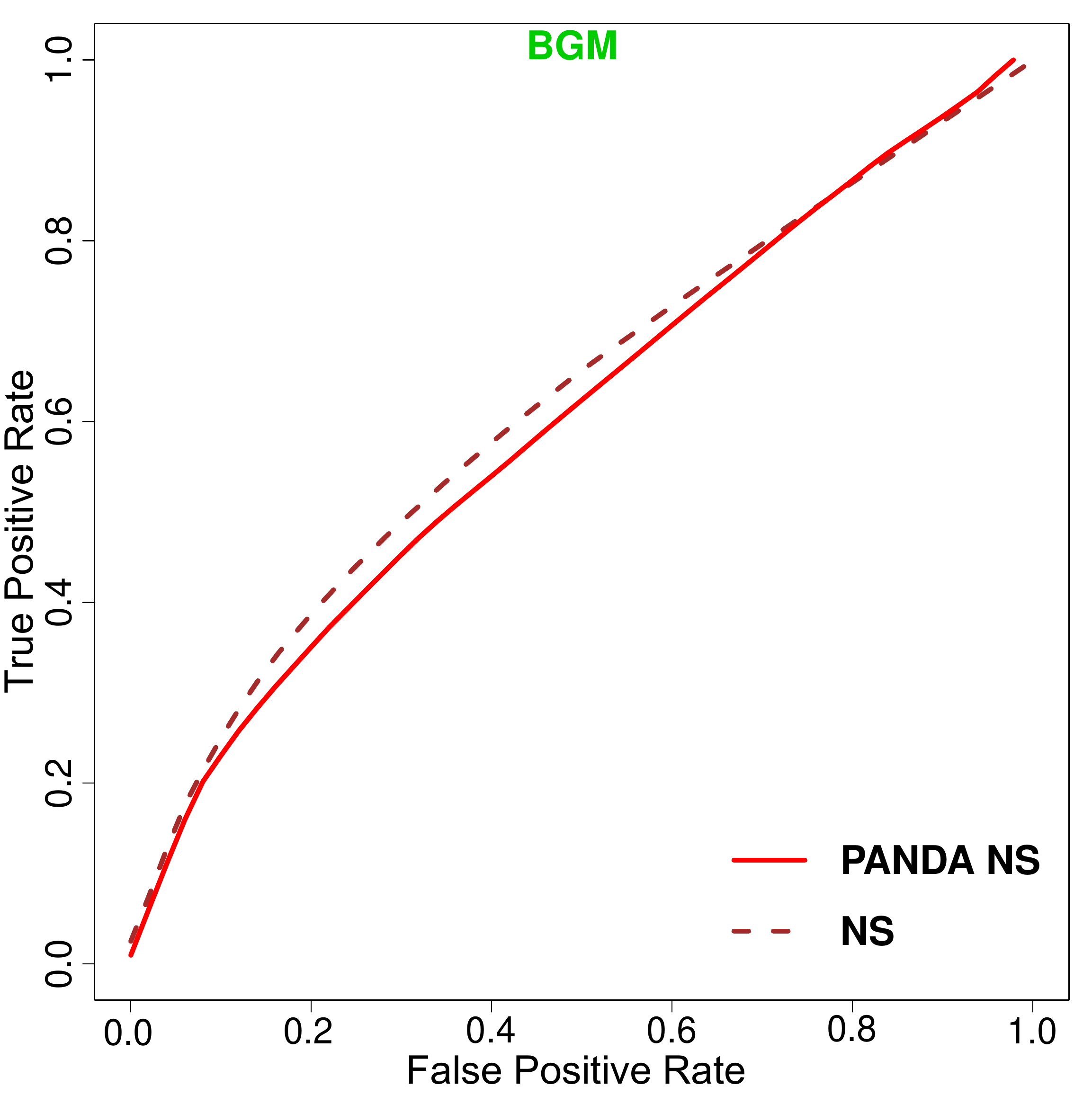}
\end{minipage}
\begin{minipage}{0.32\textwidth}
\includegraphics[width=1\linewidth]{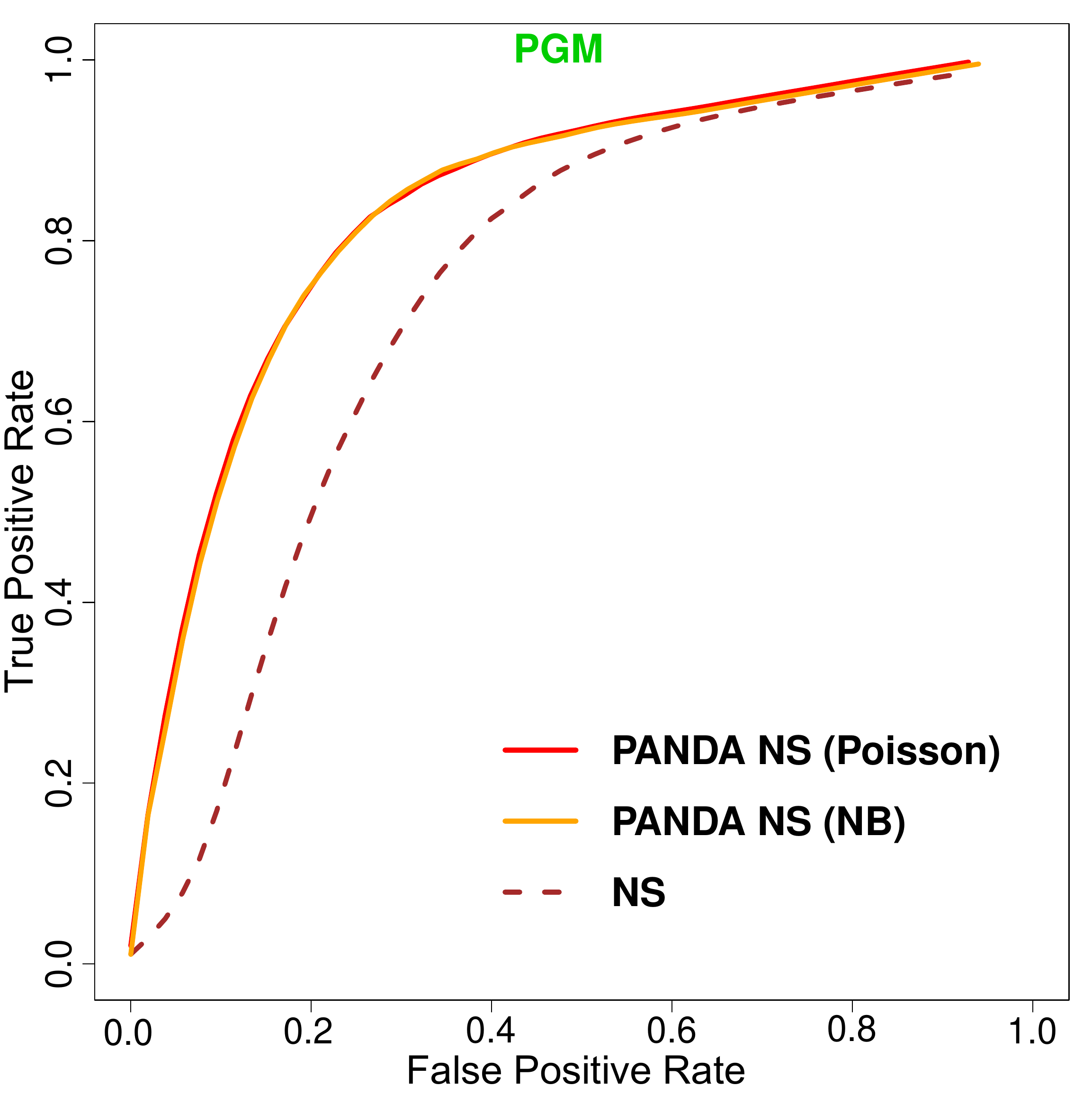}
\end{minipage}\\
\centering\textcolor{magenta}{\textbf{3-hub}}\\
\begin{minipage}{0.32\textwidth}
\includegraphics[width=1\linewidth]{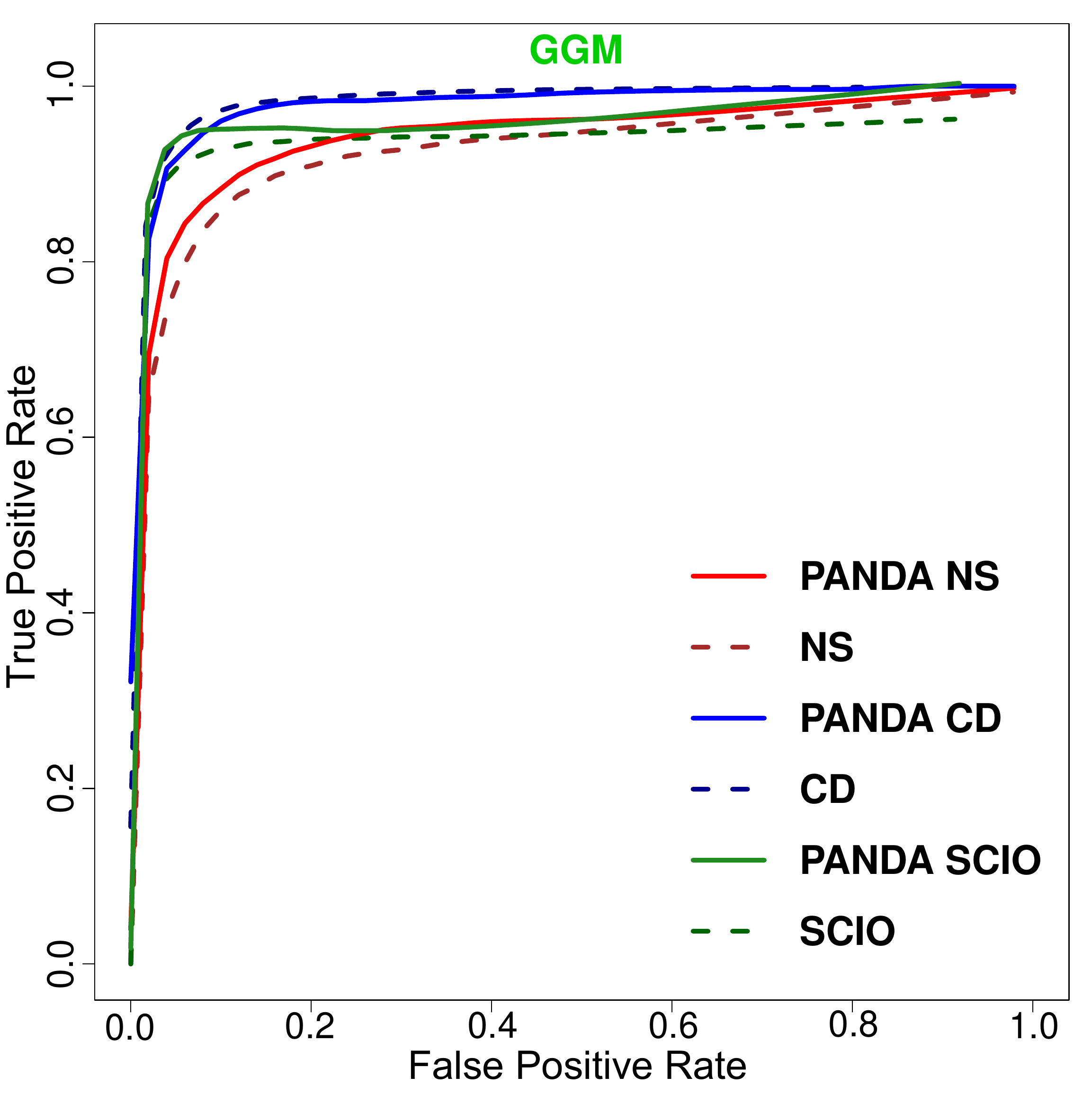}
\end{minipage}
\begin{minipage}{0.32\textwidth}
\includegraphics[width=1\textwidth]{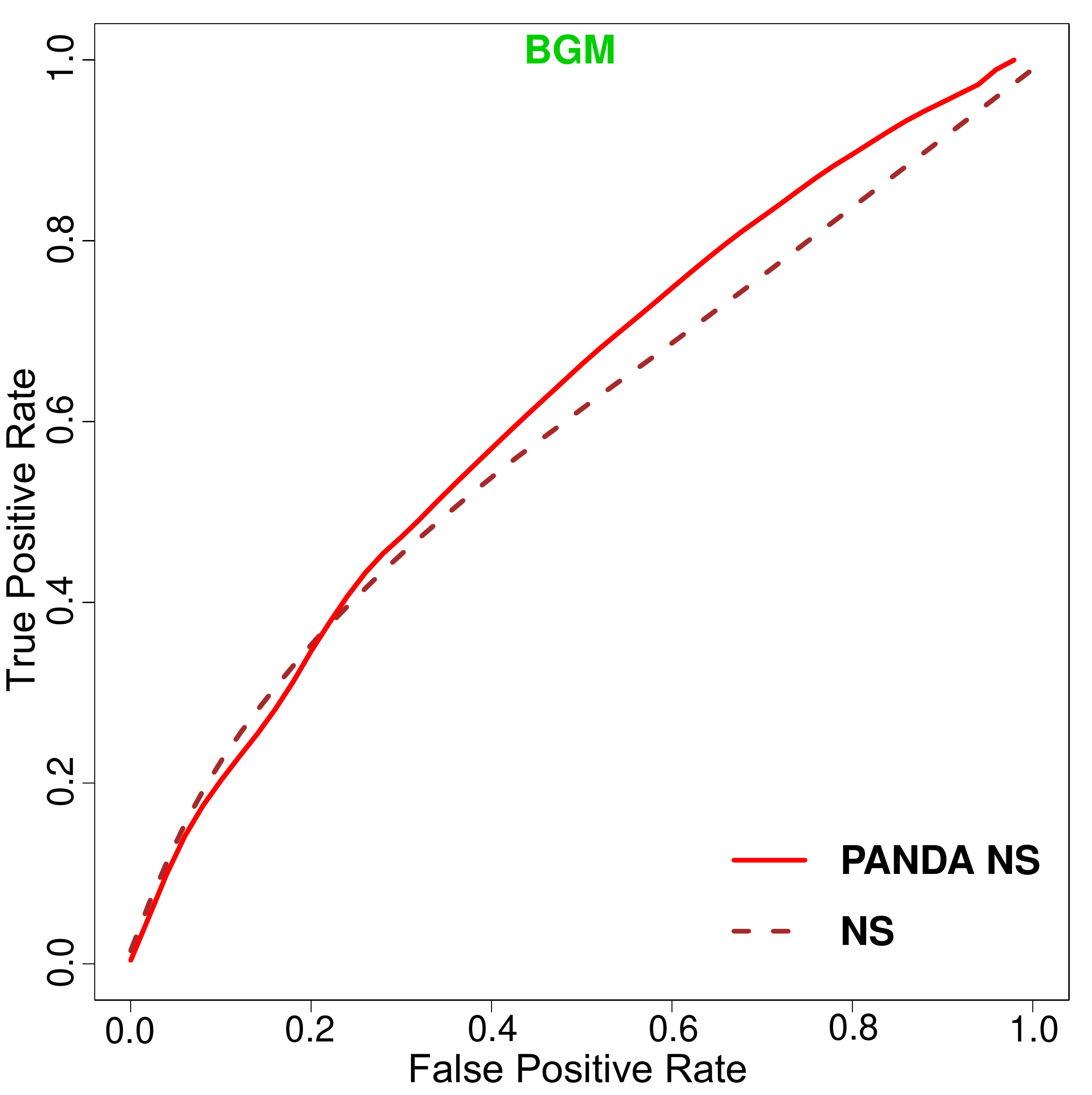}
\end{minipage}
\begin{minipage}{0.32\textwidth}
\includegraphics[width=1\linewidth]{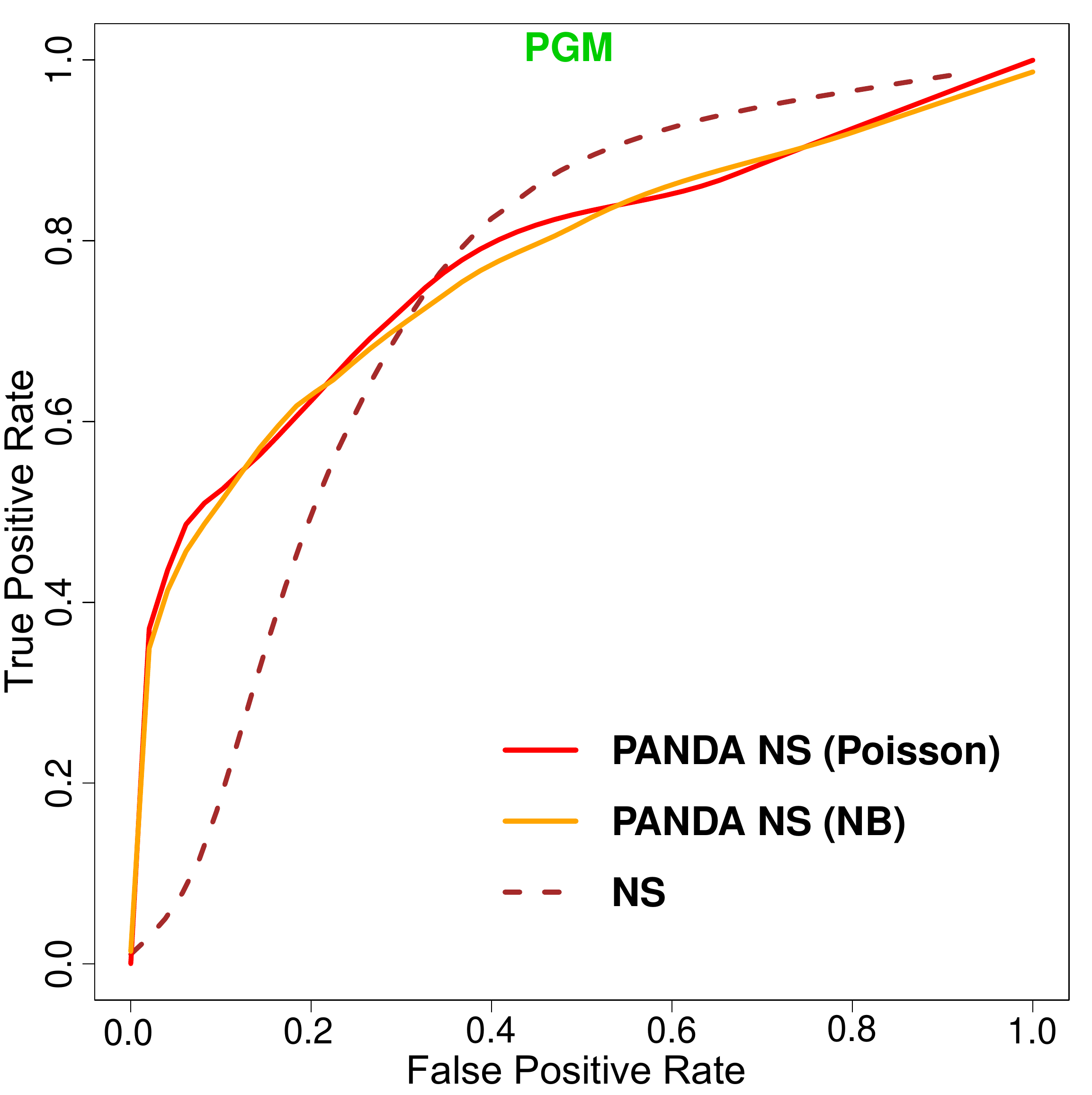}
\end{minipage}
\caption{ROC curves for non-zero edge identifications in GGM, BGM and PGM. The dash lines represent the constrained optimization and the solid lines represent PANDA.}\label{fig:single.simulation}
\end{figure}

We run 100 repetitions in each graph case with each structure, and calculated the false positive (FP) and  true positive (TP) rates at different $\lambda$ values, where ``positive'' is defined as the correct identification of a non-zero edge.  The ROC curves are depicted in Figure \ref{fig:single.simulation}. Overall, PANDA delivers either similar or superior ROC performance compared to the constrained optimization in all three graph types. The largest margin of superiority of PANDA over the constrained optimization is observed in the PGM, where PANDA is implemented in the framework of both Poisson regression and NB regression. PANDA has noticeably higher true positive rates than the constrained optimization when the false positive rate rangs 15\% to 80\% in the scale-free network, 0\% to 30\% in the 3-hub network, and  0\% to 100\% in lattice network. The ROC curves in the GGM also suggest the SCIO method (the green curves) seems to performs slightly better than the CD and NS approaches, likely because the regularization is directly imposed on the entries of the precision matrix in SCIO where both NS and CD regularize regression coefficients in linear regression, from which the precision matrix is calculated.
\subsection{Inference on GLM parameters via PANDA}\label{sec:sim2}

In this simulation, we investigate the inferential validity for the parameters $\bs{\beta}$ in GLMs based on the asymptotic distributions in Proposition \ref{prop:asymp.dist.UGM}.
We examine Gaussian  ($\sigma^2=1$), Poisson,  Bernoulli, Exponential (Exp), and Negative Binomial (NB) (number of failure was fixed at $r=5$) outcomes  with $p=30$ in each case. For the Gaussian and NB outcomes, the predictors were simulated from N$(0,1)$; for the Bernoulli, Exp, and Poisson outcomes, the predictors were simulated from Unif$(-3,3)$, Unif$(-1,2)$ and Unif$(-0.3,0.5)$, respectively. We examined three sample size scenarios $n=50,70,100$, with 200 repetitions in each simulation case. We used the lasso-type NGD to generate noise setting $n_e=n$ in logistic regression and $n_e=n/10$ in the other GLMs, and $\lambda n_e\in (1.5,7)$.  



In each repetition, we construct  the 95\% CIs for the 30 regression coefficients $\bs\beta$ ($21$ are non-zero and $9$ are zero) and examine the coverage probability (CP) and the CI width.  Tables \ref{tab:ci} presents the ranges of the CP and the corresponding CI width across the parameters, benchmarked against the post-lasso-selection inferential procedure \citep{CIJason2016,CIJonathan2017}, implemented using R package \texttt{selectiveInference}.   

\begin{table}[!htb]
\begin{center}
\resizebox{\columnwidth}{!}{
\begin{tabular}{l ccc l ccc}
\hline
& \multicolumn{3}{c}{PANDA} &&
\multicolumn{3}{c}{post-selection inference approach}\\
\cline{2-4}\cline{6-8}
outcome & n=50&n=70&n=100 &&n=50&n=70&n=100 \\
\hline
\multicolumn{8}{c}{\cellcolor{gray!15} $\beta=0$ }\\
\hline
&\multicolumn{7}{c}{\textbf{(min, max) CP (\%) among the 9 zero-valued $\beta$}}\\
\hline
Gaussian& (95.4, 96.8)&(95.6, 97.0)&(94.4, 98.6)&& N/A & N/A & N/A\\
Bernoulli &(100.0, 100.0)&(99.0, 100.0)&(95.8, 99.4)&& N/A & N/A & N/A\\
Exp  & (94.5, 98.1)&(96.2, 97.5)&(97.1, 98.3)&&-&-&-\\
Poisson & (94.8, 96.2)&(95.0, 97.6)&(96.6, 99.4)&&-&-&-\\
NB &(99.0, 99.8)&(99.6, 100)&(99.8, 100.0)&&-&-&-\\
\hline
&\multicolumn{7}{c}{\textbf{(min, max) CI width among the 9 zero-valued $\beta$}}\\
\hline
Gaussian  &(0.75, 0.77)&(0.56, 0.57)&(0.42, 0.43)&& N/A & N/A & N/A\\
Bernoulli & (11.6, 17.7)&(1.19, 1.44)&(0.85, 1.01)&& N/A & N/A & N/A\\
Exp & (1.15, 1.19)&(1.02, 1.05)&(0.99, 1.05)&& -&-&-\\
Poisson &(1.57, 1.61)&(1.03, 1.07)&(0.73, 0.76)&& -&-&-\\
NB  &(1.43, 1.48)&(1.08, 1.16)&(0.80, 0.85)&&-&-&-\\
\hline
\multicolumn{8}{c}{\cellcolor{gray!15} $\beta\ne0$ (21 $\beta$'s)}\\
\hline
&\multicolumn{7}{c}{\textbf{(min, max) CP (\%) among the 21 nonzero-valued $\beta$}}\\
\hline
Gaussian& (91.6, 95.8)&(92.8, 96.6)&(94.0, 96.2) &&(81.0, 83.6)&(92.2, 94.2)&(93.4, 95.0)\\
Bernoulli & (94.6, 100.0)&(79.6, 97.2)&(87.0, 98.2) &&(56.6, 75.2)&(66.6, 83.4)&(77.8, 88.0)\\
Exp &(92.8, 97.5)&(94.5, 97.3)&(96.0, 99.9) &&-&-&-\\
Poisson &(90.1 95.2)&(91.6, 96.0)&(93.6, 96.8) &&-&-&-\\
NB &(95.8, 99.2)&(98.6, 100)&(99.6, 100)&&-&-&-\\
\hline
& \multicolumn{7}{c}{\textbf{(min, max) CI width among the 21 zero-valued $\beta$}}\\
\hline
Gaussian &(0.80, 0.84)&(0.61, 0.62)&(0.46, 0.47)&& (28.4, 30.7)&(1.93, 2.07)&(1.22, 1.30)\\
Bernoulli &(16.7, 32.9)&(1.79, 2.50)&(1.28, 1.64)&& (20.7, 23.3)&(9.80, 11.0)&(4.26, 5.01)\\
Exp &(1.21, 1.25)&(1.00, 1.04)&(0.82, 0.89)&&-&-&-\\
Poisson &(1.63, 1.70)&(1.14, 1.18)&(0.85, 0.86)&&-&-&-\\
NB &(1.56, 1.66)&(1.30, 1.39)&(1.05, 1.09)&&-&-&-\\
\hline
\end{tabular}}
\resizebox{\columnwidth}{!}{\begin{tabular}{l}
\footnotesize NA: Package \texttt{selectiveInference} does not provide inference for $\beta$ whose estimate is 0 (that is, not selected \\
\footnotesize\hspace{16pt} by lasso). For these 9 null-valued $\beta$'s,\ many of them turned out not to be selected by lasso among the\\
\footnotesize\hspace{16pt}  200 repetitions. Therefore, no inferences are provided. \\
- \footnotesize Package \texttt{selectiveInference} only produces CIs for linear and logistic regression with the $l_1$ regularization.\\
\footnotesize\hspace{6pt} About 4 $\sim$ 18\% (the larger $n$ is, the higher the percentage) of the CIs have infinite lower/upper bounds, which are excluded in the summary.\\
\hline
\end{tabular}}
\end{center}
\caption{Range for the empirical CP and the width of 95$\%$ CI} \label{tab:ci}
\end{table}
When true $\beta=0$,  PANDA  maintains the nominal 95\% coverage for all the examined outcomes types and sample sizes. The  \texttt{selectiveInference} package does not provide inference for $\beta$ whose estimate is 0 (that is, not selected by lasso in the first place). For these 9 zero-valued $\beta$'s, many of them were not be selected by lasso among the 200 repetitions. Therefore, no inferences are provided.  When true $\beta\ne 0$, the CIs from PANDA have significantly better coverage than the post-selection procedure in most of the examined cases and  are similar in the rest. Specifically, PANDA maintains close to 95\% coverage in almost all cases and has some slight under-coverage for some $\beta$'s in logistic regression. The post-selection procedure experiences severe under-coverage in logistic regression for all $n$ and in linear regression when $n=50$. In terms of the efficiency of the inferences quantified by the CI width, PANDA yields much narrow CIs than the post-selection procedure in all cases, The CI width in the post-selection procedure can be as 30-fold higher than from the PANDA procedure. we also examined the larger $n_e$ cases ($n_e=2n$ in logistic regression and $n_e=n$ for the other GLMs), there was some under-coverage for both $\beta=0$ and $\beta\ne0$ (CP $\ge\sim90\%$ when  $\beta=0$; and $\ge\sim80\%$ when $\beta\ne0$), but improved as $n$ increased.

\section{Case study: the autism spectrum disorder data}\label{sec:case}
We apply PANDA to an autism spectrum disorder (ASD) data collected by the Dutch Association for Autism (Nederlandse Vereniging voor Autisme, NVA) and the Vrije Universiteit Amsterdam  \citep{begeer2013allemaal}. The dataset, available in the R package \texttt{mgm} \citep{haslbeck2016mgm},  contains 28 variables of various types (10 continuous, 7 categorical, and 11 count variables) from 3521 participants.  
In estimating the relationship among the variables, the continuous variables (nodes) were assumed to follow Gaussian distributions conditional on other nodes and  standardized before implementing PANDA. The count variables were assumed to follow Poisson distributions given other nodes.  For each of the 7 categorical variables, $k-1$ Bernoulli nodes were generated, where $k$ is the number of levels of the categorical variable.  All taken together, there were $p=35$ nodes used in the construction of the graph via PANDA. 

For the PANDA algorithm, we set tuning parameters $\gamma=1$ and $\sigma=0$ to obtain the lasso-type penalty. The extended BIC criterion \citep{chen2008extended, foygel2010extended, haslbeck2015structure} was used to choose $\lambda=0.667$. We used $n_e=1000,m=1$, $\tau_0=10^{-5},T=500$ and $r=500$. The computation took approximately 16 minutes in R (version 3.4.0) \citep{Rsoftware} on the Linux x86\_64 operating system.

\begin{figure}[!htb]
\includegraphics[scale=0.7]{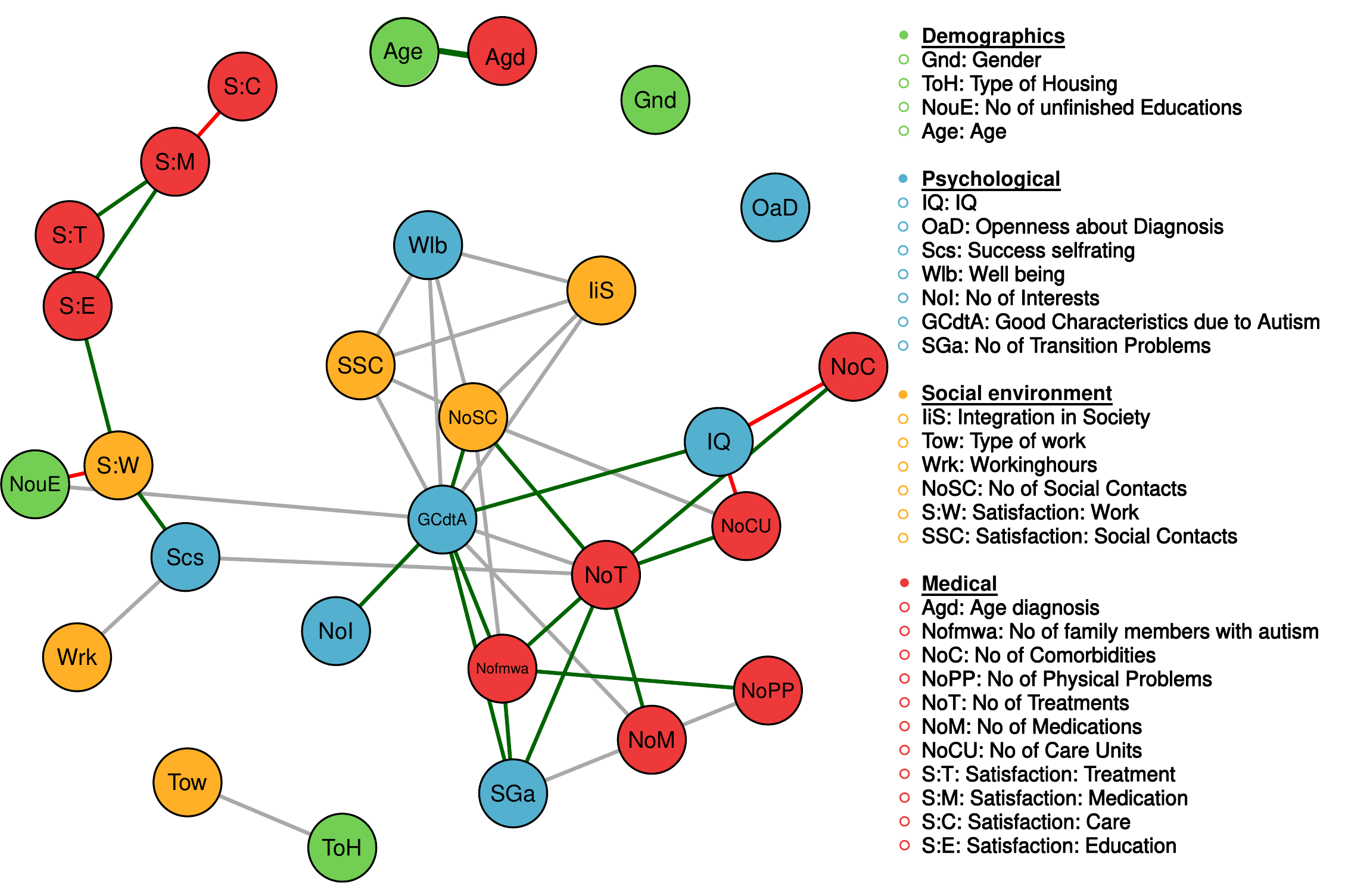}
\caption{\label{fig:asdqgraph}
Visualization of the UGM estimated via PANDA for the ASD data. Green edges indicate positive relationships, red edges indicate negative relationships, and gray edges indicate relationships involving categorical variables for which no sign is defined. The width of an edge is proportional to its weight. The colors of the nodes map to the four domains of the attributes.}
\end{figure}
Figure \ref{fig:asdqgraph} presents a visualization of the estimated UGM  via PANDA. The force-directed algorithm of \cite{fruchterman1991graph} is used to generate the graph layout. The 28 variables covered 4 domains, including demographics, social environment, diagnostic measurements and aspects of well-being. Figure \ref{fig:asdqgraph} suggests that `GCdtA'' (Good Characteristics due to Autism) is connected with multiple nodes from medical, social environment, and psychological domains  such as `NoSC'' (Number of Social Contacts), ``NoI'' (Number of Interests) and ``IQ'' (Intelligence Quotient). The connections indicate that the uniquely positive traits of autistic people connect with various aspects of their lives.  PANDA was also able to detect expected relationships among the nodes, such as the strong positive relationship between the present age of a participant (``Age'') and the age when the participant was diagnosed with autism (``Agd'').

\begin{figure}[!htb]
\includegraphics[width=0.58\textwidth, height=0.64\textheight, angle =90 ]{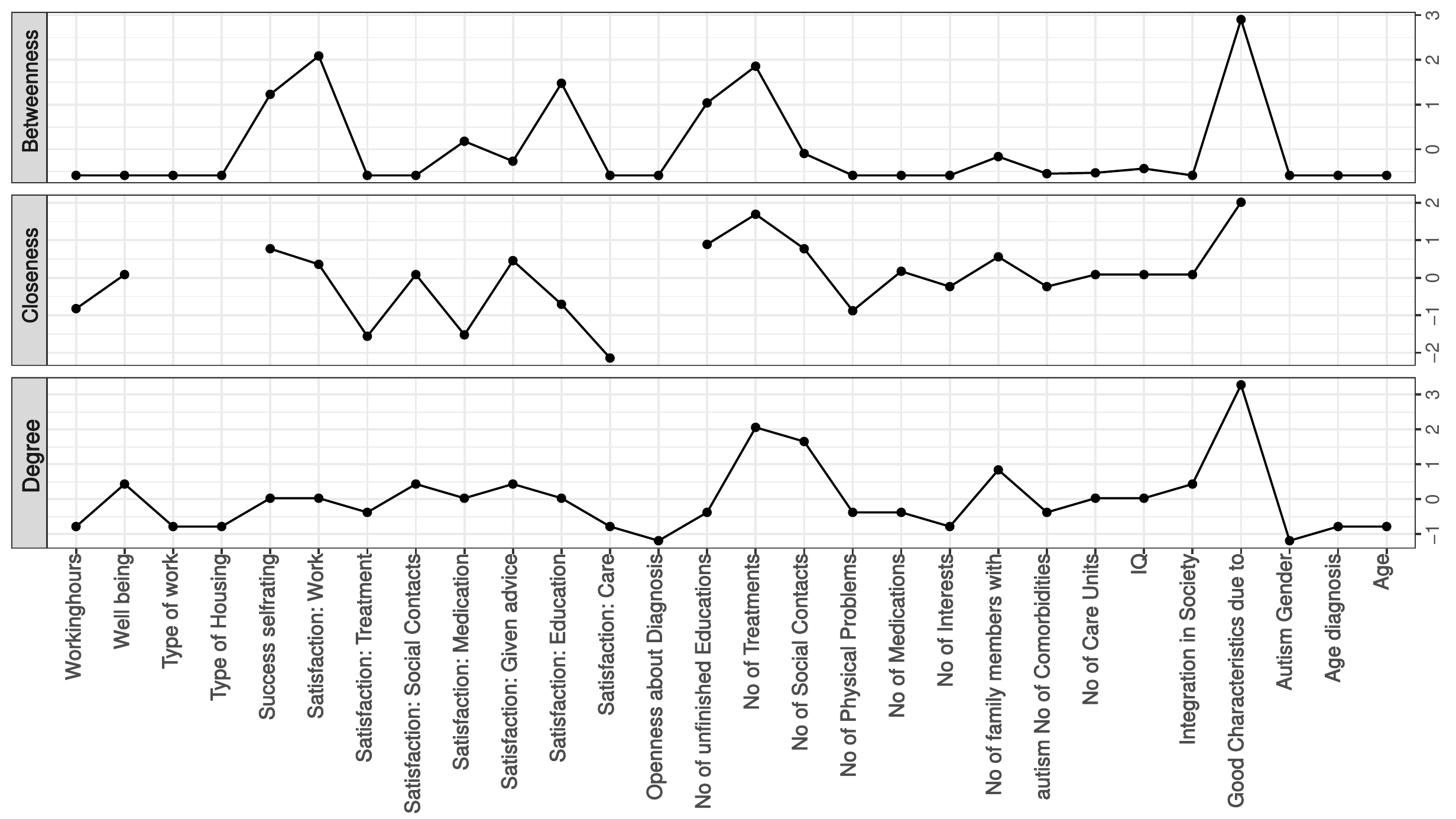}
\caption{Standardized centrality measures (degree, closeness and betweenness) for each node based on the UGM constructed by PANDA in the ASD data closeness is not defined for isolated nodes). The larger the measure, the more ``important'' the corresponding node is.}
\label{fig:asdcentr}
\end{figure}

In addition to the relationships among the variables, we can obtain some insights on the relative importance of those variables in the structure of the estimated graph. Figure \ref{fig:asdcentr} displays the standardized centrality measures (strength, closeness and betweenness) \citep{opsahl2010node} for each node. The results suggest that some variables, such as ``Good Characteristics due to Autism'', ``Satisfaction: Work'' and ``No of Social Contacts'' have relatively high centrality level whereas other variables, such as ``Openness about Diagnosis'', ``Type of Work'', ``Type of Housing'' and ``Gender'', had low centrality values, implying that those variables were not as important in the constitution of the network structure, which are not connected to the rest of the nodes, as given in Figure \ref{fig:asdqgraph}.


\section{Discussion} \label{sec:discussion}
PANDA is a regularization technique through noise augmentation. We have shown that PANDA can effectively regularize the construction of a UGM when the conditional distributions of the nodes given all other nodes in the graph are modeled by an exponential family. In the case of GGM, PANDA also offers counterparts to the  CD-based node-wise regression,  the SPACE approach,  the SCIO estimator, and the graphical ridge. We establish the Gaussian tail of the noise-augmented loss function  and the almost sure convergence to its expectation as $n_e$ or $m$ increases, which is a penalized loss function with the targeted regularizer, providing the theoretical justification for PANDA as a regularization technique and that the noise-augmented loss function is trainable. In the setting of GLMs, we propose an inferential procedure based on PANDA on top of variable selection. The simulation studies show PANDA offers non-inferior performance compared to some commonly-used graph construction methods. The case study also demonstrates the effectiveness of PANDA in constructing practically interpretable and meaningful mixed graph models.

Computationally, the PANDA algorithms are very straightforward to program; there is no need to code sophisticated optimization techniques as the algorithms can be built upon existing functions or procedures for running GLMs in any statistical software. In terms of the computational speed, a large $n_e$ or $m$ could slow down the computation, but a large $n_e$ usually leads to fast convergence with a small number of iterations. If PANDA is applied to obtain inferences in GLMs on top of variable selection, a small $n_e$ relative to $n$, with a relatively large $m$, should be used for the reasons mentioned in Sec \ref{sec:asymp.dist}. The presented PANDA algorithms in this paper calculate $\bar{\bs{\theta}}$, the  average of $m$ minimizers of $l(\Theta|\x,\e)$ from the latest $m$ iterations, so to leverage the existing software for running GLM and to maintain its computational advantage over constrained optimization that employs sophisticated optimization techniques.  Per Propositions \ref{prop:regularization} and \ref{prop:glmregularization}, one would take the average over $m$ noise-augmented loss function $l(\Theta|\x,\e)$ to yield a single minimizer $\hat{\bs{\theta}}$, which is the Monte Carlo version of $\E_{\e}(l_p(\bs{\theta}|\x,\e)$ as $m\rightarrow\infty$. When $m=1$, there is no difference between the two approaches, which can often used in practice when $n_e$ is set a large number. When $m>1$, we establish in Corollary \ref{cor:average} in the supplementary materials that $\bar{\bs{\theta}}$ and $\hat{\bs{\theta}}$  are first-order equivalent for large $m$ and $n_e$ for PANDA-NS in GGM, 
We also present simulation results in the linear regression and Poisson regression settings to  illustrate the similarity between  $\bar{\bs{\theta}}$ and $\hat{\bs{\theta}}$.

We have also extended the PANDA technique to simultaneously constructing multiple graphs that promotes the sparsity in each graph and similarity between graphs. Interested readers may refer to  \citet{pandam}.

\newpage
\section*{Appendix}
\appendix
\numberwithin{equation}{section}
\setcounter{equation}{0}
\section{Proof of Proposition \ref{prop:regularization}}\label{app:expectedloss}
The expectation of $l_p(\Theta|\x,\e)\!=\!\textstyle\sum_{i=1}^{n+n_e}\!\sum_{j=1}^{p}\!\!\left(\tilde{x}_{ij}\!-\!\sum_{k\ne j}\tilde{x}_{ik}\theta_{jk}\!\right)^2\!$ over the distribution of noise $\e$ is
\begin{align}
\E_\e (l_p(\Theta|\x,\e))\!&=\!\textstyle
\sum_{i=1}^{n}\sum_{j=1}^{p}\!\left(x_{ij}\!-\!\sum_{k\ne j}x_{ik}\theta_{jk}\!\right)^2\!\!+\!
\E_\e\!\left(\!\sum_{i=1}^{n_e}\sum_{j=1}^{p}\!\left(\!e_{ijj}\!-\!\sum_{k\ne j}e_{ijk}\theta_{jk}\right)^2\right)\label{eqn:Em}\\
&=\textstyle\sum_{i=1}^{n}\sum_{j=1}^{p}\left(x_{ij}-\sum_{k\ne j}x_{ik}\theta_{jk}\right)^2+
\sum_{i=1}^{n_e}\sum_{j=1}^{p}\E_\e \left(\sum_{k\ne j}e_{ijk}\theta_{jk}\right)^2\notag \\
&=\textstyle l(\Theta|{\x})+n_e\sum_{k\ne j}\theta_{jk}^2\mbox{V}(e_{ijk})\label{eqn:Ene}.
\end{align}
The above equations suggest there are (at least) two ways to approximate the second term in Eqn (\ref{eqn:Em}) in a Monte Carlo manner. The first approach is  straightforward, where the second term is approximated by
$\sum_{j=1}^{p}\lim_{m\rightarrow\infty} m^{-1}\sum_{t=1}^m\!
\sum_{i=1}^{n_e}\!\left(e^{(t)}_{ijj}\!-\!\sum_{k\ne j}e^{(t)}_{ijk}\theta_{jk}\right)^2$. The second approach is suggested by  Eqn  (\ref{eqn:Ene}). Under the constraint
$n_e\mbox{V}(e_{ijk})=O(1)$ and letting $n_e\rightarrow\infty$, the second term $n_e\sum_{k\ne j}\theta_{jk}^2\mbox{V}(e_{ijk})$ in  Eqn  (\ref{eqn:Ene}) is
$\textstyle
n_e\sum_{j=1}^{p}\sum_{k\ne j}\theta^2_{jk}\left(\lim_{n_e\rightarrow\infty} n_e^{-1}\!
\sum_{i=1}^{n_e}e^2_{ijk}\right)$.
The two ways of obtaining the expected penalty term in PANDA applied to GGM-NS (this proposition), UGM-NS (Proposition \ref{app:glmregularization}), and GGM-CD (Proposition \ref{prop:orderreg}).

\section{Proof of Proposition \ref{prop:glmregularization}}\label{app:glmregularization}
We first take the Taylor expansion of $l_p(\Theta|\x,\e)$, which is the negative log-likelihood, around $e_{ijk}\!=\!0$ for $i=1,\ldots,n_e$ and $k\!\ne\!j$, and then evaluate its expectation over the distribution of $e_{ijk}$.
\begin{align*}
&l_p(\Theta|\x,\e))=
l(\Theta|\x)+l_p(\Theta|\e)=
\textstyle l(\Theta|\x)) + \sum_{j=1}^{p}\!\sum_{i=1}^{n_e} l_{ij}(\bs\theta_j|\e_{i,-j})\\
=&\textstyle l(\Theta|\x)- \sum_{j=1}^{p}\!\sum_{i=1}^{n_e} \left(h_j(e_{ijj})+e_{ijj}\left(\theta_{j0}\!+\!\sum_{k\ne j}\theta_{jk}e_{ijk}\right)-B_j\left(\theta_{j0}\!+\!\sum_{k\ne j}\theta_{jk}e_{ijk}\right)\right)\\
=& l(\Theta|\x)\!+\!\textstyle\sum_{i=1}^{n_e}\sum_{j=1}^{p} \!l_{ij}(\bs\theta_j|\e_{i,-j})|_{\e_{i,-j}=0}\;\\
&\qquad\textstyle-
\sum_{i=1}^{n_e}\sum_{j=1}^{p}
\left\{e_{ijj}\!\sum_{k\ne j}\left(\theta_{jk}e_{ijk}\right)\!-\!
\sum_{k\ne j}(\theta_{jk}e_{ijk})\!
B'_j\left(\theta_{j0}\!+\!\sum_{k\ne j}\theta_{jk}e_{ijk}|_{e_{ijk=0}}\right)\right.\\
&\left.\textstyle\qquad\qquad\qquad\qquad-
\sum_{d=2}^{\infty}(d!)^{-1}\!\sum_{k\ne j}(\theta_{jk}e_{ijk})^d\!B^{(d)}_j\! \left(\theta_{j0}\!+\!\sum_{k\ne j}\theta_{jk}e_{ijk}|_{e_{ijk=0}}\right)\!\right\}\\
=&l(\Theta|\x)\!+\!C\!+\!
\textstyle\sum_{i=1}^{n_e}\!\sum_{j=1}^{p}\!\left[(B'_j(\theta_{j0})\!-\!e_{ijj})\sum_{k\ne j}(\theta_{jk}e_{ijk})\!+\!\sum_{d=2}^{\infty}(d!)^{-1}B^{(d)}_j(\theta_{j0})\!\sum_{k\ne j}(\theta_{jk}e_{ijk})^d\right],
\end{align*}
where $C=B_j(\theta_{j0})-\sum_{j=1}^{p}\sum_{i=1}^{n_e} \left(h_j(e_{ijj})+e_{ijj}\theta_{j0}\right)$, a constant independent of $\Theta$. The expectation of $l_p(\Theta|\x,\e)$  over the distribution of $e_{ijk}\sim N(0,\mbox{V}(e_{jk}))$ is
\begin{align}
\E_{\e}(l_p(\Theta|\x,\e))\!&=\!
l(\Theta|\x)\!+\!C\!+\!\textstyle
n_e\sum_{j=1}^{p}\!\left(\!\frac{1}{2}B''_j(\theta_{j0})\!\sum_{k\ne j}\!\theta^2_{jk}\mbox{V}(e_{jk})\right)\!+\! O\!\left(n_e\sum_{j=1}^{p}\!\sum_{k\ne j}\!\left(\theta_{jk}^4\E(e_{jk}^4)\right)\right)\notag\\
\!&=\!l(\Theta|\x)\!+\!C\!+\!\textstyle
n_e\sum_{j=1}^{p}\!\left(\!\frac{1}{2}B''_j(\theta_{j0})\!\sum_{k\ne j}\!\theta^2_{jk}\mbox{V}(e_{jk})\right)\!+\!O\!\left(\!n_e\!\sum_{j=1}^{p}\!\sum_{k\ne j}\!\left(\theta_{jk}^4\V^2(e_{jk})\!\right)\!\right)\notag\\
= l(\Theta|\x)&\textstyle+
n_e\sum_{j=1}^{p}\!\!\left(C_{1j}\!\sum_{k\ne j}\theta^2_{jk}\mbox{V}(e_{jk})\right)\!+\!C \!+\!O\!\left(\!n_e\!\sum_{j=1}^{p}\!\sum_{k\ne j}\!\left(\theta_{jk}^4\V^2(e_{jk})\!\right)\!\right),\label{eqn:appB}
\end{align}
where $C_{1j}=2^{-1}B''_j(\theta_{j0})$.  Similar to Proposition \ref{prop:regularization}, there are  at least 2 ways to realize the expectation in Eqn (\ref{eqn:appB}) empirically. The straightforward way is let $l_p(\Theta|\e)$ to be approximated by $\textstyle\lim_{m\rightarrow\infty} m^{-1}\sum_{t=1}^m \sum_{j=1}^{p}\sum_{i=1}^{n_e} l_{ij}(\bs\theta_j|\e^{(t)}_{i,-j})$.
The second approach, suggested by Eqn (\ref{eqn:appB}), under the constraint $n_e\mbox{V}(e_{jk})\!=\!O(1)$, is to let $n_e\rightarrow\infty$, and the second term in Eqn (\ref{eqn:appB})
$n_e\sum_{j=1}^pC_{1j}\sum_{k\ne j}\theta_{jk}^2\mbox{V}(e_{ijk})=
n_e\sum_{j=1}^{p}C_{1j}\sum_{k\ne j}\!\left(\!\theta^2_{jk}\lim_{n_e\rightarrow\infty} n_e^{-1}\!
\sum_{i=1}^{n_e}\e^2_{ijk}n_e\mbox{V}(e_{jk})\!\right)$. Between the two approaches, letting $n_e\rightarrow\infty\cap\left[n_e\mbox{V}(e_{jk})\!=\!O(1)\right]$  offers an additional benefit in that  $O\left(\sum_{j=1}^{p}\!\sum_{k\ne j}\!\left(\theta_{jk}^4n_e\V^2(e_{jk}))\!\right)\!\right)\!\rightarrow\! 0$ in Eqn (\ref{eqn:appB}); in other words, the second order Taylor approximation of $\E_{\e}(l_p(\Theta|\x,\e))$ is arbitrarily close to $\E_{\e}(l_p(\Theta|\x,\e))$, which does not hold when $m\rightarrow\infty$. Specifically, the big-$O$ term in Eqn (\ref{eqn:appB}), which is a function of $\Theta$, might adds other regularization onto $\Theta$ in addition to $\theta^2_{jk}\left(n_e\mbox{V}(e_{jk})\right)$ when it is non-ignorable.
If $X_j$ is of the same type and $\theta_{j0}=0$ for $j=1,\ldots,p$, then $B''_j(0)$ is the same across $j$ and $C_{1j}\!=\!B''(0)$ and Eqn (\ref{eqn:appB}) can be simplified to
\begin{equation}\label{eqn:appB2}
l(\Theta|\x) + \textstyle C_2\sum_{j=1}^{p}\sum_{k\ne j} \theta^2_{jk}\left(n_e\mbox{V}(e_{jk})\right)\!+\!C \!+\!O\left(\sum_{j=1}^{p}\!\sum_{k\ne j}\!\left(\theta_{jk}^4n_e\V^2(e_{jk}))\!\right)\!\right).
\end{equation}


\section{Proof of Proposition \ref{prop:orderreg}} \label{app:cholesky}
The expectation of $l_p(L,D|\x,\e)$ over the distribution of $\e$ for PANDA-CD is $\E(l_p(L,D|\x,\e))$
\begin{align*}
=&\textstyle\E\!\left(\sum_{j=1}^{p}\hat{\sigma}_j^{-2}\sum_{i=1}^{n}\left({x}_{ij}-\sum_{k=1}^{j-1}{x}_{ik}\theta_{jk}\right)^2+\sum_{j=1}^{p}\hat{\sigma}_j^{-2}\sum_{i=1}^{n_e}\left({e}_{ijj}-\sum_{k=1}^{j-1}{e}_{ijk}\theta_{jk}\right)^2\notag\right)\\
=&\textstyle l(L,D|\x)\!+\!\E\!\left(\!\sum_{i=1}^{n_e}\sum_{j=1}^{p}\hat{\sigma}_j^{-2}\left(\!\sum_{k=1}^{j-1}e_{ijk}^2\theta_{jk}^2\!\right)\!\right)=\textstyle l(L,D|\x)+\sum_{i=1}^{n_e}\sum_{j=1}^{p}\hat{\sigma}_j^{-2}\left(\sum_{k=1}^{j-1} \frac{\lambda \sigma_j^{2}}{|\theta_{jk}|^\gamma}\theta_{jk}^2\right)\\
=&
\textstyle l(L,D|\x)+\lambda n_e\sum_{j=1}^{p}\sum_{k=1}^{j-1}|\theta_{jk}|^{2-\gamma}.
 \end{align*}

\section{Proof of Proposition \ref{prop:additive}}\label{app:additive}
In NI, injected noise terms are additive to the observed data without changing the dimensionality of the original data ($n\times p$). We establish the equivalence between PANDA and NI in their expected regularization effects for each graph type separately.

\subsection{NS for GGM and UGM}
Denote the noise injected data by $\tilde{\x}_{ik}=\x_{ik}+\e_{ik}$ (for $k\ne j$) when regressing $X_j$ on $\X_{-j}$, where $e_{ijk}$ is drawn from a NGD in Eqns (\ref{eqn:bridge}) to (\ref{eqn:scad}), and $\tilde{x}_{ij}=x_{ij}$.

For GGM, the expectation of the loss function based on the noise injected data  is
\begin{align*}
\E_\e(l_p(\Theta|\tilde{\x}))=&
\textstyle \E_\e\left(\sum_{i=1}^{n}\sum_{j=1}^{p}\left(\tilde{x}_{ij}-\sum_{k\ne j}\tilde{x}_{ik}\theta_{jk}\right)^2\right)\\
=&\textstyle\sum_{i=1}^{n}\sum_{j=1}^{p}\left(x_{ij}-\sum_{k\ne j}x_{ik}\theta_{jk}\right)^2\!\!+\!
 \E_\e\left(\sum_{i=1}^{n}\sum_{j=1}^{p}\left(e_{ijj}-\sum_{k\ne j}e_{ijk}\theta_{jk}\right)^2\right)\\
=&\textstyle l(\Theta|{\x})+\sum_{i=1}^n\sum_{j=1}^{p}\sum_{k\ne j}\theta_{jk}^2\mbox{V}(e_{ijk})
\end{align*}

For non-Gaussian UGM in general, we first take the Taylor expansion of $l_p(\Theta|\tilde{\x})$, the noise-augmented negative log-likelihood, around $\tilde{x}_{ik}={x}_{ik}$, then evaluate its expectation over the distribution of $e_{jk}$ for $k\ne j=1,\ldots,p$.
\begin{align*}
&\textstyle \E_\e(l_p(\Theta|\tilde{\x}))
=-\E_\e\! \sum_{j=1}^{p}\! \sum_{i=1}^{n} \left(h_j(\tilde{x}_{ij})+{x}_{ij}\left(\!\theta_{j0}\!+\!\!\sum_{k\ne j}\theta_{jk}\tilde{x}_{ik}\!\right)\!-\!B_{j}\left(\!\theta_{j0}\!+\!\!\sum_{k\ne j}\theta_{jk}\tilde{x}_{ik}\right)\right)\\
=&\textstyle-\! \sum_{j=1}^{p}\! \sum_{i=1}^{n} \left(h_j({x}_{ij})+{x}_{ij}\!\left(\!\theta_{j0}\!+\!\!\sum_{k\ne j}\theta_{jk}\tilde{x}_{ik}\!\right)\!-\E_\e\left[ B_{j}\left(\!\theta_{j0}\!+\!\sum_{k\ne j}\theta_{jk}\tilde{x}_{ik}\right)\right]\right)\\
=& \textstyle l(\Theta|\x)+\sum_{j=1}^{p}\! \sum_{i=1}^{n} \sum_{d=2}^{\infty}(d!)^{-1}\E_\e B_j^{(d)}\left(\!\theta_{j0}\!+\sum_{k\ne j}\theta_{jk}{x}_{ik}\right)\!\sum_{k\ne j}(\theta_{jk}e_{ijk})^d\\
=& \textstyle l(\Theta|\x)+2^{-1}\sum_{j=1}^{p}\! \sum_{i=1}^{n} B_j^{(2)}\left(\!\theta_{j0}\!+\!\sum_{k\ne j}\theta_{jk}{x}_{ik}\right)\!\sum_{k\ne j}\theta_{jk}^2\mbox{V}(e_{ijk})+nO\!\left(\sum_{j=1}^{p}\bs\theta_j^4\!\cdot\!\E(\e_j^4)\right).
\end{align*}
The above equation suggests the targeted regularization with NI in UGMs with non-Gaussian nodes may be only approximated by the second order Taylor expansion when the residual term $nO\!\left(\sum_{j=1}^{p}\bs\theta_j^4\!\cdot\!\E(\e_j^4)\right)$ is ignorable.

\subsection{GGM-CD}
Let $\tilde{\x}_{ik}=\hat{\sigma}^{-1}_j(\x_{ik}+\e_{ik})$ (for $k\ne j$)in the $j$-th regression, where $\e_{ik}$ is drawn from a NGD in Eqns (\ref{eqn:bridge}) to (\ref{eqn:scad}) and $\hat{\sigma}^{-1}_j$ varies by iteration. $\tilde{\x}_{ij}=x_{ij}$. $\tilde{\x}_{ij}$.
\begin{align*}
&\E_\e(l_p(L,D|\tilde{\x}))=
\textstyle\E_\e\left(n\!\sum_{j=1}^{p}\log \sigma_{j}^2+ \sum_{i=1}^{n}\!\sum_{j=1}^{p}\!\sigma_{j}^{-2}\left(\!\tilde{x}_{ijj}-\!\sum_{k=1}^{j-1}\tilde{x}_{ijk}\theta_{jk}\!\right)^2\right)\\
&=\textstyle\E_\e\!\left(\!\!n\!\sum_{j=1}^{p}\log \sigma_{j}^2\!+\!\! \sum_{i=1}^{n}\!\sum_{j=1}^{p}\!\sigma_{j}^{-2}\!\left(\!x_{ijj}\!-\!\sum_{k=1}^{j-1}x_{ijk}\theta_{jk}\!\right)^2
\!\!\!+\!\!\sum_{i=1}^{n}\!\sum_{j=1}^{p}\!\sigma_{j}^{-2}\!\left(\!e_{ijj}\!-\!\sum_{k=1}^{j-1}\!e_{ijk}\theta_{jk}\!\right)^2\right)\\
&= \textstyle l(L,D|\x)\!+\!\E_\e\!\left(\!\sum_{i=1}^{n}\!\sum_{j=1}^{p}\!\sigma_{j}^{-2}\!\left(\!e_{ijj}\!-\!\sum_{k=1}^{j-1}\!e_{ijk}\theta_{jk}\!\right)^2\right)\!
=\! l(L,D|\x)\!+\!\sum_{i=1}^n\sum_{j=1}^{p}\!\sum_{k\ne j}\theta_{jk}^2\!\mbox{V}(e_{ijk})
\end{align*}

\subsection{GGM-SCIO}
When estimating $\bs{\theta}_j$, let $\tilde{\x}_i=\x_i+\sqrt{2}\e_i$, where $e_{ij}\equiv0$ for centralized $\X$ and $e_{ik}$ (for $k\ne j$) is drawn from a NGD in Eqns (\ref{eqn:bridge}) to (\ref{eqn:scad}).  The loss function in GGM-SCIO in the noise-injected data $\tilde{\x}$   is
$l_p(\bs{\theta}_{j}|\tilde{\x})=\frac{1}{2}\bs{\theta}_{j}^t
\left(n^{-1}\sum_{i=1}^{n}\tilde{\x}_i\tilde{\x}_i^t\right)\bs{\theta}_{j}-\mathbf{1}_j\bs{\theta}_{j}$, and its expectation is
\begin{align*}
&\textstyle\E_\e(l_p(\bs{\theta}_{j}|\tilde{\x}))
=\frac{1}{2}\textstyle\bs{\theta}_{j}^t
\left(n^{-1}\sum_{i=1}^{n}{\x}_i{\x}_i^t +2n^{-1}\E_\e\left(\sum_{i=1}^{n}\e_i\e_i^t\right)\right)\bs{\theta}_{j}-\mathbf{1}_j\bs{\theta}_{j}\\
&\textstyle=\frac{1}{2}\bs{\theta}_{j}^t\hat{\Sigma}\bs{\theta}_{j}-\1_j\bs{\theta}_{j}+\textstyle \bs{\theta}_j^t \mbox{V}(\e_{ik})\bs{\theta}_j
\end{align*}

\subsection{Graphical ridge}
\begin{center}
$\E_\e({l_p}(\Omega|\tilde{\x}))
\!=\!\E_{\e}\!\left( -\log(|\Omega|)\!+\!(1/2)\sum_{i=1}^{n}\tilde{\x}_i^T\Omega\tilde{\x}_i\right)$
\begin{align*}
&=\!\textstyle-\log(|\Omega|)+\E_\e\!\left((1/2)\sum_{i=1}^{n}(\x_i\!+\!\e_i)^T\Omega(\x_i\!+\!\e_i)\right)\\
&=\textstyle-\log(|\Omega|)+(1/2)\sum_{i=1}^{n}\x_i^T\Omega\x_i+(1/2)\E_\e\left(\sum_{i=1}^{n}\e_i^T\Omega\e_i\right)\\
&=\textstyle-\log(|\Omega|)+(1/2)\sum_{i=1}^{n}\x_i^T\Omega\x_i+(\lambda/2)\sum_{j,k=1}^{p}\omega_{jk}^2.
\end{align*}
\end{center}

\section{Proof of Theorem \ref{thm:pelf2nmelf}}\label{app:pelf2nmelf}
We prove Theorem \ref{thm:pelf2nmelf} for GGM, PGM, EGM, NBGM, and BGM, respectively. WLOG, we use the bridge-type noise $e_{ijk}\sim N(0,\lambda|\theta|^{-\gamma})$ to demonstrate the proofs, which can be easily extended to other types of noises. Prior to the proof of Theorem \ref{thm:pelf2nmelf}, we state a theoretical result in Claim \ref{cla:consistineq}, on which the subsequent proofs rely on.
\begin{cla}\label{cla:consistineq} If $l_p(\Theta|\x,\e)$ and $l_p(\Theta|\x)$ are convex functions w.r.t. $\Theta$ and share the same parameter space $\bs\Theta$, then
$$\left|\inf\limits_{\Theta}l_p(\Theta|\x,\e)-\inf\limits_{\Theta}l_p(\Theta|\x)\right|\leq \sup\limits_{\Theta}\left|l_p(\Theta|\x,\e)-l_p(\Theta|\x)\right|$$
\end{cla}
\textbf{Proof of  Claim \ref{cla:consistineq}}:  Since both $\inf\limits_{\Theta}l_p(\Theta|\x,\e)$ and $\inf\limits_{\Theta}l_p(\Theta|\x)$ are convex optimization problems, each has a global optimum, denoted by $\hat{\Theta}$ and $\tilde{\Theta}$, respectively;  thus $\left|\inf\limits_{\Theta}l_p(\Theta|\x,\e)\!-\!\inf\limits_{\Theta}l_p(\Theta|\x)\right|$ $= \left|l_p(\hat{\Theta}|\x,\e)-l_p(\tilde{\Theta}|\x)\right|$. Consider the following two scenarios,
	
i). if $l_p(\hat{\Theta}|\x,\e) \geq l_p(\tilde{\Theta}|\x)$, then $l_p(\tilde{\Theta}|\x,\e) \geq l_p(\hat{\Theta}|\x,\e) \geq l_p(\tilde{\Theta}|\x)$ and
$$\left|l_p(\hat{\Theta}|\x,\e)-l_p(\tilde{\Theta}|\x)\right| = l_p(\hat{\Theta}|\x,\e)-l_p(\tilde{\Theta}|\x) \leq l_p(\tilde{\Theta}|\x,\e)-l_p(\tilde{\Theta}|\x)=\left| l_p(\tilde{\Theta}|\x,\e)-l_p(\tilde{\Theta}|\x)\right|$$
ii). if $l_p(\hat{\Theta}|\x,\e) < l_p(\tilde{\Theta}|\x)$, then $l_p(\hat{\Theta}|\x,\e) < l_p(\tilde{\Theta}|\x) < l_p(\hat{\Theta}|\x)$ and
$$\left|l_p(\hat{\Theta}|\x,\e)-l_p(\tilde{\Theta}|\x)\right| = l_p(\tilde{\Theta}|\x) -l_p(\hat{\Theta}|\x,\e) \leq l_p(\hat{\Theta}|\x) -l_p(\hat{\Theta}|\x,\e) =\left| l_p(\hat{\Theta}|\x,\e)-l_p(\hat{\Theta}|\x)\right|.$$
All taken together,
$\left|l_p(\hat{\Theta}|\x,\e)\!-\!l_p(\tilde{\Theta}|\x)\right|\!\leq\!\max\!\left(\left| l_p(\tilde{\Theta}|\x,\e)\1-\!l_p(\tilde{\Theta}|\x)\right|, \left| l_p(\hat{\Theta}|\x,\e)-l_p(\hat{\Theta}|\x)\right|\right)$ \\ $\leq \sup\limits_{\Theta}\left|l_p(\Theta|\x,\e)-l_p(\Theta|\x)\right|$.

\subsection{GGM} 
For Gaussian nodes, the regularization effects of $n_e\rightarrow\infty$ and $m\rightarrow\infty$ are the same.  
The loss function  upon convergence is
$$\textstyle \bar{l}_p(\Theta|\x,\e)\!=\!\sum_{i=1}^{n}\!\sum_{j=1}^{p}\!\left(\!x_{ij}\!-\!\sum_{k\ne j}x_{ik}\theta_{jk}\!\right)^2\!+m^{-1}\!\sum_{t=1}^{m}\!\sum_{i=1}^{n_e}\sum_{j=1}^{p}\!\left(\!e_{ijj}\!-\!\sum_{k\ne j}e^{(t)}_{ijk}\theta_{jk}\!\right)^2.$$
Since $e^{(t)}_{ijk}=\sqrt{\lambda|\theta|^{-\gamma}}z^{(t)}_{ijk}$, where $z^{(t)}_{ijk}\sim N(0,1)$. Therefore,
\begin{align}
\bar{l}_p(\Theta|\x,\e)=&\textstyle l(\Theta|\x)\!+m^{-1}\!\sum_{t=1}^{m}\sum_{i=1}^{n_e}\sum_{j=1}^{p}\left(\sum_{k\ne j} \frac{\lambda\theta_{jk}^2}{|{\theta_{jk}}|^\gamma}z^{(t)2}_{ijk}+2\sum_{l<v\neq j}\frac{\lambda\theta_{jl}\theta_{jv}}{|\theta_{jv}\theta_{jl}|^{\frac{\gamma}{2}}}z^{(t)}_{ijl}z^{(t)}_{ijv}\right)\notag\\
=&\textstyle l(\Theta|\x)\!+\!m^{-1}\!\!\sum_{t=1}^{m}\sum_{j=1}^{p}\!\sum_{k\ne j}\!\!\left(\!\frac{\lambda\theta_{jk}^2}{|{\theta_{jk}}|^\gamma}\!\sum_{i=1}^{n_e}\!z_{ijk}^{(t)2}\!\!\right)\notag\\
&\textstyle +2m^{-1}\sum_{t=1}^{m}\!\sum_{j=1}^{p}\!\sum_{l<v\neq j}\!\left(\!\frac{\lambda\theta_{jl}\theta_{jv}}{|\theta_{jv}\theta_{jl} |^{\frac{\gamma}{2}}}\!\sum_{i=1}^{n_e}\! z^{(t)}_{ijl}z^{(t)}_{ijv}\!\!\right).\notag
\end{align}
Since $\sum_{i=1}^{n_e}\!z_{ijk}^{(t)2}\sim \Gamma\left(\frac{n_e}{2},2\right)$ and
$\Gamma\left(\frac{n_e}{2},2\right)
\approx N(n_e,2n_e)=n_e+ (2n_e)^{1/2}N(0,1)=n_e+(2n_e)^{1/2}z_1$ as  $n_e\rightarrow\infty$;
$\sum_{i=1}^{n_e}\! z^{(t)}_{ijl}z^{(t)}_{ijv}\sim  \Gamma\left(\frac{n_e}{2},2\right)-\Gamma\left(\frac{n_e}{2},2\right)\approx N(0,4n_e)= 2n_e^{1/2}N(0,1)=2n_e^{1/2}z_2$ as  $n_e\rightarrow\infty$, where $z_1\sim N(0,1)$ and $z_2\sim N(0,1)$.  Therefore, the distribution of  $\bar{l}_p(\Theta|\x,\e)$ can be approximated by
\begin{align}
&\textstyle l(\Theta|\x)\!+\sum_{j=1}^{p}\sum_{k\ne j}n_e\lambda|\theta_{jk}|^{2-\gamma}\label{app:asymptgauss}\\
&+\sum_{j=1}^{p}\sum_{k\ne j}\!\left(\!n_e\lambda |\theta_{jk}|^{2-\gamma} 2^{1/2}n_e^{-1/2}\!\left(\!\frac{1}{m}\!\sum_{t=1}^{m}z_1^{(t)}\!\right)\!\right)\!+\!
\sum_{j=1}^{p}\!\sum_{l<v\neq j}\!\!\left(\!\frac{\lambda n_e\theta_{jl}\theta_{jv}}{|\theta_{jv}\theta_{jl}|^{\frac{\gamma}{2}}}\! \left(\!2n_e^{-1/2}\!\right)\!\left(\! \frac{1}{m}\!\sum_{t=1}^{m} z_2^{(t)}\!\right)\!\right)\notag\\
=&l_p(\Theta|\x)\!+(mn_e)^{-1/2}C_1N(0,1)\mbox{ where } C_1\!=\!n_e\lambda \bigg(2\sum_{j=1}^{p}  \bigg|\bigg|\bigg(\frac{\bs\theta_{j,-j}}{|{\bs\theta_{j,-j}}|^{\frac{\gamma}{2}}}\bigg)\bigg(\frac{\bs\theta_{j,-j}}{|{\bs\theta_{j,-j}}|^{\frac{\gamma}{2}}}\bigg)^T\bigg|\bigg|_2^2\bigg)^{1/2}, \label{app:asygauss1}
\end{align}
where $l_p(\Theta|\x)=l(\Theta|\x)\!+\sum_{j=1}^{p}\sum_{k\ne j}n_e\lambda|\theta_{jk}|^{2-\gamma}=\E_{\e}(l_p(\Theta|\x,\e))$ per Appendix \ref{app:expectedloss} and   Proposition \ref{prop:regularization}.Exactly the same Eqn (\ref{app:asygauss1}) can be obtained by letting $m\rightarrow\infty$ rather than $n_e\rightarrow\infty$.

Per the strong law of large numbers (LLN), Eqn (\ref{app:asygauss1}) suggests $\bar{l}_p(\Theta|\x,\e)$ converges almost surely to its mean for all $\Theta\in\bs\Theta$ as  $m\rightarrow\infty$ or $n_e\rightarrow\infty$ (with $n_e\lambda=O(1)$),  assuming $|\theta_{jk}|$ belongs to a compact parameter space and is bounded by $B$.  Consequently, $\sup\limits_{\Theta}\left|\bar{l}_p(\Theta|\x,\e)-l_p(\Theta|\x)\right|\overset{\mbox{a.s.}}{\longrightarrow}0$
as $m\rightarrow\infty \mbox{ or } n_e\rightarrow\infty$. Per Claim \ref{cla:consistineq}, 
$\inf\limits_{\Theta}\bar{l}_p(\Theta|\x,\e)\overset{\mbox{a.s.} }{\longrightarrow}\inf\limits_{\Theta}l_p(\Theta|\x)$, and
$\arg\inf\limits_{\Theta}\bar{l}_p(\Theta|\x,\e)\overset{\mbox{a.s.}}{\longrightarrow}\arg\inf\limits_{\Theta}l_p(\Theta|\x)$
due to the convexity of the loss function.
  	
\subsection{PGM}
The averaged noise-augmented loss function over $m$ iterations upon convergence is
\begin{align}
\!\!&\!\!\bar{l}_p(\Theta|\x,\e)\!=\!l(\Theta|\x)\!-\!\frac{1}{m}\!\sum_{t=1}^{m}\!\sum_{i=1}^{n_e}\!\sum_{j=1}^{p}\!\!\left(\!e_{ijj}\!\!\left(\!\theta_{j0}\!+\!\!\sum_{k\ne j}\!e_{ijk}^{(t)}\theta_{jk}\!\!\right)\!\!-\! \log(e_{ijj}!)\!-\!\exp\!\!\left(\!\theta_{j0}\!+\!\!\sum_{k\ne j}\!e_{ijk}^{(t)}\theta_{jk} \!\right)\!\!\right)\!\!\!\label{app:poiexp0}\\
=&l(\Theta|\x)\!-\frac{1}{m}\!\sum_{t=1}^{m}\sum_{j=1}^{p}e_{ijj}\sum_{k\ne j} \sum_{i=1}^{n_e} \theta_{jk}e_{ijk}^{(t)}+\frac{1}{m}\!\sum_{t=1}^{m}\sum_{j=1}^{p}\sum_{i=1}^{n_e}\exp\left(\!\theta_{j0}\!+\!\sum_{k\ne j}\theta_{jk}e_{ijk}^{(t)} \right)+C\notag\\	
=&l(\Theta|\x)\boxed{\!-\frac{1}{m}\!\sum_{t=1}^{m}\sum_{j=1}^{p}\!e_{ijj}\!\sum_{k\ne j}\left(\!\frac{\sqrt{\lambda}\theta_{jk}}{|\theta_{jk}|^{\frac{\gamma}{2}}}\!\sum_{i=1}^{n_e}z^{(t)}_{ijk}\!\right)\!\!+\!\frac{1}{m}\!\sum_{t=1}^{m}\!\sum_{j=1}^{p}\!\sum_{i=1}^{n_e}\!\exp\!\left(\!\theta_{j0}\!+\!\sum_{k\ne j}\!\frac{\sqrt{\lambda} \theta_{jk}}{|\theta_{jk}|^{\frac{\gamma}{2}}}z^{(t)}_{ijk}\!\!\right)}\!\!+\!C,\label{app:poiexp1}\\
=&l(\Theta|\x)+P(\Theta)+C,\notag
\end{align}
where $P(\Theta)$ refers to the boxed expression in Eqn (\ref{app:poiexp1}), $z^{(t)}_{ijk}\sim N(0,1),e_{ijj}\equiv n^{-1}\sum_{i'=1}^nx_{i'j}$ that is a constant across $i=1,\ldots, n_e$ for a given $j$, and $C$ is a constant not related to $\Theta$. The regularizer $P(\Theta)$ is different for $n_e\rightarrow\infty$ vs $m\rightarrow\infty$. We thus consider each case separately.   

\emph{Case 1: $n_e\rightarrow\infty$, $n_e\lambda=O(1)$ and  fixed $m$}\\
Assume $m=1$ WLOG, then $z^{(t)}_{ijk}$ can be abbreviated as $z_{ijk}$. $n_e\rightarrow\infty$ and $\lambda n_e=O(1)$ implies that $\lambda\rightarrow0$, therefore, $\sum_{k\ne j}\!\frac{\sqrt{\lambda} \theta_{jk}}{|\theta_{jk}|^{\frac{\gamma}{2}}}z_{ijk}\rightarrow0$ in Eqn (\ref{app:poiexp1}). Apply the second order Taylor expansion around $\sum_{k\ne j}\theta_{jk}z_{ijk}=0$ to Eqn (\ref{app:poiexp1}), as $n_e\rightarrow\infty$, 
\begin{align}
\bar{l}_p(\Theta&|\x,\e)\!\rightarrow\textstyle
l(\Theta|\x)\!-\!\sum_{j=1}^{p}e_{ijj}\sum_{k\ne j}\!\left( \frac{\sqrt{\lambda}\theta_{jk}}{|\theta_{jk}|^{\frac{\gamma}{2}}}\sum_{i=1}^{n_e}z_{ijk}\!\right)\!+\!\!\sum_{j=1}^{p}\!\exp(\theta_{j0})\sum_{i=1}^{n_e}\sum_{k\ne j}\frac{\sqrt{\lambda} \theta_{jk}}{|\theta_{jk}|^{\frac{\gamma}{2}}}z_{ijk}\notag\\
&\ \ \ \ \ \ \ \ \ \textstyle+\!\frac{1}{2}\!\sum_{j=1}^{p}\exp(\theta_{j0})\sum_{i=1}^{n_e}\left(\sum_{k\ne j}\frac{\sqrt{\lambda} \theta_{jk}}{|\theta_{jk}|^{\frac{\gamma}{2}}}z_{ijk} \right)^2+O\left(n_e^{-1}\right)C_1(\Theta)N(1,1)+C\label{app:poiexp2}\\
\approx&\textstyle l(\Theta|\x)\!+\!\frac{1}{2}\!\sum_{j=1}^{p}\!\exp(\theta_{j0})\!\sum_{k\ne j}\!\left( \frac{{\lambda}\theta_{jk}^2}{|\theta_{jk}|^{\gamma}} \sum_{i=1}^{n_e}z_{ijk}^2\!\right)\!\!+\!\!\sum_{j=1}^{p}\!\exp(\theta_{j0})\!\sum_{k<l\ne j}\!\left(\! \frac{\!\lambda\theta_{jk}\theta_{jl}}{|\theta_{jk}\theta_{jl}|^{\frac{\gamma}{2}}}\!\sum_{i=1}^{n_e}\!z_{ijk}z_{ijl}\!\right)\notag\\
&\ \ \ \ \ \ \ \ \ +O\left(n_e^{-1}\right)C_1(\Theta)N(1,1)+C\label{app:poiexp3}\\
\rightarrow &\textstyle l(\Theta|\x)+\frac{\lambda n_e}{2}\sum_{j=1}^{p}\!\exp(\theta_{j0})\!\sum_{j\neq k}|{\theta_{jk}}|^{2-\gamma}\notag\\
&\qquad\quad+\!O\left(n_e^{-0.5}\right)\!C_2(\Theta)\!N(0,1)\!+\!O\left(n_e^{-1}\right)\!C_1(\Theta)N(1,1)\!+\!C.\label{eqn:2O}
\end{align}
In PGM, $e_{ijj}\equiv  n^{-1}\sum_{i'=1}^nx_{i'j}$, the average of the observations in the outcome node, the log of which estimates $\theta_{j0})$ with the canonical log link function. In other words, when $n_e\rightarrow\infty$ $e_{ijj}=\exp(\theta_{j0})$; therefore, the second and third terms in Eqn (\ref{app:poiexp2}) cancel out. $C_1(\Theta)$ and $C_2(\Theta)$ are functions of $\Theta$ and the standard deviations associated with the  two asymptotic normality terms in Eqn (\ref{eqn:2O}) which result from the summation over $n_e$ noise terms per the CLT, and the $C_2(\Theta)$ term is the rate-limiting term and
\begin{equation}\label{app:asypoi1}
C_2(\Theta)=\frac{\lambda n_e}{2} \bigg(2\sum_{j=1}^{p} \!\exp(2\theta_{j0})\! \bigg|\bigg|\bigg(|{\bs\theta_{j,-j}}|^{1-\frac{\gamma}{2}}\bigg)\bigg(|{\bs\theta_{j,-j}}|^{1-\frac{\gamma}{2}}\bigg)^T\bigg|\bigg|_2^2\bigg)^{1/2} \mbox{ where } \lambda n_e=O(1).
\end{equation}
Note that $l(\Theta|\x)+\frac{\lambda n_e}{2}\sum_{j=1}^{p}\!\exp(\theta_{j0})\!\sum_{j\neq k}|{\theta_{jk}}|^{2-\gamma}$ in Eqn (\ref{eqn:2O}) is $l_p(\Theta|\x)=\E_{\e}(l_p(\Theta|\x,\e)$ per Proposition \ref{prop:glmregularization} and Appendix \ref{app:glmregularization}.
As $n_e\rightarrow\infty$ and $\lambda n_e=O(1)$, per the strong LLN and Eqn (\ref{eqn:2O}),  $\bar{l}_p(\Theta|\x,\e)$  converges almost surely to $l_p(\Theta|\x)$.
Given the convexity of the loss function and per Claim \ref{cla:consistineq},
$\arg\inf\limits_{\Theta}\bar{l}_p(\Theta|\x,\e)\overset{{a.s.}}{\longrightarrow}\arg\inf\limits_{\Theta}l_p(\Theta|\x)$.

\emph{Case 2: $m\rightarrow\infty$ and  fixed $n_e$}\\
The 2nd term in Eqn (\ref{app:poiexp1}) is the summation of Gaussian variables, and the 3rd term follows a log-normal distribution. Therefore, we can rewrite Eqn (\ref{app:poiexp1})  as
\begin{align}
&\bar{l}_p(\Theta|\x,\e)=\textstyle l(\Theta|\x)\!-\sum_{j=1}^{p}\!e_{ijj}\!\sum_{k\ne j}\frac{\sqrt{\lambda }n_e\theta_{jk}}{\sqrt{m}|\theta_{jk}|^{\frac{\gamma}{2}}}N(0,1)\notag\\
&\textstyle\qquad\qquad\quad+m^{-1}\!\sum_{t=1}^{m}\sum_{j=1}^{p}\sum_{i=1}^{n_e}\mbox{LogN}\left(\theta_{j0}, \sum_{k\ne j} \frac{\lambda\theta_{jk}^2}{|\theta_{jk}|^\gamma}\right)+C.\label{eqn:PGMm}
\end{align}
Applying the CLT to Eqn (\ref{eqn:PGMm}) as $m\rightarrow\infty$, 
\begin{align}
&\textstyle\bar{l}_p(\Theta|\x,\e)\rightarrow l(\Theta|\x)\!-\sum_{j=1}^{p}e_{ijj}\sum_{k\ne j}\frac{\sqrt{\lambda} n_e\theta_{jk}}{\sqrt{m}|\theta_{jk}|^{\frac{\gamma}{2}}}N(0,1) \label{app:asymptpoi}\\
&+ \left\{ \frac{n_e}{m}\sum_{j=1}^{p}\left(\exp \left(\sum_{k\ne j} \frac{\sqrt{\lambda}\theta_{jk}}{|\theta_{jk}|^{\frac{\gamma}{2}}}\right)^2-1 \right)\exp\!\left(2\theta_{j0}+\left(\sum_{k\ne j} \frac{\sqrt{\lambda}\theta_{jk}}{|\theta_{jk}|^{\frac{\gamma}{2}}}\right)^2  \right)\right\}^{1/2}\!N(0,1)+C,\notag
\end{align}
suggesting that $\bar{l}_p(\Theta|\x,\e)$ follows a Gaussian distribution asymptotically.
Per the strong LLN as $m\rightarrow\infty$, Eqn (\ref{app:asymptpoi}) converges almost surely to
\begin{align}
&\E_{\e}(l_p(\Theta|\x,\e))
= l_p(\Theta|\x)=l(\Theta|\x)+P(\Theta)+C\notag\\
&=l(\Theta|\x)+n_e\sum_{j=1}^{p}\exp\left(\theta_{j0}\right)\exp\left(2^{-1}\lambda\left( \sum_{k\neq j} {|\theta_{jk}|^{1-\frac{\gamma}{2}}}\right)^2 \right)+C \label{app:poiexp4}
\end{align}
for all $\Theta\in\bs\Theta$ assuming $\bs\Theta$ to be compact. 
Per claim \ref{cla:consistineq},
$\sup\limits_{\Theta}\left|\bar{l}_p(\Theta|\x,\e)-l_p(\Theta|\x)\right|\overset{\mbox{a.s.}}{\longrightarrow}0\mbox{ , as $m\rightarrow\infty$}
\Rightarrow\inf\limits_{\Theta}\bar{l}_p(\Theta|\x,\e)\overset{\mbox{a.s.}}{\longrightarrow}\inf\limits_{\Theta}l_p(\Theta|\x)
\Rightarrow\arg\inf\limits_{\Theta}\bar{l}_p(\Theta|\x,\e)\overset{\mbox{a.s.}}{\longrightarrow}\arg\inf\limits_{\Theta}l_p(\Theta|\x)$ given the convexity of the loss function.

\subsection{EGM}
The averaged noise-augmented loss function over $m$ iterations upon convergence is
$$ \textstyle \bar{l}_p({\Theta}|\x,\e)=l(\Theta|\x)-\textstyle m^{-1}\!\sum_{t=1}^{m}\!\sum_{i=1}^{n_e}\!\sum_{j=1}^{p}\!\left(\theta_{j0}+\sum_{k\ne j}e^{(t)}_{ijk}\theta_{jk}\!-e_{ijj}\exp\left(\theta_{j0}+\sum_{k\ne j}e^{(t)}_{ijk}\theta_{jk} \right) \right), $$
where $e_{ijj}=n^{-1}\sum_{i'=1}^nx_{i'j}$. The above loss function is equivalent to the loss function in Eqn (\ref{app:poiexp0}) in the PGM case except for the constant term that does not involve $\Theta$. Therefore, the proof for PGM  also applies in the case of EGM.

\subsection{NBGM}
The averaged noise-augmented loss function over $m$ iterations upon convergence is
\begin{align}
&\bar{l}_p({\Theta}|\x,\e)=\textstyle l(\Theta|\x)-\!\frac{1}{m}\!\sum_{t=1}^{m}\!\sum_{i=1}^{n_e}\!\sum_{j=1}^{p}\!\bigg(\!\log\!\left(\!\frac{\Gamma(e_{ijj}\!+\!r_j)r_j^{r_j}}{\Gamma(e_{ijj}\!\!+\!1)\Gamma(r_j)}\!\right)\!+\!e_{ijj}\!\!\sum_{k\ne j}\!e_{ijk}^{(t)}\theta_{jk}\notag\\
&\textstyle\qquad\qquad\quad -\!(r_j\!+\!e_{ijj})\log\!\left(\!r_j\!+\!\exp\!\left(\!\theta_{j0}\!+\!\sum_{k\ne j}\!e_{ijk}^{(t)}\theta_{jk}\!\right)\right)\!\bigg)\label{app:nb0}\\
&=l(\Theta|\x)+\!C\notag\\
&\quad\!-\!\frac{1}{m}\!\sum_{t=1}^{m}\!\sum_{i=1}^{n_e}\!\sum_{j=1}^{p}\!e_{ijj}\!\!\sum_{k\ne j}\!e_{ijk}^{(t)}\theta_{jk}\!+\!\frac{1}{m}\!\sum_{t=1}^{m}\!\sum_{i=1}^{n_e}\!\sum_{j=1}^{p}(r_j\!\!+\!e_{ijj})\log\!\left(\!\!r_j\!+\!\exp\!\left(\!\theta_{j0}\!+\!\!\sum_{k\ne j}e_{ijk}^{(t)}\theta_{jk}\!\!\right)\!\!\right)\label{app:nb01}\\
&=l(\Theta|\x)+C\notag\\
\!&\boxed{-\frac{1}{m}\!\sum_{t=1}^{m}\!\sum_{j=1}^{p}\!e_{ijj}\!\sum_{k\ne j}\!\!\left(\! \frac{\sqrt{\lambda}\theta_{jk}}{|\theta_{jk}|^{\frac{\gamma}{2}}}\!\sum_{i=1}^{n_e}\!z_{ijk}^{(t)}\!\!\right) \!+\!\frac{1}{m}\!\sum_{t=1}^{m}\!\sum_{j=1}^{p}\!\sum_{i=1}^{n_e}(r_j+1)\!\log\!\!\left(\!\!r_j\!\exp\!\left(\!\theta_{j0}\!+\!\sum_{k\ne j}\!\frac{\sqrt{\lambda} \theta_{jk}}{|\theta_{jk}|^{\frac{\gamma}{2}}}z_{ijk}^{(t)}\!\right)\!\!\right)\!\!}\!\!\label{app:nb1}\\
&= l(\Theta|\x)+P(\Theta)+C=l_p(\Theta|\x)+C,\notag
\end{align}
where $P(\Theta$) refers to the boxed expression in Eqn (\ref{app:nb1}), $z^{(t)}_{ijk}\sim N(0,1),e_{ijj}\equiv n^{-1}\sum_{i'=1}^nx_{i'j}$ that is a constant across $i=1,\ldots, n_e$ for a given $j$, and $C$ is a constant not related to $\Theta$. The regularizer $P(\Theta)$ is different for $n_e\rightarrow\infty$ vs $m\rightarrow\infty$. We thus consider each case separately.

\emph{Case 1: $n_e\rightarrow\infty$ and $n_e\lambda=O(1)$ and  fixed $m$}\\
Let $m=1$ WLOG, thus $z_{ijk}^{(t)}$ can be abbreviated as $z_{ijk}$. Since $n_e\rightarrow\infty$ and $\lambda n_e=O(1)$, implying $\lambda\rightarrow 0$ and thus $\exp\left(\sum_{k\ne j}\frac{\sqrt{\lambda} \theta_{jk}}{|\theta_{jk}|^{\frac{\gamma}{2}}}z_{ijk} \right)\rightarrow 1$. Applying the second order Taylor expansion  around $\sum_{k\ne j}\theta_{jk}z_{ijk}=0$ to Eqn (\ref{app:poiexp1}), we have
\begin{align}
\bar{l}_p(\Theta&|\x,\e)\!=l(\Theta|\x)\!-\!\sum_{j=1}^{p}e_{ijj}\sum_{k\ne j}\left( \frac{\sqrt{\lambda}\theta_{jk}}{|\theta_{jk}|^{\frac{\gamma}{2}}}\sum_{i=1}^{n_e}z_{ijk}\right)+\!\sum_{j=1}^{p}\frac{(r_j+e_{ijj})\exp(\theta_{j0})}{r_j+\exp(\theta_{j0})}\sum_{i=1}^{n_e}\sum_{k\ne j}\frac{\sqrt{\lambda} \theta_{jk}}{|\theta_{jk}|^{\frac{\gamma}{2}}}z_{ijk}\notag\\
&\ \ \ \ +\!\frac{1}{2}\!\sum_{j=1}^{p}\sum_{i=1}^{n_e}\frac{(r_j+e_{ijj})r_j\exp(\theta_{j0})}{(r_j+\exp(\theta_{j0}))^2}\left(\sum_{k\ne j}\frac{\sqrt{\lambda} \theta_{jk}}{|\theta_{jk}|^{\frac{\gamma}{2}}}z_{ijk} \right)^2+O\left(n_e^{-1}\right)N(1,C_1(\Theta)) +C\label{app:nb2}\\
\rightarrow &l(\Theta|\x)\!+\!\frac{1}{2}\!\sum_{j=1}^{p}\!\sum_{k\ne j}\!\frac{r_j\exp(\theta_{j0})}{r_j\!+\!\exp(\theta_{j0})}\!\!\left(\!\frac{{\lambda}\theta_{jk}^2}{|\theta_{jk}|^{\gamma}}\!\sum_{i=1}^{n_e}z_{ijk}^2\!\!\right)\!\!+\!\!\sum_{j=1}^{p}\!\sum_{k<l\ne j}\!\frac{r_j\exp(\theta_{j0})}{r_j\!+\!\exp(\theta_{j0})}\!\!\left(\!\frac{{\lambda}\theta_{jk}\theta_{jl}}{|\theta_{jk}\!\theta_{jl}|^{\frac{\gamma}{2}}}\!\!\sum_{i=1}^{n_e}\!z_{ijk}e_{0ijl}\!\! \right)\notag \\
&+O\left(n_e^{-1}\right)N(1,C_1(\Theta))+\!C\label{app:nb3}\\
\rightarrow &\textstyle l(\Theta|\x)\!+\!\frac{1}{2}\!\sum_{j=1}^{p}\!\sum_{k\ne j}\!\frac{r_j\exp(\theta_{j0})}{r_j\!+\!\exp(\theta_{j0})}\!\!\left(\!\frac{{\lambda}\theta_{jk}^2}{|\theta_{jk}|^{\gamma}}\!\sum_{i=1}^{n_e}z_{ijk}^2\right)\notag \\
&\qquad+O\left(n_e^{-1}\right)N(1,C_1(\Theta))\!+\!O\!\left(n_e^{-0.5}\right)C_2(\Theta)N(0,1)+\!C\label{eqn:2ONB}
\end{align}
In NBGM, $e_{ijj}\equiv  n^{-1}\sum_{i'=1}^nx_{i'j}$, the average of the observations in the outcome node, the logarithm of which estimates $\theta_{j0}$ with the canonical log link function. In other words, when $n_e\rightarrow\infty$ $e_{ijj}=\exp(\theta_{j0})$, and $r_j+\exp(\theta_{j0})=r_j+e_{ijj}$; therefore, the second and third terms in Eqn (\ref{app:nb2}) cancel out and the forth term can be simplied as shown above. $C_1(\Theta)$ and $C_2(\Theta)$ are functions of $\Theta$ and the standard deviations associated with the  two asymptotic normality terms in Eqn (\ref{eqn:2ONB}) that result from the summation over $n_e$ noise terms per the CLT, and the $C_2(\Theta)$ term is the rate-limiting term and
\begin{equation}\label{app:asynb1}
C_2(\Theta)=\frac{\lambda n_e}{2} \bigg(2\sum_{j=1}^{p} \bigg(\frac{r_j\exp(\theta_{j0})}{r_j\!+\!\exp(\theta_{j0})}\bigg)^2 \bigg|\bigg|\bigg(|{\bs\theta_{j,-j}}|^{1-\frac{\gamma}{2}}\bigg)\bigg(|{\bs\theta_{j,-j}}|^{1-\frac{\gamma}{2}}\bigg)^T\bigg|\bigg|_2^2\bigg)^{1/2}.
\end{equation}
Note that $l(\Theta|\x)+\frac{1}{2}\!\sum_{j=1}^{p}\!\sum_{k\ne j}\!\frac{r_j\exp(\theta_{j0})}{r_j\!+\!\exp(\theta_{j0})}\!\!\left(\!\frac{{\lambda}\theta_{jk}^2}{|\theta_{jk}|^{\gamma}}\!\sum_{i=1}^{n_e}z_{ijk}^2\right)$ in Eqn (\ref{eqn:2ONB}) is $l_p(\Theta|\x)\!=\!\E_{\e}(l_p(\Theta|\x,\e)$ per Proposition \ref{prop:glmregularization} and Appendix \ref{app:glmregularization}.
As $n_e\rightarrow\infty$ and $\lambda n_e=O(1)$, per the strong LLN and Eqn (\ref{eqn:2ONB}),  $\bar{l}_p(\Theta|\x,\e)$  converges almost surely to $l_p(\Theta|\x)$. Given the convexity of the loss function and per Claim \ref{cla:consistineq},
$$\arg\inf\limits_{\Theta}{l}_p(\Theta|\x,\e)\overset{{a.s.}}{\longrightarrow}\arg\inf\limits_{\Theta}l_p(\Theta|\x).$$

\emph{Case 2: $m\rightarrow\infty$ and  fixed $n_e$}\\
The second term in Eqn (\ref{app:nb01}) is the summation over Gaussian variables, therefore, the equation can be written as
\begin{align}
\bar{l}_p(\Theta|\x,\e)&=\textstyle l(\Theta|\x)-\!\sum_{j=1}^{p}e_{ijj}\sum_{k\ne j}\frac{\sqrt{\lambda} n_e\theta_{jk}}{\sqrt{m}|\theta_{jk}^{(t-1)}|^{\frac{\gamma}{2}}}N(0,1)\!+\!\frac{1}{m}\!\sum_{t=1}^{m}\!\sum_{i=1}^{n_e}U^{(t)}_i+C,\notag\\
&=\textstyle l(\Theta|\x)-\!\sum_{j=1}^{p}e_{ijj}\sum_{k\ne j}\frac{\sqrt{\lambda} n_e\theta_{jk}}{\sqrt{m}|\theta_{jk}^{(t-1)}|^{\frac{\gamma}{2}}}N(0,1)\!+\!\frac{n_e}{m}\!\sum_{t=1}^{m}\!U^{(t)}+C,\label{eqn:NBm}
\end{align}
where $U^{(t)}_i\!=\!\sum_{j=1}^p\!(r_j\!+e_{ijj})\log\!\left(\!r_j\!+\!\exp\left(\!\sum_{k\ne j}\!e_{ijk}^{(t)}\theta_{jk}\!\right)\!\right)$. The second equation holds because $U^{(t)}_i$ is the same for all $i=1,\ldots,n_e$. Applying the CLT to the $U$-term in Eqn (\ref{eqn:NBm}) as $m\rightarrow\infty$,
\begin{align}
&\textstyle\bar{l}_p(\Theta|\x,\e)\rightarrow l(\Theta|\x)\!-\sum_{j=1}^{p}e_{ijj}\sum_{k\ne j}\frac{\sqrt{\lambda} n_e\theta_{jk}}{\sqrt{m}|\theta_{jk}|^{\frac{\gamma}{2}}}N(0,1)+n_e\E\left(U^{(t)}\right)+\frac{n_e}{\sqrt{m}}N\left(0,\sigma_U\right)\notag\\
&\textstyle= l(\Theta|\x)+n_e E(U^{(t)})\!-\sum_{j=1}^{p}e_{ijj}\sum_{k\ne j}\frac{\sqrt{\lambda} n_e\theta_{jk}}{\sqrt{m}|\theta_{jk}|^{\frac{\gamma}{2}}}N(0,1)+\frac{n_e}{\sigma_U\sqrt{m}}N(0,1),\label{app:asymptnb}
\end{align}
where $\sigma_U$ is the standard deviation of $U^{(t)}$. Since $\log(r_j+\exp(*))\!\rightarrow\!\max\{\log(r_j),*\}$, as $*\rightarrow\pm\infty$, $\sigma_U$ is a finite.  Eqn (\ref{app:asymptnb}) suggests that $\bar{l}_p(\Theta|\x,\e)$ follows a Gaussian distribution as $m\rightarrow\infty$.

Additionally, applying the strong LLN to Eqn (\ref{app:nb01}), $\bar{l}_p(\Theta|\x,\e)$  converges almost surely to its mean $l_p(\Theta|\x)=\E(l_p(\Theta|\x,\e))$ for all $\Theta\in\bs\Theta$ as $m\rightarrow\infty$, assuming $\bs\Theta$ to be compact; that is,
\begin{align}
\bar{l}_p(\Theta|\x,\e)&\rightarrow l_p(\Theta|\x)+C= l(\Theta|\x)\!+n_e \E(U^{(t)}_i)+C.\label{app:nbbin1}
\end{align}
It follows that $\sup\limits_{\Theta}\left|\bar{l}_p(\Theta|\x,\e)-l_p(\Theta|\x)\right|\overset{\mbox{a.s.}}{\longrightarrow}0\mbox{ as $m\rightarrow\infty$}
\Rightarrow\inf\limits_{\Theta}l_p(\Theta|\x,\e)\overset{\mbox{a.s.}}{\longrightarrow}\inf\limits_{\Theta}l_p(\Theta|\x)
\Rightarrow\arg\inf\limits_{\Theta}l_p(\Theta|\x,\e)\overset{\mbox{a.s.}}{\longrightarrow}\arg\inf\limits_{\Theta}l_p(\Theta|\x)$ given the convexity of the loss function.

\subsection{BGM}
The averaged noise-augmented loss function over $m$ iterations upon convergence is
$$\bar{l}_p({\Theta}|\x,\e)=l(\Theta|\x)-\frac{1}{m}\sum_{t=1}^{m}\sum_{i=1}^{n_e}\sum_{j=1}^{p}\left(e_{ijj}\sum_{k\ne j}e^{(t)}_{ijk}\theta_{jk}-\log\left(1+\exp\left(\theta_{j0}+\sum_{k\ne j}e^{(t)}_{ijk}\theta_{jk}\right)\right)\right),$$
which is a special case of Eqn (\ref{app:nb0}) by setting  $r_j=1$. As such, the proof for NBGM also applies to BGM.  


\subsection{Graphical Ridge for GGM}
Let $\z=\sqrt{n^{-1}(n+n_e)}\x$ and $\mathbf{d}=\sqrt{n_e^{-1}(n+n_e)}\e$.
By Claim \ref{cla:consistineq}, we only need to prove $(n+n_e)^{-1}\left|l_p(\Omega|\z,\mathbf{d})-l_p(\Omega|\z)\right|$ $\rightarrow 0, \forall\;\Omega\in\bs\Omega$ almost surely and obtain its probability bound. The noise-augmented loss function is $l_p(\Omega|\z,\mathbf{d})$
\begin{align*}
&\textstyle=-(n+n_e)\log(\Omega)+\frac{n+n_e}{n}\sum_{i=1}^{n}\x_i^T\Omega\x_i+\frac{n+n_e}{n_e}\frac{1}{m}\sum_{t=1}^{m}\sum_{i=1}^{n_e}\e_i^T\Omega\e_i\\
&\textstyle=\frac{n+n_e}{n}l(\Omega|\x)+\frac{n+n_e}{m n_e}\sum_{t=1}^{m}\sum_{i=1}^{n_e}\sum_{j,k=1}^{p}(e_{ij}e_{ik})\omega_{jk}\\
&\textstyle=\frac{n+n_e}{n}l(\Omega|\x)+\frac{\lambda(n+n_e)}{m}\sum_{t=1}^{m}\left(\sum_{j,k=1}^{p}\omega_{jk}\frac{1}{n_e}\sum_{i=1}^{n_e}(e_{ij}e_{ik})\right)\\%
&=\frac{n+n_e}{n}l(\Omega|\x)\!+\!\frac{\lambda(n+n_e)}{m}\sum_{t=1}^{m}\!\left(\!\sum_{j,k=1}^{p}\!\omega_{jk}\frac{1}{n_e}\!\sum_{i=1}^{n_e}\left(\!e_{ij}\!\!\left(\!\frac{\omega_{jk}^{(t-1)}}{\omega_{jj}^{(t-1)}}e_{ij}\!+\!\sqrt{\frac{\omega_{jj}^{(t-1)}\omega_{kk}^{(t-1)}\!\!-\omega_{jk}^{(t-1)}}{\omega_{jj}^{(t-1)}}}e_{0i}\!\right)\!\!\right)\!\!\right),
\end{align*}
where $e_{ij}$ and $e_{0i}$ are independent  with mean 0; and $e_{ij}^2$ and $e_{ij}e_{0i}$ are uncorrelated.
\begin{align*}
l_p(\Omega|\z,\mathbf{d})
&\sim\!l_p(\Omega|\z)\!+\!\frac{\lambda(n\!+\!n_e)}{m}\!\sum_{t=1}^{m}\!\!\left(\!\sum_{j,k=1}^{p}\!\!\omega_{jk}\omega_{jk}^{(t-1)}\!\frac{1}{n_e}\!\sum_{i=1}^{n_e}\!\!\left(\!\!\chi^2_1\!-\!1\!+\!\!\sqrt{ \frac{\omega_{jj}^{(t-1)}\!\omega_{kk}^{(t-1)}\!-\!\omega_{jk}^{(t-1)}}{\omega_{jk}^{2(t-1)}} }\frac{\chi_1^2\!-\!\chi_1^2}{2}\!\right)\!\!\right)\notag\\
&\rightarrow\!l_p(\Omega|\z)\!+\!\frac{\lambda(n\!+\!n_e)}{\sqrt{mn_e}}\!\left(\!\sum_{j,k=1}^{p}\!\!\omega_{jk}\omega_{jk}^{(t-1)}\!\sqrt{2}N(0,1)\!+\!\!\!\sum_{j,k=1}^{p}\!\!\omega_{jk}\sqrt{ \omega_{jj}^{(t-1)}\!\omega_{kk}^{(t-1)}\!\!-\!\omega_{jk}^{(t-1)}}N(0,1)\!\!  \right),
\end{align*}
where $\textstyle l_p(\Omega|\z)=\frac{n+n_e}{n}l(\Omega|\x)+\lambda(n+n_e)\sum_{j,k=1}^{p}\omega_{jk}^2 \mbox{ as } n_e\rightarrow\infty \mbox{ or } m\rightarrow\infty$. Assume  $\omega_{jk}\subset\Omega\in\bs\Omega$ is bounded by $B$, then
\begin{align*}
&(n+n_e)^{-1}\sup\limits_{\Omega}\left|l_p(\Omega|\z,\mathbf{d})-l_p(\Omega|\z)\right|\rightarrow \lambda p^2(m n_e)^{-1/2}(\sqrt{2}B^2+B\sqrt{B^2+B}) N(0,1)\rightarrow 0\\
\Rightarrow&\inf\limits_{\Omega}l_p(\Omega|\x,\e)\overset{a.s.}{\longrightarrow}\inf\limits_{\Omega}l_p(\Omega|\x) \mbox{ as } m\rightarrow\infty \mbox{ or } n_e\rightarrow\infty\Rightarrow\arg\inf\limits_{\Omega}l_p(\Omega|\x,\e)\overset{a.s.}{\longrightarrow}\arg\inf\limits_{\Omega}l_p(\Omega|\x).
\end{align*}

\section{proof of Proposition \ref{prop:consistmcl}}\label{app:consistmcl}
In the case of multicollinearity, PANDA with  sparsity regularization might experience difficulty in learning minimizer $\hat{\Theta}_p^{(n_e)}$ (or $\hat{\Theta}_p^{(m)}$) when $n_e($ or $m)\rightarrow\infty$. In such a case, we prove that there exists $\epsilon>0$ and a sub-sequence $[n_e]_i$ (or $[m]_i$), such that letting $\nu_i\overset{\Delta}{=}\hat{\Theta}_p^{[n_e]_i}$ (or $\hat{\Theta}_p^{[m]_i}$), then $d\!\left(\!\nu_i,\hat{\bs{\Theta}}^0 \!\right)\!>\!\epsilon$. Denote $\mu_i\!=\!l_p(\nu_i|\x,\e)$, then by Eqn (\ref{eqn:sup}), there exists a sub-sequence $[i]_k$, such that,
\begin{align} \label{eqn:supboundmulti1}
  \Pr\left( \sup\limits_{\Theta}\left|\bar{l}_p(\Theta|\x,\e)-\bar{l}_p(\Theta|\x)\right|>\delta \right)<k^{-1}, k\in N.
  \end{align}
  Since $\bs\Theta$ is compact, the subsequence $[i]_k$ converges to a point $\hat{\Theta}^*\in\bs\Theta$ and $  d\left(\hat{\Theta}^*,\hat{\bs{\Theta}}^0 \right)\geq\epsilon$, so $\hat{\Theta}^*\notin\hat{\bs{\Theta}}^0$. On the other hand, for any $\Theta \in \bs\Theta$, we have
  \begin{align*}
  \bar{l}_p(\hat{\Theta}^*|\x,\e )-\bar{l}_p(\Theta|\x)=& (\bar{l}_p(\hat{\Theta}^*|\x,\e )-\bar{l}_p(\nu_{[i]_k}|\x,\e ) ) + (\bar{l}_p(\nu_{[i]_k}|\x,\e ) -\mu_{[i]_k}(\nu_{[i]_k})) \\
  &+(\mu_{[i]_k}(\nu_{[i]_k}) -\mu_{[i]_k}(\Theta) ) +( \mu_{[i]_k}(\Theta)- \bar{l}_p(\Theta|\x)).
\end{align*}
By the continuity of the loss function and $\lim\limits_{i_k\rightarrow\infty}\nu_{[i]_k}=\hat{ \Theta }^*$, the first term in the above equation is arbitrarily small with $i_k \rightarrow\infty$; by equation (\ref{eqn:supboundmulti1}), the second and forth terms are arbitrarily small with $i_k \rightarrow\infty $, and the third term is non-positive. By the arbitrariness of $\Theta\in\bs\Theta$, we must have $\hat{ \Theta }^*\in\hat{\bs{\Theta}}^0$, which is a contradiction and the Proposition is proved.

\section{Proof of Proposition \ref{prop:fisher}}\label{app:fisher}
WLOG, we derive the Fisher information with the bridge-type noise. The proofs for other types of noise are similar. In the GLM framework when regressing $X_j$ on $\X_{-j}$, the Fisher information matrix $I_{\tilde{\x}}(\bs\theta_j)$ on the augmented data $\tilde{\x}$ is obtained by taking the expectation of the negative second derivative of the noise-augmented loss function in Eqn (\ref{eqn:GLMloss}) over the distribution of data $\x$ and augmented noise $\e$.
\begin{align*}
{I_{\tilde{\x}}}(\bs\theta_j)
=&\E_\x\left(\x_{-j}^T\bs{B}''_j(\x_{-j})\x_{-j}\right)+\E_\e\left(\e_{j,-j}^T\bs{B}''_j(\e_{j,-j})\e_{j,-j}\right)\\
=&\textstyle{I_\x}(\bs\theta_j)+\E_\e\left(\sum_{i=1}^{n_e}\e_{ij,-j}^T{B}_j''(\e_{ij,-j}\bs\theta_j)\e_{ij,-j}\right),
\end{align*}
where $\bs{B}''_j(\x_{-j})=\diag\{B_j''(\x_{1,-j}\bs\theta_j),\ldots,B_j''(\x_{n,-j}\bs\theta_j\!)\}$ and $\bs{B}_j(\e_{j,-j})=\diag\{B_j''(\e_{1,-j}\bs\theta_j),\ldots,$ $B_j''(\e_{n_e,-j}\bs\theta_j\!)\}$.  Let $\lambda n_e=O(1)$ and $\mbox{V}(\e_{ij,-j})$ denote the covariance matrix of $\e_{ij,-j}$; take the second-order Taylor expansion around $\e_{ij,-j}\bs\theta_j=0$, we have
\begin{align*}
I_{\tilde{\x}}(\bs\theta_j)=&{I_\x}(\bs\theta_j)+ n_e{B}_j''(0)\mbox{V}(\e_{ij,-j})+O(\lambda n_e^{1/2})J_p\notag\\
=&\textstyle {I_\x}(\bs\theta_j)+(\lambda n_e){B}_j''(0) \mbox{diag}\{|\theta_{j1}|^{-\gamma},\ldots,|\theta_{jp}|^{-\gamma}\} +O(\lambda n_e^{1/2})J_p,
\end{align*}  
where $J_p$ is a $p\times p$ matrix with all elements equal to 1.

\section{Proof of Corollary \ref{prop:asymp.dist.UGM}}\label{app:CI.UGM}
It is known that $n^{-1/2}l'(\bs\theta|\x)\overset{d}{\rightarrow} N(0, I^{-1}_1(\bs{\theta}))$, where $l'(\bs\theta|\x)$ is the first derivative of the negative log-likelihood function given the observed data $\x$ over $\bs\theta$, and $I_1(\bs{\theta})$ is the information matrix over one observation. It follows that
\begin{equation}\label{eqn:H1}
n^{-1/2}(l'(\bs\theta|\x)+l'(\bs\theta|\e))=n^{-1/2}l'(\bs\theta|\x,\e)\overset{d}{\rightarrow} N(n^{-1/2} l'(\bs\theta|\e), I_1(\bs{\theta}))
\end{equation}
where $\e$ is the augmented noise and $l'(\bs\theta|\e)\!=\!\sum_{i=1}^{n_e}l'(\bs\theta|{\e_i})$. Let $\bs\phi(\e)\!=\! n^{-1/2} l'(\bs\theta|\e)$ and it expectation over the distribution of $\e$ can be worked out for different types of noise. For example, in the regression of  $X_j$ on $\X_{-j}$  with the bridge-type noise,  $\bs\phi_j(\e)\!=\! n^{-1/2} l'(\bs\theta_j|\e)$  and $\E_{\e}(\bs\phi_j)=\frac{\lambda n_e}{\sqrt{n}}\sigma^2\mbox{sgn}(\bs\theta_0)$ for Gaussian outcome nodes,  $\frac{\lambda n_e}{8\sqrt{n}} \mbox{sgn}(\bs\theta_0)+\frac{\lambda^2 n_e}{\sqrt{n}}O(|\bs\theta_0|)$ for Bernoulli outcome nodes,  $\frac{\lambda n_e}{2\sqrt{n}}\mbox{sgn}(\bs\theta_0)\!+\!\frac{\lambda^2 n_e}{{n}}O(|\bs\theta_0|)$ for exponential and Poisson outcome nodes and $\frac{\lambda n_e r}{2(r+1){n}}\mbox{sgn}(\bs\theta_0)\!+\!\frac{\lambda^2 n_e}{\sqrt{n}}O(|\bs\theta_0|)\!$ for NB outcome nodes. If  $\lambda n_e\! =\!o(\sqrt{n})$, then $\E_{\e}(\bs\phi_j)\!\rightarrow\!0$ as $n\!\rightarrow\!\infty$.

Upon the convergence of the PANDA algorithm, in the $j$-th regression, the MLE of $\bs\theta_j$ based on $(\x,\e)$ is the minimizer $\hat{\bs\theta}_{j,\e}$  from solving $l'(\hat{\bs\theta}_{j,\e})=0$, its  first-order Taylor expansion around $\bs\theta_j$ is
$l'(\hat{\bs\theta}_{j,\e})\approx l'(\bs\theta_j|\x,\e)+l''(\bs\theta_j|\x,\e)(\hat{\bs\theta}_{j,\e}-\bs\theta_j)=0$. Therefore, $\hat{\bs\theta}_{j,\e}-\theta_j=-(l''(\bs\theta_j|\x,\e))^{-1}l'(\bs\theta_j|\x,\e)$ and  $\sqrt{n}\left(\hat{\bs\theta}_{j,\e}-\theta_j\right)=-(n^{-1}l''(\bs\theta_j|\x,\e))^{-1}\left(n^{-1/2}l'(\bs\theta_j|\x,\e)\right)$, where $l''(\bs\theta_j|\x,\e)$ is the Hessian matrix and $l''(\bs\theta_j|\x,\e)\rightarrow I_p(\bs\theta_j)$ as $n\rightarrow \infty$. Taken together with Eqn (\ref{eqn:H1}),  assume $\lambda n_e\! =\!o(\sqrt{n})$,  by Slutsky's theorem, then as $n\rightarrow\infty$
\begin{align}\label{eqn:ci1}
\sqrt{n}\left(\hat{\bs{\theta}}_{j,\e}-\bs{\theta}_j)\right)&\!=\! (n^{-1}l''(\bs\theta_j|\x,\e))^{-1}\left(n^{-1/2}l'(\bs\theta_j|\x,\e)\right)\notag\\
&\!\overset{d}{\rightarrow}\! N\left(\0 , I_p(\bs\theta_j)^{-1}I(\bs\theta_j)I_p(\bs\theta_j)^{-1} \right)\!\overset{\Delta}{=}N(\0,\Sigma_{j,\e}).
\end{align}
When the mean of $m>1$ estimates over  consecutive iteration are taken as the final estimate for $\bs{\theta}_j$, that is $\bar{\bs\theta}_j=m^{-1}\sum_{j=1}^m\hat{\bs{\theta}}_{j,\e}^{(t)}$, the variability among the $m$ consecutive estimates will need to be accounted for and be reflected in the variance of the final estimate. It is easy to establish this in the Bayesian framework. Specifically,
\begin{align*}
\E(\bs\theta_j|\x)&=\textstyle\E_\e(\E(\bs\theta_j|\x,\e))=\E_\e(\hat{\bs{\theta}}_{j,\e})=
m^{-1}\sum_{t=1}^m\bs{\theta}^{(t)}_{j,\e}\triangleq \bar{\bs\theta}_j\mbox{ as }m\rightarrow\infty\\
\V(\bs\theta_j|\x)&=
\E_\e(\V(\bs\theta_j|\x,\e))+\V_\e(\E(\bs\theta_j|\x,\e))=\E_\e(\Sigma_{j,\e})+\V_\e(\hat{\bs{\theta}}_{j,\e})\triangleq \bar{\Sigma}_j+ \Lambda_j\\
&=\textstyle m^{-1}\sum_{t=1}^m \Sigma_{j,\e}^{(t)}+(m-1)^{-1}\sum_{t=1}^m
\left(\hat{\bs{\theta}}_{j}^{(t)}-\bar{\bs\theta}_j\right)
\left(\hat{\bs{\theta}}_{j,\e}^{(t)}-\bar{\bs\theta}_j\right)' \mbox{ as }m\rightarrow\infty
\end{align*}
Per the large-sample Bayesian theory, the posterior mean and variance of $\bs{\theta}_j$ given $\x$ are asymptotically equivalent ($n\rightarrow\infty$) to the MLE for $\bs{\theta}$ and the inverse information matrix of $\bs\theta_j$ contained in $\x$. In other words,
$$\textstyle \sqrt{n}(\bar{\bs{\theta}}_j-\bs{\theta}_j)\rightarrow N\left( \0,\bar{\Sigma}_j+ \Lambda_j \right).$$
In the case of a finite $m$ (as in practical application), $\V(\bs\theta_j|\x)$ is estimated
by $\bar{\Sigma}_j+(1+m^{-1})\Lambda_j$ with the correction for the finite $m$.


Applying Proposition \ref{prop:asymp.dist.UGM} to GGMs with lasso-type noise, we have 
$$\sqrt{n}(\hat{\bs{\theta}}_j-\bs{\theta}_j)
\rightarrow N\left(n^{-1/2}\lambda n_e \mbox{sgn}(\bs\theta) M_j^{-1},\sigma^2_jM^{-1}(\x'_{-j}\x_{-j})M_j^{-1} \right),$$
where $M_j=(\x'_{-j}\x_{-j}+\diag(\lambda n_e|\bs\theta|^{-1}))$ and $\sigma^2_j$ is the variance of the error term in the linear regression, and is estimated by 
\begin{align*}
\hat{\sigma}_j^2=&\mbox{SSE}_j(n-\nu_j)^{-1}
=(n-\nu_j)^{-1}(\x_{-j}\bs\theta+\epsilon_j)'(I-H_j)(\x_{-j}\bs\theta+\epsilon_j)\\
=&(n-\nu_j)^{-1}\epsilon_j'(I-H_j)\epsilon_j+(n-\nu_j)^{-1}\left(\bs\theta'\x_{-j}'(I-H_j)\x_{-j}\bs\theta+2\bs\theta'\x_{-j}'(I-H_j)\epsilon_j\right)
\end{align*}
where $H_j=\x_{-j}(\x'_{-j}\x_{-j}+\diag(\lambda n_e|\bs\theta|^{-1}))^{-1}\x'_{-j} \mbox{ and  }\nu_j=\mbox{trace}(H_j)$.

\bibliographystyle{apalike}

\setcounter{section}{0}
\setcounter{algorithm}{0}
\setcounter{cor}{0}
\renewcommand\thesection{S.\arabic{section}}
\renewcommand\thealgorithm{S.\arabic{algorithm}}
\renewcommand\thefigure{S.\arabic{figure}}
\renewcommand\thecor{S.\arabic{cor}}
\renewcommand{\theHcor}{S.\arabic{cor}}

\clearpage
\setstretch{2}
\begin{center}
\Large  \textbf{Supplementary Materials to }\\
\textbf{\emph{AdaPtive Noisy Data Augmentation for Regularization of Undirected Graphical Models}}\\
\normalsize Yinan Li$^{1}$, Xiao Liu$^{2}$, and Fang Liu$^1$\\
\small$^1$ Department of Applied and Computational Mathematics and Statistics\\
\small$^2$ Department of Psychology\\
\small University of Notre Dame, Notre Dame, IN 46556, U.S.A.
\end{center}
\clearpage

\setstretch{1}
\section{PANDA for NS in a single UGM}
\begin{algorithm}[!htp]
\caption{PANDA for NS in a single UGM}\label{alg:NSUGM}
\begin{algorithmic}[1]
\State \textbf{Input}
\begin{enumerate}[leftmargin=0.18in]\setlength\itemsep{-1pt}
\item random initial parameter estimates $\bar{\bs{\theta}}_j^{(0)}$ for $j=1,\ldots,p$.
\item  A NGD in Eqns (\ref{eqn:bridge}) to (\ref{eqn:scad}) and the associated tuning parameters, maximum iteration $T$,  noisy data size $n_e$, width of moving average (MA) window $m$,  threshold $\tau_0$, banked parameter estimates after convergence $r$.
\end{enumerate}
\State $t\leftarrow 0$; convergence $\leftarrow 0$
\State \textbf{WHILE} $t<T$ \textbf{AND} convergence $= 0$
\State \hspace{0.4cm} $t\leftarrow t+1$
\State \textbf{\hspace{0.3cm} FOR} $j = 1$ to $p$
\begin{enumerate}[leftmargin=0.5in]\setlength\itemsep{-1pt}
\item[a)]Generate  $\e_j$ from the NGD with $\bar{\bs{\theta}}_j^{(t-1)}$ plugged in the variance term of the NGD.
\item[b)]Centerize the observed data $\x_{-j}$ in each covariate node, and obtain augmented data $\tilde{\x}_{j}$ by row-combining $(\x_j,\x_{-j})$ with  $(\mathbf{e}_{jj},\e_{.j.})$
\item[c)]Obtain MLE $\hat{\bs{\theta}}_{j}^{(t)}$ by regressing $\tilde{\x}_{j}$ on all other columns $\tilde{\x}_{-j}$ with a proper GLM
\item[d)] If $t>m$, calculate the MA $\bar{\bs{\theta}}^{(t)}_j=m^{-1}\sum_{l=t-m+1}^t \hat{\bs{\theta}}_{j}^{(l)}$; otherwise $\bar{\bs{\theta}}^{(t)}_j=\hat{\bs{\theta}}^{(t)}_j$. Calculate $l^{(t)}_{j}$ with $\bar{\bs\theta}^{(l)}$  plugged in, where $l$ is the negative  log-likelihood  in Eqn (\ref{eqn:GLMloss})..
\end{enumerate}
\hspace{0.45cm} \textbf{End FOR}
\State Calculate the loss function $\bar{l}^{(t)}=m^{-1}\!\sum_{l=t-m+1}^t \sum_{j=1}^p l^{(l)}_{j}$ and apply one of the convergence criteria listed in Remark \ref{rem:converge} to $\bar{l}^{(t)}$. Let convergence $\leftarrow 1 $ if the convergence is reached.
\State\textbf{End WHILE}
\State Continue to execute the command lines 4 and 5  for another $r$ iterations, and record $\bar{\bs\theta}^{(l)}_j$ for $l=t+1,\ldots,t+r$. Let $\bar{\bs{\theta}}_{jk}=(\bar{\theta}_{jk}^{(t+1)},\ldots,\bar{\theta}_{jk}^{(t+r)})$.
\State  Set $\hat{\theta}_{jk}=\hat{\theta}_{kj}=0$ and claim there is no edge between nodes $j$ and $k$
if $\left(\big|\max\{\bar{\bs{\theta}}_{jk}\}\cdot\min\{\bar{\bs{\theta}}_{jk}\}\big|<\tau_0\right) \cap \left(\max\{\bar{\bs{\theta}}_{jk}\}\cdot\min\{\bar{\bs{\theta}}_{jk}\}<0\right)$ or
$\left(\big|\max\{\bar{\bs{\theta}}_{kj}\}\!\cdot\!\min\{\bar{\bs{\theta}}_{kj}\}\big|<\tau_0\right) \cap \left(\max\{\bar{\bs{\theta}}_{kj}\}\!\cdot\!\min\{\bar{\bs{\theta}}_{kj}\}<0\right)$; otherwise, there is an edge between nodes $j$ and $k$. 
\end{algorithmic}
\end{algorithm}

\clearpage
\section{PANDA-CD Algorithm}
\begin{algorithm}[!htb]
\caption{PANDA-CD in a single GGM}\label{alg:CD}
\begin{algorithmic}[1]
\State \textbf{Pre-processing}:  standardize observed data $\x$.
\State \textbf{Input}
\begin{itemize}[leftmargin=0.18in]\setlength\itemsep{-1pt}
\item random initial parameter estimates $\bar{\bs\theta}_j^{(0)}$ and $\hat{\sigma}_j^2$ for $j=2\ldots,p$; let $\hat{\sigma}_1^2=s^2$ where $s^2$ is the sample variance of $\x_1$.
\item A NGD in Eqns (\ref{eqn:bridge}) to (\ref{eqn:scad}) and the associated tuning parameters, maximum iteration $T$, noisy data size $n_e$, MA  window  width $m$, threshold $\tau_0$,  banked parameter estimates after convergence $r$, inner loop $K$ in alternatively estimating $\bs{\theta}_j$ and $\sigma_j^2$
\end{itemize}
\State $t\leftarrow 0$; convergence $\leftarrow 0$
\State \textbf{WHILE} $t < T$ \textbf{AND} convergence $= 0$
\State \hspace{0.4cm} $t\leftarrow t+1$
\State \textbf{\hspace{0.3cm} FOR} $j = 2$ to $p$
\State \textbf{\hspace{0.6cm} FOR} $k = 1:K$
\begin{enumerate}[leftmargin=0.5in]\setlength\itemsep{-1pt}
\item[a)]Generate $n_e$ rows of noisy data $\e_{1:(j-1),}$ from the NGD  with $\bar{\bs{\theta}}_j^{(t-1)}$ plugged in the variance term of the NGD to obtain augmented data as depcited in Figure \ref{fig:pandaCD1}.
\item[b)] Obtain the OLS estimate $\hat{\bs{\theta}}_{j}^{(t)}$ by regressing $\tilde{\x}_{j}$ on $\tilde{\x}_{1:j-1}$, according to Eqn (\ref{eqn:CDreg}).
\item[c)]  If $t>m$, calculate the MA $\bar{\bs{\theta}}^{(t)}_j=m^{-1}\sum_{l=t-m+1}^t \hat{\bs{\theta}}_{j}^{(l)}$; otherwise $\bar{\bs{\theta}}^{(t)}_j=\hat{\bs{\theta}}^{(t)}_j$. Calculate the sum of squared error SSE$^{(t)}_{j}$ given $\bar{\bs{\theta}}^{(t)}_j$ and $\hat{\sigma}_{j}^{2(t)}=$ SSE$^{(t)}_{j}/n$ (Eqn \ref{eqn:CDsigma}).
\end{enumerate}
\textbf{\hspace{0.6cm}END FOR}\\
\textbf{\hspace{0.3cm}END FOR}\\
\textbf{END WHILE}
\State Continue to execute the command lines 5 and 6 for another $r$ iterations, and record $\bar{\bs\theta}^{(l)}_j$ for $l=t+1,\ldots,t+r$, calculate the degrees of freedom $\nu_j^{(t)}=\mbox{trace}(\x_j(\tilde{\x}'_j\tilde{\x}_j)^{-1}\x'_j)$ and $\hat{\sigma}_{j}^{2(l)}=$ SSE$^{(t)}_{j}/(n-\nu_j^{(l)})$. Let $\bar{\bs{\theta}}_{jk}=(\bar{\theta}_{jk}^{(t+1)},\ldots,\bar{\theta}_{jk}^{(t+r)})$.
\State  Set $\hat{\theta}_{jk}=0$ if $\left(\big|\max\{\bar{\bs{\theta}}_{jk}\}\cdot\min\{\bar{\bs{\theta}}_{jk}\}\big|<\tau_0\right) \cap \left(\max\{\bar{\bs{\theta}}_{jk}\}\cdot\min\{\bar{\bs{\theta}}_{jk}\}\right)<0$  for $k> j$; otherwise, set $\hat{\theta}_{jk}=r^{-1}\sum_{l=t+1}^{t+r}\bar{{\theta}}_{jk}^{(l)}$. Set $\hat{D}=r^{-1}\sum_{l=t+1}^{t+r}\diag(\hat{\sigma}_1^{2(l)},\ldots,\hat{\sigma}_p^{2(l)})$. Calculate $\hat\Omega\!=\!\hat{L}'\hat{D}\hat{L} $
\end{algorithmic}
\end{algorithm}

\clearpage
\section{PANDA-SCIO Algorithm}
\begin{algorithm}[!htp]
\caption{PANDA-SCIO in a single GGM}\label{alg:scio}
\begin{algorithmic}[1]
\State \textbf{Pre-processing}: standardize or  observed data $\x$.
\State \textbf{Input}
\begin{itemize}[leftmargin=0.18in]\setlength\itemsep{-1pt}
\item initial parameter estimates $\hat{\bs{\theta}}_{j}^{(0)}=\hat{\Omega}^{(0)}=\left(n^{-1}\x^T\x+0.1I\right)^{-1}$
\item tuning parameters, maximum iteration $T$, noisy data size $n_e$, thresholds $\tau_0,\tau_1$, MA window width $m$, banked parameter estimates after convergence $r$
\end{itemize}
\State $t\leftarrow 0$  and $d\leftarrow C$ ($C$ is a large positive number)
\State \textbf{WHILE} $t < T$  \textbf{AND} $d<\tau$
\State $t\leftarrow t+1$
\State \textbf{\hspace{0.3cm} FOR} $j = 1$ to $p$
\begin{enumerate}[leftmargin=0.45in]\setlength\itemsep{-1pt}
\item[a)]Set $\bar{{\Omega}}_j^{(t-1)}\!\!=\!\bar{{\Omega}}_j^{(t-1)}\1\left(|\bar{{\Omega}}_j^{(t-1)}|\!>\!\tau_1\right)+\!\tau_1\1\left(0\!<\!\bar{{\Omega}}_j^{(t-1)}\!\!<\tau_1\right)-\!\tau_1\1\left(0\!>\!\bar{{\Omega}}_j^{(t-1)}\!\!>\!-\tau_1\right)$
\item[b)]Generate Gaussian noisy data $\e_j$ from a NGD in Eqns (\ref{eqn:bridge}) to (\ref{eqn:scad}) with $\bar{{\Omega}}_j^{(t-1)}$ plugged in the variance term of the NGD.
\item[c)]Obtain augmented data $\tilde{\x}_{j}$ by row-combining $\z=\sqrt{\frac{n+n_e}{n}}\x$ and $\mathbf{d}_j=\sqrt{\frac{2(n+n_e)}{n_e}}\e_j$
\item[d)] Calculate $\hat{\bs\theta}_{j}^{(t)}=(n+n_e)(\tilde{\x}_{j}^T\tilde{\x}_{j})^{-1}_j\1_j$
\item[e)] If $t>m$, calculate MA $\bar{\bs\theta}_{j}^{(t)}=m^{-1}\sum_{l=t-m+1}^t\hat{\bs\theta}_{j}^{(l)}$; otherwise $\bar{\bs\theta}_{j}^{(t)}=\hat{\bs\theta}_{j}^{(t)}$
\item[f)] If $t>m$, calculate $d = m^{-1}\left(\sum^t_{l=t-m+1}l_j^{(l)}-\sum^{t-1}_{l=t-m}l_j^{(l)}\right)$,  where $l_j^{(l)}$ is the loss functions in Eqn (\ref{eqn:qo})  with $\bar{\bs{\theta}}^{(l)}_j$ plugged in.
\end{enumerate}
\textbf{\hspace{0.3cm}End FOR}
\State Calculate the loss function $\bar{l}^{(t)}=m^{-1}\!\sum_{l=t-m+1}^t \sum_{j=1}^p l^{(l)}_{j}$ and apply one of the convergence criteria listed in Remark \ref{rem:converge} to $\bar{l}^{(t)}$. Let convergence $\leftarrow 1 $ if the convergence is reached.
\State\textbf{End WHILE}
\State Continue to execute the command lines 5 to 7 for another $r$ iterations, and record $\bar{\bs\theta}^{(l)}_j$ for $l=t+1,\ldots,t+r$. Let $\bar{\bs{\theta}}_{jk}=(\bar{\theta}_{jk}^{(t+1)},\ldots,\bar{\theta}_{jk}^{(t+r)})$.
\State  Set $\hat{\omega}_{jk}=\hat{\omega}_{kj}=0$ if $\left(\big|\max\{\bar{\bs{\theta}}_{jk}\}\cdot\min\{\bar{\bs{\theta}}_{jk}\}\big|<\tau_0\right) \cap \left(\max\{\bar{\bs{\theta}}_{jk}\}\cdot\min\{\bar{\bs{\theta}}_{jk}\}<0\right)$ or
$\left(\big|\max\{\bar{\bs{\theta}}_{kj}\}\!\cdot\!\min\{\bar{\bs{\theta}}_{kj}\}\big|<\tau_0\right) \cap \left(\max\{\bar{\bs{\theta}}_{kj}\}\!\cdot\!\min\{\bar{\bs{\theta}}_{kj}\}<0\right)$ for $k\ne j=1,\ldots,p$; otherwise, set $\hat{\omega}_{jk}=\!=\!\min\left\{\bar{\bs\theta}_{jk},\bar{\bs{\theta}}_{kj}\right\}$
\end{algorithmic}
\end{algorithm}

\section{Minimizer of averaged noise-augmented loss function vs averaged minimizer
of noise-augmented loss functions}
Per Propositions \ref{prop:regularization} and \ref{prop:glmregularization}, one would take the average over $m$ noise-augmented loss function $l(\Theta|\x,\e)$ to yield a single minimizer $\hat{\bs{\theta}}$, which is the Monte Carlo version of $\E_{\e}(l_p(\bs{\theta}|\x,\e)$ as $m\rightarrow\infty$. However, PANDA would lose its computational edge. To maintain the computational advantage for PANDA, we  instead calculate $\bar{\bs{\theta}}$, the  average of $m$ minimizers of $l(\Theta|\x,\e)$ from the latest $m$ iterations, which is the approach that the PANDA algorithms take in Section \ref{sec:single}.  
We establish in Corollary \ref{cor:average} that $\bar{\bs{\theta}}$ and $\hat{\bs{\theta}}$  are equivalent under some regularity conditions in teh framework of PANDA-NS for GGM.   We also present some numerical examples below to  illustrate the similarity between  $\bar{\bs{\theta}}$ and $\hat{\bs{\theta}}$.
\begin{cor}[\textbf{First-order equivalence between minimizer of averaged noise-augmented loss functions vs averaged minimizers of single noise-augmented loss functions}]\label{cor:average}
The average $\bar{\bs\theta}$ of $m$ minimizers of the $m$ perturbed loss functions in PANDA-NS for GGM upon convergence is first-order equivalent to the minimizer $\hat{\bs\theta}$ of the averaged $m$  noise-augmented loss functions as $m\rightarrow\infty$ or  as $n_e\rightarrow\infty$ while $V(\theta_{jk}n_e)=O(1)$. In addition, The  higher-order difference between  $\bar{\bs\theta}$ and $\hat{\bs\theta}$ also approaches 0 as $n_e\rightarrow\infty$ while $V(\theta_{jk}n_e)=O(1)$.
\end{cor}
Proof: WLOG, we work with the bridge-type noise. in this proof. During the regression with outcome node $X_j$, the average of the minimizers of the $m$ loss functions is
\begin{align}\label{eqn:thetabar}
\bar{\bs\theta}_j=\textstyle  m^{-1}\!\sum_{t=1}^{m}\left(\x_{-j}'\x_{-j}+\sum_{i=1}^{n_e}\e_{i,j,-j}^{(t)'}\e^{(t)}_{i,j,-j}\right)^{-1}\!\!\x_{-j}'\x_j,
\end{align}
where $e_{ijk}\sim N(0,\lambda|\theta_{jk}|^{-1})$. Let $\sum_{i=1}^{n_e}\!\e_{i,j,-j}^{(t)'}\e^{(t)}_{i,j,-j}\!=\!\E\left(\!\sum_{i=1}^{n_e}\!\e_{i,j,-j}^{(t)'}\e^{(t)}_{i,j,-j}\!\right) + A^{(t)}\!=\!\diag(\lambda n_e|\bs\theta_j|^{-\gamma})\\+\bar{A}^{(t)}$; so $A^{(t)}$ can be regarded as the sample deviation of $\sum_{i=1}^{n_e}\e_{i,j,-j}^{(t)'}\e^{(t)}_{i,j,-j}$ from its mean. Let $\bar{A}=m^{-1}\sum_{t=1}^{m}\bar{A}^{(t)}$, the elements of which are
\begin{align}\label{eqn:Abar}
\begin{cases}
\bar{A}[k,k]
\!=\!m^{-1}\sum_{t=1}^{m}\!\sum_{i=1}^{n_e}e_{ijk}^{(t)2}-\lambda n_e|\theta_{jk}|^{-1}&\sim \lambda|m\theta_{jk}|^{-1}(\chi^2_{n_em}\!-\!n_em)\\
\bar{A}[k,l]
\!=\! m^{-1}\sum_{t=1}^{m}\sum_{i=1}^{n_e}e^{(t)}_{ijk}e^{(t)}_{ijl}
&\sim \lambda|\theta_{jk}\theta_{jl}|^{-\frac{1}{2}}m^{-1}\!\sum_{t=1}^{m}\!\sum_{i=1}^{n_e}\!z_{ti}z'_{ti}
\end{cases},
\end{align}
where $z_{ti}\sim N(0,1)$ and $z_{ti}'\sim N(0,1)$ independently.  Let $S_j=(\x_{-j}'\x_{-j}\!+\diag(\lambda n_e|\bs\theta_j|^{-1}))^{-1}$. The Taylor expansion of the inverse of the sum of two matrices, assuming $A^{(t)}$ to be  a small increment, is $(S_j^{-1}+A^{(t)})^{-1}\!=\!S_j-S_jA^{(t)}S_j+S_jA^{(t)}S_jA^{(t)}S_j+\ldots$ Therefore, Eqn (\ref{eqn:thetabar}) becomes
\begin{align}\label{eqn:bar}
\bar{\bs\theta}_j=\textstyle S_j\x_{-j}'\x_j- S_j\left(\bar{A}+O(\lambda^2 n_e)\right)S_j\x_{-j}'\x_j. \end{align}
On the other hand, the minimizer of the average of $m$ loss functions  is
\begin{align}
\hat{\bs\theta}_j=&\textstyle\!\left(\x_{-j}'\x_{-j}+\sum_{i=1}^{n_em}\hat\e_{ij}'\hat\e_{ij}\right)^{-1}\!\!\x_{-j}'\x_j\!=\! \left(\x_{-j}'\x_{-j}\!+\!\diag(\lambda n_e|\bs\theta_j|^{-1})\!+\!\hat{A}\right)^{-1}\!\!\x_{-j}'\x_j,\notag\\
=&S_j\x_{-j}'\x_j- S_j\left(\hat{A} +O(\lambda^2 n_e)\right)S_j\x_{-j}'\x_j,
\label{eqn:hat}
\end{align}
where $e_{ijk}\sim N(0,\lambda|m\theta_{jk}|^{-1})$ for the sake of yielding the same regularization effect as imposed on $\bar{\bs\theta}_j$; and $\hat{A}$ is defined in a similar manner as $\bar{A}$, the elements of which are
\begin{align}\label{eqn:Ahat}
\begin{cases}
\hat{A}[k,k]=\textstyle\sum_{i=1}^{n_em}e_{ijk}^2-\lambda n_e|\theta_{jk}|^{-1}&\sim \lambda|m\theta_{jk}|^{-1}(\chi^2_{n_em}-n_em)\\
\hat{A}[k,l]=\textstyle\sum_{i=1}^{n_em}e_{ijk}e_{ijl} &\sim\frac{\lambda}{m}|\theta_{jk}\theta_{jl}|^{-\frac{1}{2}}\sum_{i=1}^{n_em}z_iz_i'\\
\end{cases},
\end{align}
where $z_{i}\sim N(0,1)$ and $z_{i}'\sim N(0,1)$ independently. $\bar{A}$ and $\hat{A}$ in Eqn (\ref{eqn:Abar}) and (\ref{eqn:Ahat}) follow the same distribution. The expected values of $\bar{A}[k,k], \bar{A}[k,l], \hat{A}[k,k]$, and $\hat{A}[k,l]$ are all equal to zero; the variance of $\bar{A}[k,k]$ and $\hat{A}[k,k]$ is $\lambda^2|m\theta_{jk}|^{-2}2n_em=2\lambda (\lambda n_e)|\theta_{jk}|^{-2}2/m$, and that of $\bar{A}[k,l]$ and $\hat{A}[k,l]$ is $\lambda^2m^{-2}|\theta_{jk}\theta_{jl}|^{-1}n_em=\lambda(\lambda n_e)|\theta_{jk}|^{-2}2/m$. As $m$ increases, both variance terms shrink to 0. As $n_e$ increases while $O(n_e\lambda)=1$, then both variance terms shrinks to 0 as well. In other words, we expect $\bar{A}$ and $\hat{A}$ to be very similar. As such, $\bar{\bs\theta}_j$ in Eqn (\ref{eqn:bar}) and $\hat{\bs\theta}_j$ in Eqn (\ref{eqn:hat}) are also very similar. In addition, as $n_e$ increases and $\lambda n_e=O(1)$, the higher-order terms also goes to 0.

To first illustrate the similarity between $\bar{\bs\theta}$ and $\hat{\bs\theta}$, we simulated data ($n=30$) from linear regression and a Poisson regression models, where the linear predictor is $\X^T\bs\theta=X_1+0.75X_2+0.5X_3+0X_4$. $\X$ and the error in the linear regression was simulated from N$(0, 1)$ independently. The PANDA augmented noises $\e$ in both cases were drawn from N$(0,\lambda^2)$ with $n_e=200$. We examined $m=30,60,90,120$ and $\lambda^2=0.25,0.5,1,2$, calculated $\hat{\bs{\theta}}$ and $\bar{\bs{\theta}}$, and plotted their differences the figure below. The results show minimal difference between $\hat{\bs{\theta}}$ and $\bar{\bs{\theta}}$.
\begin{figure}[!htb]
\centering
linear regression\\
\includegraphics[width=0.75\linewidth, height=0.6\linewidth]{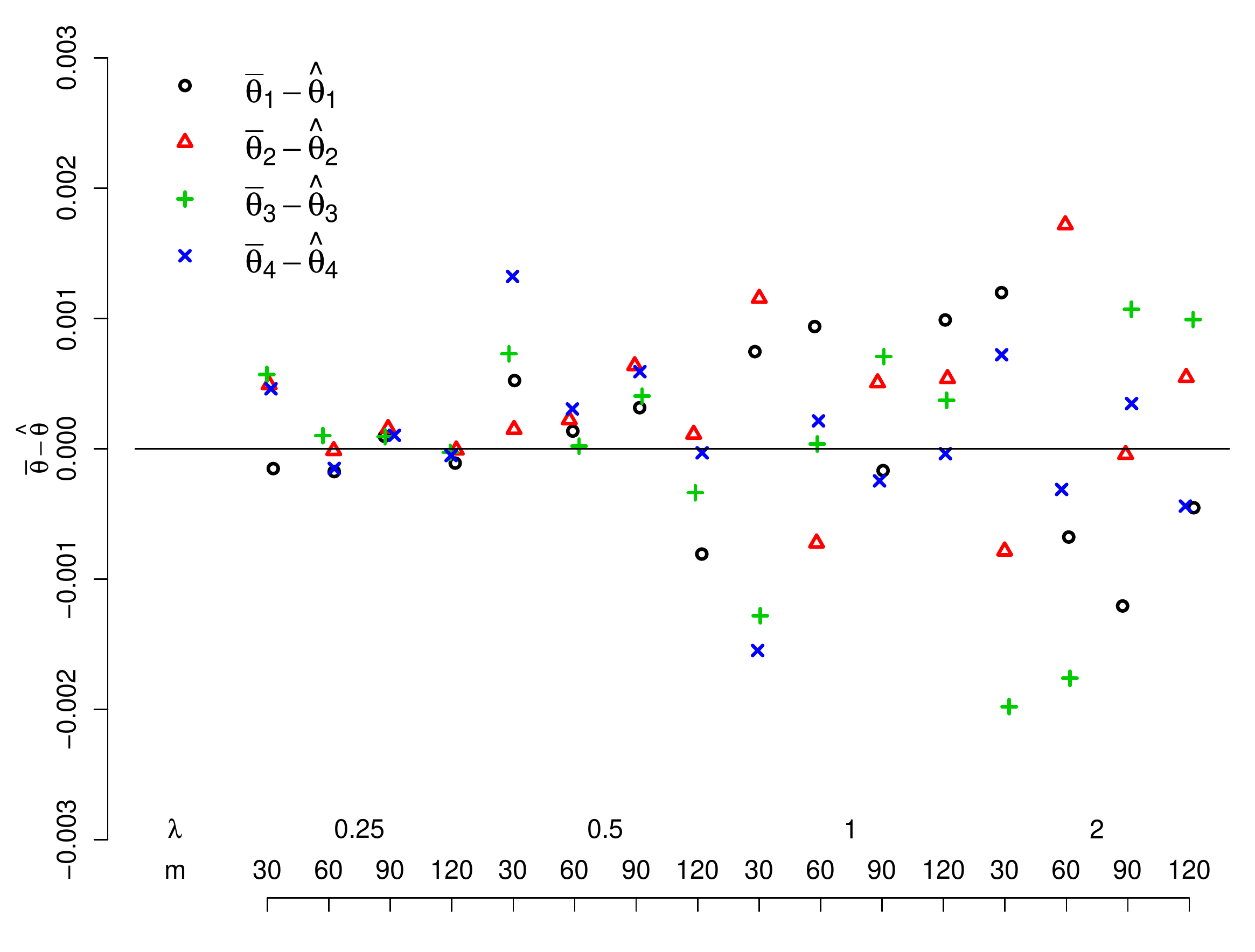}\\
Poisson regression \\
\includegraphics[width=0.75\textwidth,
height=0.6\linewidth]{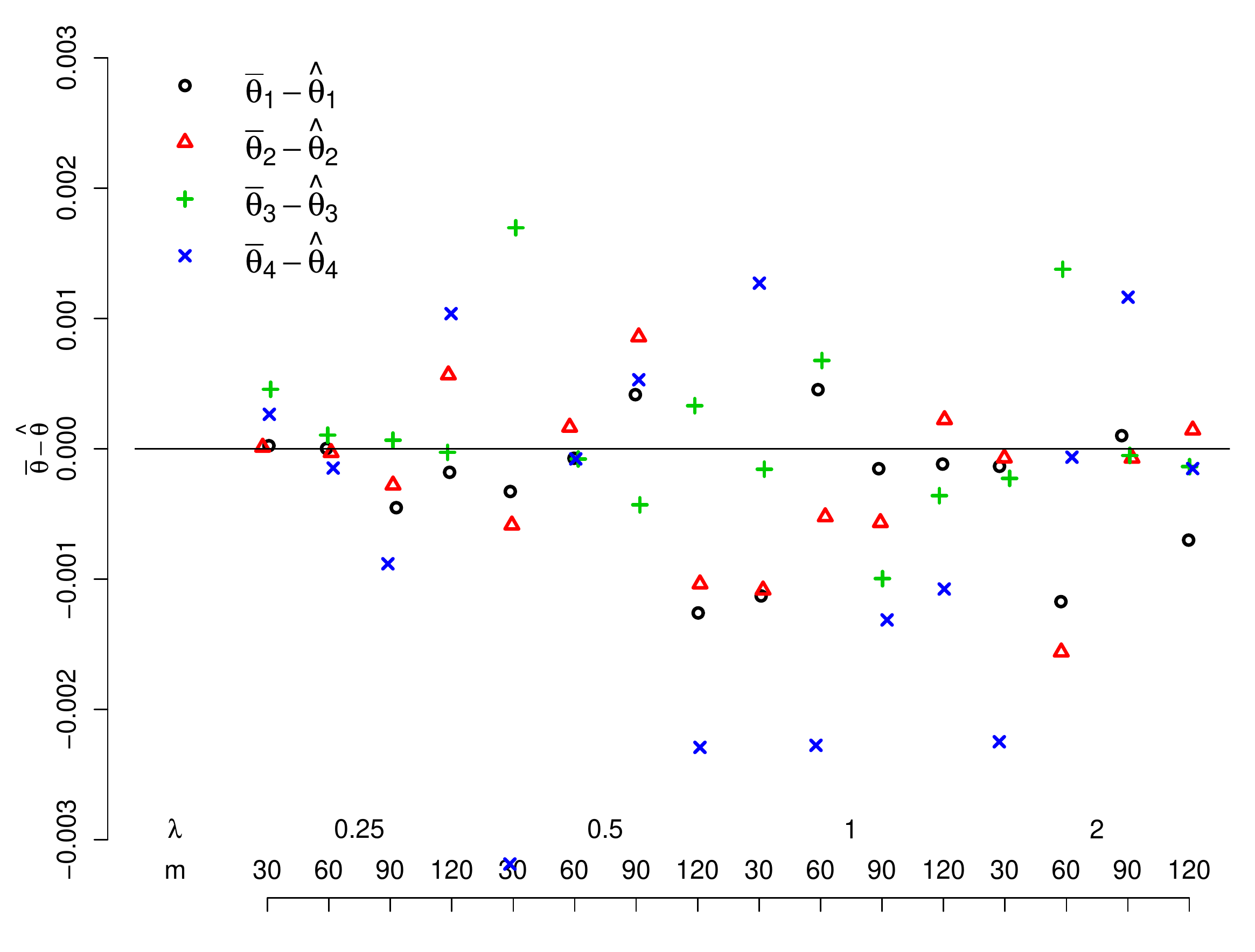}
\end{figure}

\end{document}